\documentclass[10pt,twocolumn,letterpaper]{article}

\usepackage[pagenumbers]{wacv} 

\definecolor{wacvblue}{rgb}{0.21,0.49,0.74}
\usepackage[pagebackref,breaklinks,colorlinks,allcolors=wacvblue]{hyperref}

\usepackage{graphicx}
\usepackage{booktabs}

\usepackage{booktabs}
\usepackage{dsfont,adjustbox,multirow,amsfonts}
\usepackage{diagbox}
\usepackage{algorithm}
\usepackage{algpseudocode}
\usepackage{tikz,pgfplots}
\usepackage{colortbl}
\usepackage{xcolor}
\usepackage{vcell}
\usepackage{tocloft}

\usepackage{tabularx}
\usepackage{booktabs}
\usepackage{makecell}

\newcommand{\pa}{SloMo-Fast}
\newcommand*\rot{\rotatebox{65}}

\newcommand{\second}[1]{\textcolor{gray}{\textbf{#1}}}

\def\wacvPaperID{1075} 
\def\confName{WACV}
\def\confYear{2027}

\title{SloMo-Fast: Slow-Momentum and Fast-Adaptive Teachers for Source-Free Continual Test-Time Adaptation}

\author{Md Akil Raihan Iftee$^{1,}$\thanks{Corresponding author: \href{mailto:iftee1807002@gmail.com}{iftee1807002@gmail.com}}, Mir Sazzat Hossain$^{1}$, Rakibul Hasan Rajib$^{2}$, Tariq Iqbal$^{3}$,\\ Md Mofijul Islam$^{4,}$\thanks{Work does not relate to position at Amazon.}, M Ashraful Amin$^{1}$, Amin Ahsan Ali$^{1}$ and A K M Mahbubur Rahman$^{1}$\\
\\
$^1$Center for Computational \& Data Sciences, Independent University, Bangladesh\\
$^2$University of Central Florida, USA,
$^3$University of Virginia, USA,\\ $^4$Amazon GenAI, USA\\
}

\begin{document}
\maketitle
\vspace{-5pt}
\begin{abstract}
Continual Test-Time Adaptation (CTTA) is crucial for deploying models in real-world applications with unseen, evolving target domains. Existing CTTA methods, however, often rely on source data or prototypes, limiting their applicability in privacy-sensitive and resource-constrained settings. Although several methods attempt to mitigate catastrophic forgetting, they often fail to preserve long-term domain-specific knowledge across many domain shifts. Moreover, their relatively slow adaptation rates during domain transitions can cause error accumulation, allowing mistakes to propagate before effective adaptation occurs. To address these challenges, we propose SloMo-Fast, a source-free, dual-teacher CTTA framework designed for enhanced quick adaptability and generalization. It includes two complementary teachers: the Slow-Teacher, which exhibits slow forgetting and retains long-term knowledge of previously encountered domains to ensure robust generalization, and the Fast-Teacher rapidly adapts to new domains while accumulating and integrating knowledge across them. This framework preserves knowledge of past domains and adapts efficiently to new ones. Our extensive experiments show that SloMo-Fast consistently outperforms state-of-the-art methods across Cyclic Test-Time Adaptation (Cyclic-TTA), a CTTA benchmark that simulates recurring domain shifts, along with ten other CTTA settings, highlighting its ability to both adapt and generalize across evolving, revisited domains.
\end{abstract}
\vspace{-15pt}

\section{Introduction}
\label{sec:intro}

\begin{figure*}[!t]
    \centering

    \begin{minipage}{0.3\textwidth}
        \centering
        \includegraphics[width=\linewidth]{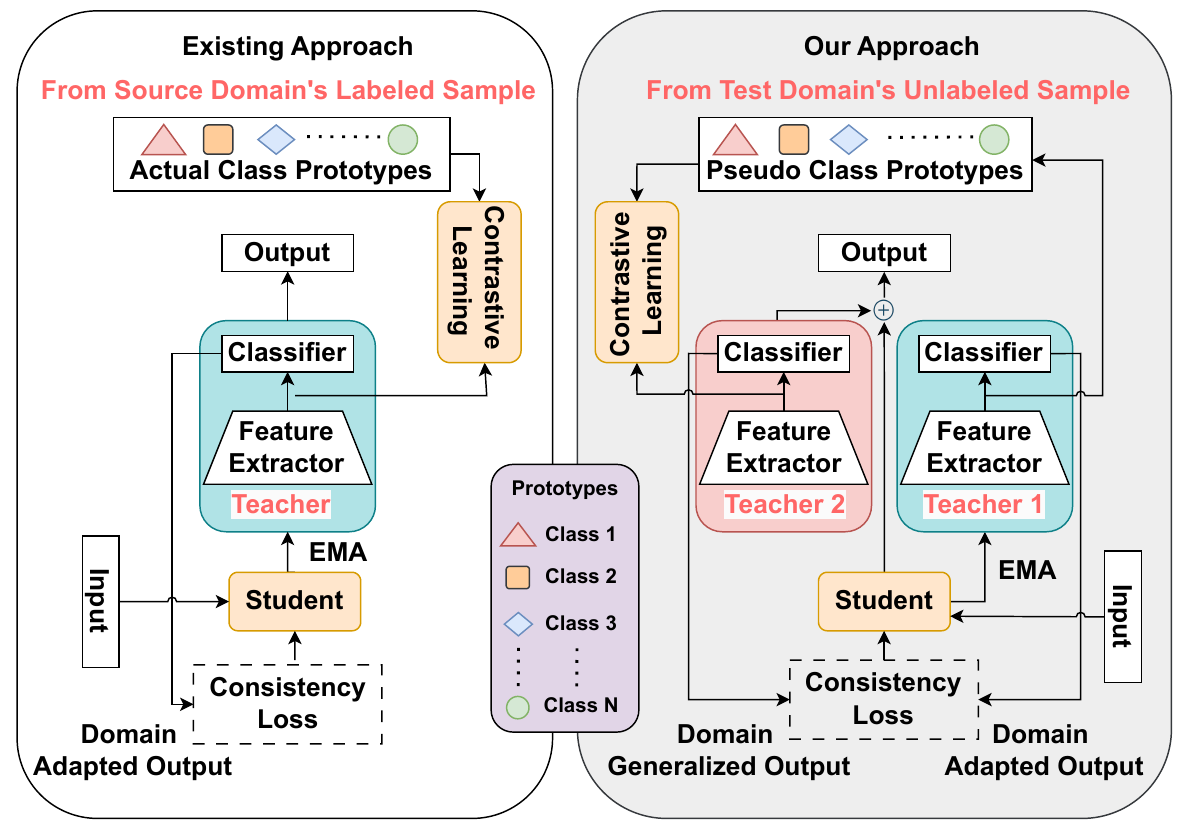}
    \end{minipage}
    \hspace{0.01\textwidth}
    \begin{minipage}{0.5\textwidth}
        \centering
        \includegraphics[width=\linewidth]{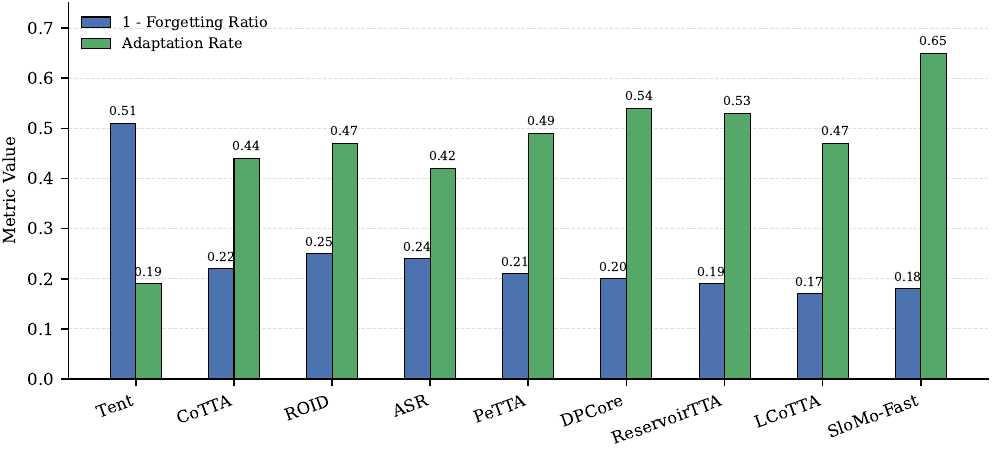}
    \end{minipage}

    \caption{Overview of CTTA methods using teacher--student and contrastive learning (left) versus SloMo-Fast with dual-teacher prototype generation during testing (right), achieving the best trade-off between forgetting and adaptation rate.}
    \label{fig:framework_intro}
    \vspace{-18pt}
\end{figure*}

Adapting models to dynamic environments is crucial for real-world deployment in areas such as autonomous driving, healthcare, and robotics, where systems must operate effectively under evolving conditions without prior knowledge of these changes. Unlike Test-Time Adaptation (TTA) such as TENT \cite{wang_tent_2021}, MEMO \cite{zhang_memo_2022}, and EATA \cite{niu_eata_2022}, which allows models to adapt to a single, unseen domain using only unlabeled test data, Continual Test-Time Adaptation (CTTA) \cite{wang_cotta_2022, sojka2023ar, liu2024continual, yang2024navigating} extends this by enabling models to adapt over time to a sequence of changing domains, crucial for real-world scenarios like autonomous driving, where environments evolve continuously with unpredictable weather and lighting conditions.

Recent advances in CTTA challenge the ability to achieve practical, effective adaptation for real-world applications. First, models must adapt to evolving, source-free \cite{kundu2020universal, xu2024revisiting, schlachter2025memory, deng2025multi} data streams in an online fashion \cite{mancini2018kitting, lee2024stationary}, requiring simultaneous prediction and continuous model updating for adaptation. Second, the dynamic nature of distribution shifts across continually changing target domains can compromise the reliability of pseudo-labels \cite{lee_deyo_2024, zhu2024reshaping}, hinder source knowledge retention, and lead to error accumulation \cite{wang_cotta_2022, wang2024continual} and catastrophic forgetting \cite{yang2023exploring, liu_vida_2024}. Third, ensuring robust generalization \cite{chen2023improved,iwasawa2021test} is crucial, as previously encountered domains may reappear during tests, necessitating the long-term retention of prior knowledge. 

To mitigate error accumulation and catastrophic forgetting in the source domain, CoTTA \cite{wang_cotta_2022} employs knowledge distillation in conjunction with stochastic restoration of the source model. Other approaches reduce knowledge forgetting by enforcing alignment in the parameter-space \cite{niloy2024effective, press2023rdumb}, feature-space \cite{chakrabarty2023santa, dobler_rmt_2023}, or through logit-driven energy minimization strategies \cite{yuan2024tea, choiadaptive}. However, achieving reliable alignment remains challenging in the absence of ground-truth labels during test time.

To better align target features with the source, it's essential to reference feature class boundaries to avoid excessive drift during adaptation. Self-training or knowledge distillation-based CTTA methods like RMT \cite{dobler_rmt_2023}, RoTTA \cite{yuan_rotta_2023}, AR-TTA \cite{sojka2023ar}, and DPLOT \cite{yu_dplot_2024} rely heavily on the quality of pseudo-labels from a teacher model. These methods often use memory banks to store source prototypes or labeled features, which help guide adaptation and preserve domain-specific characteristics. However, in real-world scenarios, accessing source data or prototypes is often restricted due to privacy concerns \cite{karani2021test}, storage limitations, or transmission constraints \cite{wang2024continual,niu_eata_2022}, making such approaches less viable in sensitive fields such as healthcare, finance, autonomous driving, and surveillance.

Identifying reliable samples during test-time adaptation remains a core challenge, especially under distribution shifts. To address this, several TTA methods directly minimize the entropy score \cite{wang_tent_2021, han2025ranked, wu2025multi} to encourage confident predictions. In contrast, others \cite{niu_eata_2022, niu2023towards, yuan_rotta_2023} rely on uncertainty or entropy-based sample selection to filter out high-entropy samples that may hinder adaptation. However, these approaches often degrade under spurious correlation shifts \cite{beery2018recognition}, TRAP factors \cite{lee_deyo_2024}, and biasness \cite{marsden2024universal}. Recent studies, such as Deyo \cite{lee_deyo_2024} and FATA \cite{cho2024feature}, have demonstrated that confidence-based metrics alone are insufficient for identifying trustworthy samples under diverse distribution shifts.

Recent methods, such as ROID \cite{marsden2024universal}, employ diversity and certainty-based weighting to retain past domain knowledge. VIDA \cite{liu_vida_2024} uses high- and low-rank adapters, MGG \cite{deng2025learning} refines gradients through historical memory, and TCA \cite{ni2025maintaining} enforces inter-class structure via topological constraints to mitigate CTTA challenges. Although effective in some continual settings, none of these approaches explicitly handle real-world CTTA scenarios involving cyclic domain arrivals, where domains may repeat over time, such as in autonomous driving or UAV applications, where weather patterns can recur. Furthermore, many of these methods lack robust domain generalizability and require re-adaptation when previously seen domains return, demonstrating a limited ability to retain long-term domain-specific knowledge.

While recent works \cite{hoang2024persistent, zhang2024dpcore, vray2026reservoirtta, duan2026lifelong, lim2026and} have begun to explore long-term and recurring domain scenarios, they each expose critical gaps when evaluated through the dual lens of Forgetting Ratio i.e. the accumulated performance drop on previously encountered domains  and Adaptation Rate i.e. the speed and stability of a model recovers performance upon re-encountering a familiar domain shown in Figure \ref{fig:framework_intro}.

To address these challenges, we propose SloMo-Fast, a dual-teacher framework that incorporates a student model, eliminating the need for source data while enhancing adaptability and generalization (Figure~\ref{fig:proposed_approach}). It employs two teachers: the Fast-Teacher ($T_1$), which adapts quickly to new domains, and the Slow-Teacher ($T_2$), which adapts gradually to ensure robust generalization (Figure~\ref{fig:framework_intro}). Unlike prior work that relies on source-based prototypes derived from labeled data, SloMo-Fast introduces a novel method for generating class-wise prototypes directly from unlabeled target data, utilizing high-confidence features filtered by the pseudo label probability difference (PLPD). The Fast-Teacher, updated via exponential moving average (EMA) without backpropagation, produces these prototypes, which guide the Slow-Teacher through contrastive learning to support robust cross-domain generalization. While ($T_1$) ensures fast and low-cost adaptation, ($T_2$) updates only batch normalization layers, thereby maintaining efficiency. This dual-teacher design enables effective adaptation to current domains while preserving knowledge of previously encountered ones, ensuring reliable pseudo-labels and robust generalization performance in dynamic, continually evolving real-world environments.

The key contributions of our work are as follows:
\begin{itemize}
    \item We propose SloMo-Fast, a dual-teacher CTTA framework that eliminates the need for source data while enhancing adaptability and generalization. The Fast-Teacher ($T_1$) adapts quickly to new domains, while the Slow-Teacher ($T_2$) ensures robust generalization by adapting gradually. 
    
    
    \item We introduce a test time reliable prototype generation approach to refine the Slow-Teacher for learning generalized representations across domains. The feature prototypes of each class are generated dynamically at test time without requiring source data.
    

    \item SloMo-Fast consistently outperforms with a mean error rate of 34.7\%, surpassing state-of-the-art methods by at least 1.5\% across 11 diverse CTTA scenarios across five datasets, demonstrating robust test time domain adaptation and generalization.

\end{itemize}

\section{Related Works}

CTTA aims to adapt models online under evolving, unlabeled target domains while preventing catastrophic forgetting and maintaining generalization challenges. 

\noindent
\textbf{Continual Test Time Adaptation (CTTA):} Recent CTTA methods address challenges such as sample selection, forgetting, and domain generalization. To improve sample reliability, DeYo \cite{lee_deyo_2024} introduced the Pseudo-Label Probability Difference, enabling more stable selection under distribution shifts. 
To alleviate catastrophic forgetting across sequential domains, CMF \cite{lee_cmf_2024} proposed Continual Momentum Filtering, while BECoTTA \cite{lee_becotta_2024} addressed evolving domain characteristics by selectively updating low-rank domain experts. Beyond these, MGG \cite{deng2025learning} introduced a Meta Gradient Generator with a lightweight Gradient Memory Layer to refine noisy gradients. TCA \cite{ni2025maintaining} strengthened feature consistency by enforcing topological constraints that minimize centroid distortion across class relationships. Finally, PaLM \cite{maharana2025palm} selects which layers to update based on gradient norms from KL divergence, while freezing others and applying layer-specific learning rates. 

\noindent
\textbf{Knowledge Distillation based CTTA:} Knowledge distillation has been widely applied in CTTA to strike a balance between stability and adaptability. CoTTA \cite{wang_cotta_2022} introduced a teacher–student framework specifically designed for non-stationary environments, laying the foundation for many subsequent approaches. EcoTTA \cite{song_ecotta_2023} extended this idea by improving memory efficiency through meta-networks and self-distilled regularization, while RoTTA \cite{yuan_rotta_2023} incorporated a time-aware reweighting strategy to account for sample timeliness and uncertainty. Seeking to avoid overfitting during continual updates, PSMT \cite{tian2024parameter} selectively updates specific network parameters, whereas VIDA \cite{liu_vida_2024} explores adaptability–forgetting trade-offs using high- and low-rank domain adapters. Additionally, OBAO \cite{zhu2024reshaping} introduced an uncertainty-aware buffer for reliable samples along with a graph-based constraint to preserve class relations. 


\noindent
\textbf{CTTA under Diverse TTA Settings:} Several methods address adaptation across different shift patterns. Tent introduces \emph{episodic TTA}, adapting to a single unseen domain, whereas CoTTA extends this to \emph{continual} adaptation over evolving domains. RMT \cite{dobler_rmt_2023} tackles continual and gradual shifts using a robust mean-teacher with contrastive alignment. GTTA \cite{marsden_gtta_2024} handles long sequences by generating intermediate domains via mixup and lightweight style transfer. ROID \cite{marsden2024universal} introduced \emph{universal} TTA by evaluating mixed or shuffled domain orders through weight ensembling, certainty/diversity-based weighting, and adaptive prior correction. The CCC benchmark \cite{press2023rdumb} further tests TTA methods under infinitely long, smoothly evolving domains to assess long-horizon stability.

\textbf{Long-term or recurring domain shifts:} PeTTA~\cite{hoang2024persistent} introduces a recurring TTA diagnostic and proposes collapse-sensing adaptation, yet validates only on CIFAR-10-C with a single baseline and offers no structured cyclic benchmark or per-cycle knowledge recovery. DPCore~\cite{zhang2024dpcore} proposes a Continual Dynamic Change setting where domains recur with variable duration, but relies on source statistics. ReservoirTTA~\cite{vray2026reservoirtta} maintains a reservoir of domain-specialized models to handle recurring and evolving shifts, yet its multi-model design grows unboundedly with domain count and is never evaluated on structured corruption-subgroup cycles or multi-dataset benchmarks. LCoTTA~\cite{duan2026lifelong} suppresses entropy-deceptive gradient updates by constraining adaptation to a low-dimensional gradient subspace for lifelong stability, but accumulates no domain memory and therefore cannot improve performance when previously seen domains return. ASR~\cite{lim2026and} proposes adaptive and selective resets to prevent collapse under long-term shifts, but deliberately erases acquired knowledge at reset events, a behavior that is actively harmful in cyclic settings where prior domain knowledge should be reused.

\section{Methodology}
\label{sec:methodology}

We address the task of adapting a pre-trained model to perform effectively in a continuously evolving target domain. The initial model, \( f_{\theta_0} \) with parameters \( \theta_0 \), is trained on a source dataset \((X^s, Y^s)\). Our goal is to enhance the model's performance during inference in a dynamic environment where data distributions evolve, without access to the source data. At each time step \( t \), the model receives new target data \( x_t \), predicts \( f_{\theta_t}(x_t) \), and updates its parameters from \( \theta_t \) to \( \theta_{t+1} \) to better adapt to the changing distributions.

\begin{figure*}[!t]
    \centering
    \includegraphics[width=0.85\linewidth]{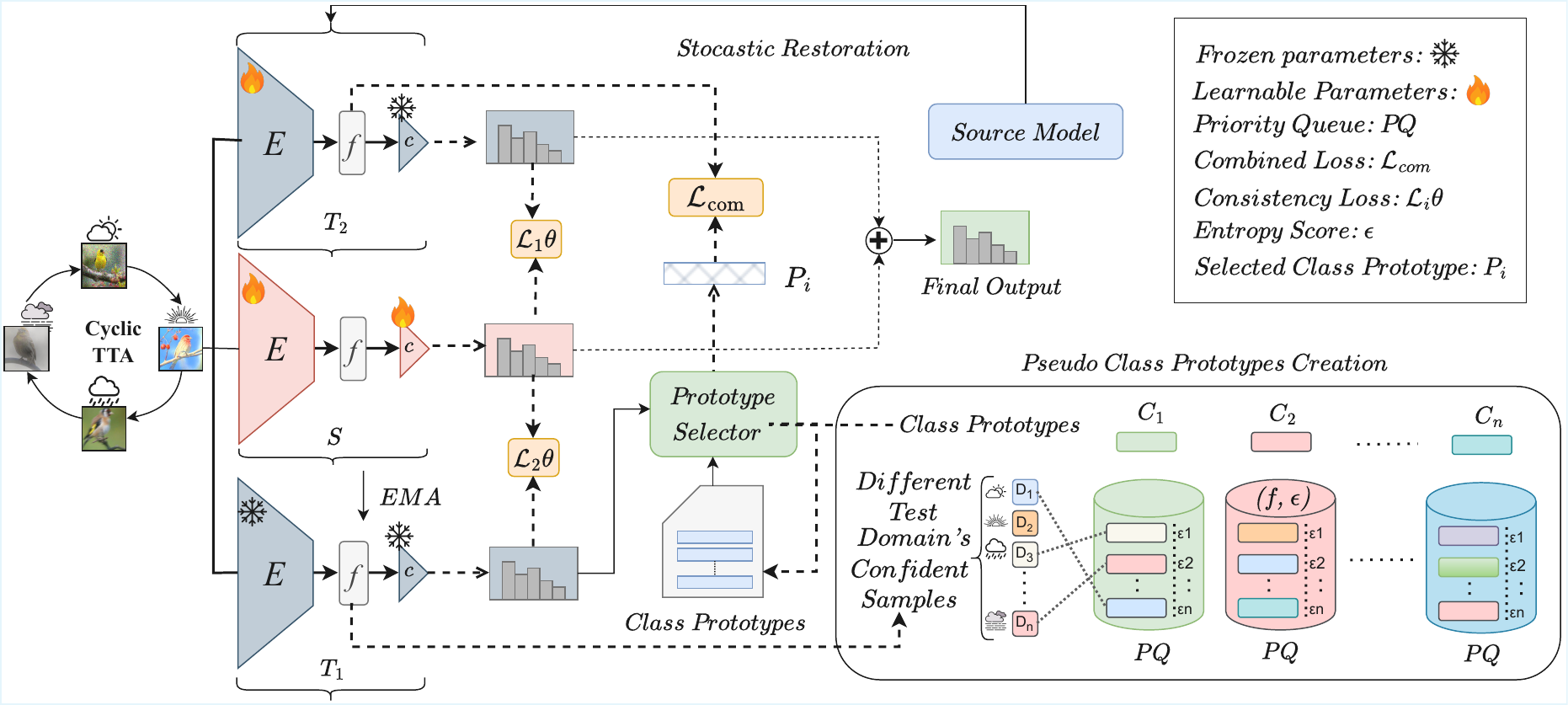}
    
    \caption{The SloMo-FAST framework comprises a dual-teacher and student model. The fast teacher $T_1$ quickly adapts to the current domain by taking the exponential moving average (EMA) of the student. Confident feature vectors from $T_1$ are used to construct robust class prototypes via a priority queue, which refines the slow teacher $T_2$ through contrastive learning. This enables $T_2$ to learn domain-invariant generalized representations while preserving knowledge from previous domains.}
    \label{fig:proposed_approach}
\end{figure*}

Fig. \ref{fig:proposed_approach} provides an overview of our method, which incorporates two teacher models and a student model. All models share the same architecture, feature extractor, and classifier and are initialized with $\theta_0$, but differ in update strategies. The student model $S$, with weights $\theta_S$, is updated using symmetric cross-entropy, leveraging pseudo-labels from both teacher models. The fast-teacher model, $T_1$, updates its weights $\theta_{T_1}$ using an EMA of the student's weights, smoothing the student's learning process. The slow-teacher model, $T_2$, updates its weights $\theta_{T_2}$ by optimizing contrastive, mean squared error (MSE), and information maximization loss to learn domain-invariant features.

\subsection{Self-training with Dual Teachers}\label{ssec:self-training}
For each test sample $x_t$, the student model $S$ aligns its predictions with those of teachers $T_1$ and $T_2$ using symmetric cross-entropy (SCE)
instead of CE loss due to its robustness to noisy labels. For distributions $p$ and $q$:
\begin{equation}
    \mathcal{L}_{SCE}(p, q) = -\sum_{c=1}^C p(c)\log q(c) - \sum_{c=1}^C q(c)\log p(c).
\end{equation}

\noindent
The training objective for the student model $S$, leveraging predictions from teacher models $T_1$ and $T_2$, results in the following self-training or consistency loss:
\begin{equation}
\mathcal{L}_{ST}(x_t) = \mathcal{L}_{SCE}\big(f_{\theta_S}(x_t), f_{\theta_{T_1}}(x_t)\big) + \mathcal{L}_{SCE}\big(f_{\theta_S}(x_t), f_{\theta_{T_2}}(x_t)\big)
\end{equation}

After updating $S$, teacher $T_1$ is refreshed via EMA: $\theta_{T_1}^{t+1} = \alpha \theta_{T_1}^t + (1-\alpha)\theta_S^{t+1}$,

where $\alpha$ is a smoothing factor.

\subsection{Class Specific Prototype Generation}
To enable robust domain-generalized adaptation, we construct class-specific priority queues that store high-quality features from Fast-Teacher, $T1$. These queues are iteratively updated during test using a dual-criterion strategy based on entropy (confidence) and sensitivity (stability).

\noindent
\textbf{Pseudo Label Assignment: }  
For each test sample $x_t$, the Fast-Teacher $T_1$ generates a pseudo label:
$
    \hat{y}_{T_1} = \arg \max_{c} \, y_{T_1}(c),
$
where $y_{T_1}(c)$ denotes the softmax probability for class $c$.

\noindent
\textbf{Dual-Criterion Priority Queue Update:} Each class-specific priority queue $\mathcal{Q}_c$ maintains a maximum of $K$ elements, storing tuples $(z_t, \mathcal{H}_t)$ where $z_t$ is the extracted feature and $\mathcal{H}_t$ is the entropy of prediction:
\begin{equation}
    \mathcal{H}_t = - \sum_{c=1}^{C} y_{T_1}(c) \log y_{T_1}(c).
\end{equation}
\noindent
To ensure reliability, each feature is evaluated using two criteria: \textit{(i) Confidence Criterion:} The entropy $\mathcal{H}_t$ must be less than a predefined threshold $\sigma$. \textit{(ii)Sensitivity Criterion:} Entropy can mistakenly include overconfident but incorrect features, especially under domain shifts. To address this, \cite{lee_deyo_2024} introduces a sensitivity criterion where the prediction sensitivity $\Delta p_t$ must exceed a threshold $\delta$, defined as:
    \begin{equation}
        \Delta p_t = p(\hat{y}_z \mid x_t) - \mathbb{E}_{x' \sim \mathcal{A}(x_t)}[p(\hat{y}_z \mid x')],
    \end{equation}
    where $\hat{y}_z = \arg \max_c p(y = c \mid x_t, z)$ and $\mathcal{A}(x_t)$ denotes the set of augmentations of $x_t$.
The sensitivity criterion checks whether the prediction remains stable when the input is slightly altered using augmentations such as pixel, patch shuffling, or center occlusion. If a feature is sensitive, it is likely unreliable. By retaining only stable features, the model learns more general and robust representations that are more effective across different domains.

A feature $(z_t, \mathcal{H}_t)$ is inserted into $\mathcal{Q}_c$ only if both criteria are met. If the queue is not full, the feature is inserted directly. If full, the new feature replaces the one with the highest entropy in the queue, but only if $\mathcal{H}_t$ is lower.

\noindent
\textbf{Queue Maintenance: }  
To maintain diversity, every $p$ time steps, the element with the \emph{lowest} entropy is removed from each queue $\mathcal{Q}_c$. This prevents overfitting to early high-confidence predictions and allows inclusion of more representative features over time.

\noindent
\textbf{Reliable Feature Set: }  
The resulting set of retained features for class $c$ is:
\begin{equation}
    \mathcal{S}_c = \left\{ z_t \in \mathcal{Q}_c \;\middle|\; \mathcal{H}_t \leq \sigma, \; \Delta p_t \geq \delta \right\}.
\end{equation}
Only both confident and structurally stable features contribute to prototype construction for $T_2$.

\noindent
\textbf{Prototype Generation: }We generate class prototypes using the reliable feature subset stored in each priority queue. Prototypes are computed as a weighted average, where the weight is the inverse of entropy, normalized for consistency. For class $c$, the prototype $P_c$ is computed as:
\begin{equation} \label{eq:proto_gen}
    P_c = \frac{1}{w} \sum_{(z, \mathcal{H}(z)) \in \mathcal{S}_c} w_z z,
\end{equation}
where $w_z = \frac{1}{\mathcal{H}(z)}$ and $w = \sum_{(z, \mathcal{H}(z)) \in \mathcal{S}_c} w_z$. This encourages confident, diverse prototypes that generalize across domains. (See the Algorithm~\ref{algo} in Section~\ref{sec:prototype_creation} of the supplementary materials)

\noindent 
\textbf{Contrastive Learning with Class Prototypes: }  
To learn domain-generalized features, we apply contrastive learning with class prototypes, aligning same-class features across domains while separating different classes. During inference, we update class-specific priority queues with features from the teacher model $T_1$ and recompute prototypes using Equation~(\ref{eq:proto_gen}).

We focus on samples where $T_1$ is confident but $T_2$ is not, using the binary variable:
\begin{equation}
    n_i = \mathds{1}[\mathcal{H}(y_{T_1}^{(i)}) \le \sigma] \cdot \mathds{1}[\mathcal{H}(y_{T_2}^{(i)}) > \sigma]
\end{equation}

\noindent The selected features form the set $\mathbb{S} = \{s_i \mid n_i = 1\}$.

where $s_i$ is the feature from $T_2$ for the $i$-th sample.
For each feature in $\mathbb{S}$, we compute the cosine similarity with all class prototypes and select the nearest prototype to form a positive pair. To ensure invariance to input changes, we include the test sample's augmented view, resulting in a batch size of $3N$, where $N$ is the size of $\mathbb{S}$. Each batch consists of original features, augmented views, and prototypes. For \( i \in I := \{1, \dots, 3N\} \), let \( A(i) := I \setminus \{i\} \) and \( V(i) \) represent different views of sample $i$. 
Here, we use a non-linear projection layer to obtain \( z = \text{Proj}(s_i) \). The contrastive loss is defined as:
\begin{equation}
    \mathcal{L_{CL}} = - \sum_{i \in I} \sum_{v \in V(i)} \log \left( \frac{\exp \left( \text{sim}(z_i, z_v) / \tau \right)}{\sum_{a \in A(i)} \exp \left( \text{sim}(z_i, z_a) / \tau \right)} \right)
\end{equation}
where $\tau$ is the temperature, and $\text{sim}(u, v) = \frac{u^T v}{\|u\| \|v\|}$ is the cosine similarity.

\medskip
\noindent \textbf{Feature Alignment with MSE Loss:} We also apply an MSE loss that aligns features from the $T_2$ model with their corresponding class prototypes. Each test sample's feature \( z_i \) is matched with the prototype \( P_{\hat{y}_{T_1}^i} \) based on the pseudo-label from $T_1$:
\begin{equation}
    \mathcal{L}_{\text{MSE}} = \frac{1}{N} \sum_{i=1}^{N} \| z_i - P_{\hat{y}_{T_1}^i} \|^2
\end{equation}
\noindent where \( N \) is the number of selected test samples, and \( \hat{y}_{T_1}^i \) denotes the predicted final pseudo-label.

\begin{table*}[!t]

    \centering

    \resizebox{0.8\textwidth}{!}{%

    \begin{tabular}{clcccccccccccc|c}

        \toprule

        \textbf{Setting} & \textbf{Dataset} & \textbf{Source} & \textbf{TENT} & \textbf{RoTTA} & \textbf{CoTTA}  & \textbf{RMT} & \textbf{ViDA} & \textbf{SANTA} & \textbf{OBAO} & \textbf{PALM} & \textbf{TCA} & \textbf{ROID} &  \textbf{SloMo-Fast} & \textbf{SloMo-Fast*}\\

        \midrule

        \multirow{5}{*}{\textit{Continual}} 

        & CF10-C   & 43.5 & 20.0 & 19.3 & 16.5 & 16.9 & 18.7 & 16.1 & 15.8 & 15.5 & \second{15.4} & 16.2 & 15.8$\pm$0.09 & \textbf{14.8$\pm$0.07}  \\

        & CF100-C  & 46.4 & 62.2  &34.8 & 32.8 & 30.2 & \second{28.1} & 30.3 &29.0 & 30.1& 29.9 & 29.3 & 28.2$\pm$0.11 & \textbf{27.9$\pm$0.12} \\

        & IN-C  & 82.0 & 82.5  & 78.1 & 76.0 & 59.9 & 61.2& 60.3 &59.0 & 60.1 &  59.8& 54.5 & \second{54.2$\pm$0.10} & \textbf{52.8$\pm$0.23} \\

        & IN-R & 63.8 & 57.6  & 60.7 & 57.4 &  56.1 & 59.1 & 57.2 & 58.6 & 59.9 &  58.9 & 51.2 &\second{51.0$\pm$0.05} & \textbf{50.4$\pm$0.07} \\ 

         & IN-S & 75.9 & 69.5  & 70.8 &  69.5 & 68.4 & 70.7 & 69.5 & 68.7 & 68.5 & 69.2& \second{64.3} & 65.2$\pm$0.13 & \textbf{64.1$\pm$0.21} \\

        \midrule

        \multirow{3}{*}{\textit{Mixed}} 

        & CF10-C   & 43.5 & 44.1 &  32.5 &   33.4 & 31.0 & 35.2 & 29.5 & 31.2 & 30.1 & 30.4 & \second{28.4} & 29.7$\pm$0.09  & \textbf{28.0$\pm$0.06} \\

        & CF100-C  & 46.4 & 82.5  & 43.1 &  45.4 & 38.6 & 38.9 & 38.2 & 37.2 & 38.1 & 37.6 &  35.0 & \second{34.4$\pm$0.15} & \textbf{33.5$\pm$0.02} \\

        & IN-C  & 82.0 & 86.4 & 78.1  &  79.4 & 75.4 & 76.0 &75.5 & 74.1& 74.6 & 73.9 & \textbf{69.5} & 71.3$\pm$0.11 & \second{70.8$\pm$0.27} \\

        \midrule

        \multirow{3}{*}{\textit{Gradual}} 

        & CF10-C  & 43.5 & 26.2 &  11.8 & 10.8 & 10.4 &12.1 & 10.9 & 10.5 & \second{9.7} & 10.1 &10.5  & 10.4$\pm$0.11 & \textbf{8.9$\pm$0.09}\\

        & CF100-C & 46.4 & 75.9  & 33.4 &    27.0 & 26.9& 24.7 & 25.8& 25.8 & 26.2 & 26.6 & 24.3 & \second{24.1$\pm$0.25} & \textbf{23.3$\pm$0.33}\\

        & IN-C & 82.0 & 91.6  & 96.4 &  67.7 & 41.5 & 43.8 & 42.1 & 41.9 & 40.9 & 41.3 & \second{38.8} & 39.1$\pm$0.06 &  \textbf{37.9$\pm$0.17}\\

        \midrule

        \multirow{3}{*}{\textit{Episodic}} 

        & CF10-C  & 43.5 & 18.2 & 21.6  &  18.3 & 17.2 &18.1 & \second{16.9} & 18.7 & 17.1 & 18.8 &17.5 & 17.2$\pm$0.08 & \textbf{16.7$\pm$0.15}\\

        & CF100-C & 46.4 & 31.1  & 41.9   & 34.5 & 32.5 & 31.9 & \second{30.2}& 33.2& 31.9& 32.2&  30.4 & 30.7$\pm$ 0.31 & \textbf{30.1$\pm$0.13}\\

        & IN-C & 82.0 & 57.3 & 63.70 &  61.5 & 57.3 & 60.0 & 57.1 & 59.4& 57.3& 59.0 & \second{52.6} & 53.1$\pm$0.25 & \textbf{52.7$\pm$0.14}\\

        \midrule

        \multirow{3}{*}{

            \textit{\shortstack{Cross Group\\(Continual)}}

        }

        & CF10-C   & 43.5 & 15.8  & 18.8 & 19.7 &  16.5 & 17.1 & \second{15.3} & 16.8 & 15.9 & 15.8 &16.4 & 16.0$\pm$0.18 &  \textbf{14.7$\pm$0.09} \\

        & CF100-C  & 46.4 & 61.5  & 32.5 &   34.9 & 28.7 & \second{28.5} & 28.9 &29.2 & 30.1& 30.8 & 29.5 & 28.7$\pm$0.14 & \textbf{27.9$\pm$0.11} \\

        & IN-C  & 82.0 & 62.2  & 68.6 & 59.2 & 58.3& 59.9 & 58.7& 59.5& 57.3& 56.9 & 55.7 & \second{54.3$\pm$0.22} & \textbf{52.3$\pm$0.18}\\

        \midrule

        \multirow{3}{*}{

            \textit{\shortstack{Easy2Hard\\(Continual)}}

        }

        & CF10-C  & 43.5 & 19.6  & 17.8 & 15.7 &16.1 & 17.5 & 15.3 & \second{14.8} & 14.9 & 15.7 &15.9 & 15.5$\pm$0.18 & \textbf{13.9$\pm$0.13}\\

        & CF100-C & 46.4 & 52.8  & 33.0 & 32.2 & 29.7 & 29.3 & \second{28.9} & 29.6& 30.4 & 29.9& 29.3 & 29.2$\pm$0.12 & \textbf{28.2$\pm$0.14} \\

        & IN-C & 82.0 & 60.0   & 65.1 & \second{52.5} & 57.1 &58.5 & 55.9 & 59.3 & 58.3 & 59.2&  54.3 & 52.8$\pm$0.31 & \textbf{48.6$\pm$0.13}\\

        \midrule

         \multirow{3}{*}{

            \textit{\shortstack{Hard2Easy\\(Continual)}}

        } 

        & CF10-C  & 43.5 & 21.6  & 19.4 & 17.1 & 16.6 &18.5 & 17.5 & \second{16.1} & 16.3 & 16.7 &16.3 & 16.1$\pm$0.20 & \textbf{15.3$\pm$0.04}\\

        & CF100-C & 46.4 & 66.7  & 35.7 &   33.0 & 29.9 & 29.7 & 29.8 & 30.0& 30.3& 29.8& \second{29.5} & 30.0$\pm$ 0.11& \textbf{28.2$\pm$0.06} \\

        & IN-C & 82.0 & 62.8  & 68.4 & 63.2 & 58.3& 61.3 & 60.9& 60.4& 60.0& 59.1 & 55.1 & \second{54.3$\pm$0.22} & \textbf{52.8$\pm$0.17}\\

        \midrule

        \multirow{3}{*}{

            \textit{\shortstack{Mixed\\ After Continual\\(Overlapping)}}

        } 

        & CF10-C   & 43.5 & 21.3  & 19.9 & 16.8 & \second{16.5}  & 19.3 & 17.5 & 17.1 & 17.9 & 16.5 &16.9 & 16.7$\pm$0.12& \textbf{16.2$\pm0.15$}\\

        & CF100-C  & 46.4 & 63.7  & 35.1 &  33.2 & 30.4&30.9& 32.1& 29.7 &31.8 & 30.5 &\second{29.6} & 30.2$\pm$0.17 & \textbf{28.1$\pm0.10$}\\

        & IN-C  & 82.0 & 91.8 & 75.4 & 70.9 & 61.2 & 63.6& 63.5 & 61.8& 62.5& 59.0 &\textbf{52.5} & 53.4$\pm$0.19& \second{52.6$\pm$0.11} \\

        \midrule

        \multirow{3}{*}{

            \textit{\shortstack{Continual\\ After Mixed\\(Overlapping)}}

        } 

        & CF10-C   & 43.5 & 46.0   & 18.9 & 17.8 & 17.6 & 18.8 & 16.7 & \second{16.2} & 17.4 & 17.1 &16.7 &16.6$\pm$0.06 & \textbf{14.6$\pm$0.07}\\

        & CF100-C  & 46.4 & 97.1 & 34.5 & 33.3 & 33.2 & 35.5 & 36.1 & 32.3& 31.4& 30.2 &29.4 & \second{29.1$\pm$ 0.18} & \textbf{26.6$\pm$0.17} \\

        & IN-C  & 82.0 & 85.6 &  64.2 &  55.3 & 58.3 & 59.4& 60.7& 59.7& 60.6& 60.2 &\second{55.1} & 55.6$\pm$0.20 & \textbf{54.7$\pm$0.20}   \\

        \midrule 
        \textit{CCC} & IN-C & 93.4 & 98.1 & 94.7 & 92.1 & 68.7 & 78.8 & 70.1 & 69.7 & 67.5 & 67.1 & 67.2 & \second{65.3$\pm$0.23} & \textbf{64.7$\pm$0.12} \\
        \midrule \midrule

        \rowcolor[gray]{0.9}

        & CF10-C  & 43.5 & 17.0 & 19.4 & 16.7  & 16.8 & 17.4& 17.5 &17.2 & 16.7& 16.9&16.3 & \second{15.2$\pm$0.07} & \textbf{14.6$\pm$ 0.08} \\

        \rowcolor[gray]{0.9}

        &  CF100-C & 46.4 & 34.7  & 37.8 & 33.7  & 32.2& 33.8 & 34.8& 31.5& 29.3 & 30.1 &29.4 &  \second{28.6$\pm$0.27} & \textbf{27.5$\pm$0.15}\\

        \rowcolor[gray]{0.9}

        \multirow{-3}{*}{\textit{Cyclic}} & IN-C & 82.0 & 59.7 & 66.1 & 63.7 & 65.5& 66.5 & 66.9& 63.6& 59.3 & 60.5&54.6 & \second{52.7$\pm$0.11} & \textbf{51.9$\pm$0.23} \\

        \midrule \midrule

    \rowcolor[gray]{0.9}
\textit{Mean Error Rates} & All 
& 59.2 
& 55.8 
& 44.1 
& 42.1 
& 38.6 
& 40.1 
& 38.9 
& 38.7 
& 38.4 
& 38.3 
& 36.3 
& \second{35.9$\pm$0.13} & \textbf{34.7$\pm$0.21}
\\

        \bottomrule

    \end{tabular}}

    \caption{Average online classification error rate (\%) over five runs for different TTA settings (definition of them in \ref{benchmarks} and \ref{cyclic}) across multiple datasets. The table includes results for various TTA methods: TENT-cont., RoTTA, CoTTA, ROID, SloMo-Fast, and SloMo-Fast* (where all parameters of the student model are updated), evaluated in different settings. Results are shown for CIFAR10-C (CF10-C), CIFAR100-C (CF100-C), ImageNet-C (IN-C), ImageNet-R (IN-R), and ImageNet-Sketch (IN-S) datasets. Results in bold represent the best performance, while those in gray represent the second-best performance. (See the Details Results of Each TTA setting in Section~\ref{sec:details_result} of the Supplementary Materials)}

    \label{tab:summary_table}

\end{table*}

\medskip
\noindent \textbf{Information Maximization Loss:} To ensure $T_2$ provides strong guidance to the student model $S$, it must maintain both discriminability and diversity in its predictions. Following state-of-the-art unsupervised domain adaptation methods, 
 we use an information maximization loss, $\mathcal{L}_{\text{IM}}$, comprising two components:
\begin{equation} \label{eq:information_loss}
    \mathcal{L}_{\text{IM}} = - \mathbb{E}_{x_t \in X_t}\sum_{c=1}^{C}y_{T_2}(c) \log (y_{T_2}(c)) - \sum_{c=1}^{C}\bar{q}(c)\log(\bar{q}(c)),
\end{equation}
where the first term enhances individual prediction certainty, and the second term promotes variation across class distributions.

The overall training objective for the teacher model \( T_2 \) is defined as follows:
\begin{equation}
    \mathcal{L}_{T2} = \lambda_{\text{cl}} \mathcal{L}_{\text{CL}} + \lambda_{\text{mse}} \mathcal{L}_{\text{MSE}} + \lambda_{\text{im}} \mathcal{L}_{\text{IM}}
\end{equation}
where \( \lambda_{\text{cl}} \), \( \lambda_{\text{mse}} \), and \( \lambda_{\text{im}} \) represent the weighting factors for the contrastive loss, the MSE loss, and the information maximization loss, respectively.

To address error accumulation from distribution shifts, we use a stochastic restoration method \cite{wang_cotta_2022} that combines the pretrained source model's original weights with updated weights after each gradient step. This approach mitigates catastrophic forgetting by selectively restoring weights, preserving knowledge from the source model.

\subsection{Prediction Ensembling }
\label{ssec:pred_ensemble}
 We combine the outputs of both the student and \( T_2 \) models as the student model learns the current domain from \( T_1 \), while the \( T_2 \) model provides generalized predictions across domains. For a test sample \( x_t \), the final prediction is:
$\label{eq: final_pred}
    y_t = f_{\theta_S}(x_t) + f_{\theta_{T_2}}(x_t) $
This combination leverages their complementary strengths, improving prediction robustness and accuracy in dynamic environments.
\medskip

\noindent \textbf{Dynamic Label Calibration (DLC):} 
To mitigate performance degradation caused by the shift in label distributions between the source domain and the target domain, we recalibrate the predictive posterior $P_t(y|x)$ as follows:
\begin{equation}
P_t(y|x) \propto P_s(y|x) \cdot \frac{\pi_t(y)}{\pi_s(y)}
\end{equation}
\noindent Assuming a uniform source domain prior $\pi_s(y)$, we estimate the target domain prior $\pi_t(y)$ by averaging softmax outputs across the current batch. To ensure stability during small-batch inference, we employ a smoothed estimation for each class:
\begin{equation}
\bar{\pi}_t = \frac{\sum_{i=1}^{B} \hat{y}_i + \alpha}{B + \alpha K}
\end{equation}
\noindent where $B$ is the batch size, $K$ is the number of classes, and $\alpha$ is a smoothing hyperparameter. This allows the model to dynamically adapt its decision boundaries to the evolving test stream.

\section{Result and Discussion}
\label{sec:result_and_discussion}

In this section, we analyzes experimental results across diverse datasets and CTTA settings of {\pa}  and comparison with baselines.

\subsection{Implementation Details}
We evaluate our approach on diverse domain shifts, including artificial corruptions and natural variations. Following \cite{marsden2024universal}, we use the corruption benchmark on CIFAR10-C (C10-C), CIFAR100-C (C100-C), and ImageNet-C \cite{hendrycks2019benchmarking} (IN-C), which applies 15 corruption types at five severity levels. Additionally, we assess our method on ImageNet-R (IN-R) \cite{hendrycks2021many} and ImageNet-Sketch (IN-S) \cite{wang2019learning}. We use a priority queue size of 10, a batch size of 200 for C10-C and C100-C, and 64 for IN-C, IN-R, and IN-S. We use a single set of hyperparameters shown \cref{sec:4.6} across all benchmarks (11 CTTAs, 5 Datasets) (\cref{tab:summary_table}). So, our performance gains result from a principled design rather than exhaustive, per-dataset tuning.

\subsection{Result for Different TTA Setting}

The proposed SloMo-Fast framework consistently achieves state-of-the-art (SOTA) performance across nearly all evaluated benchmarks, reducing the overall mean error rate from 59.2\% (Source) to a leading 34.7\%, as shown in Table \ref{tab:summary_table} (See their definitions in the supplementary material in section \ref{benchmarks}. Our method demonstrates exceptional robustness in high-complexity and long term domain knowledge retaining scenarios like the CCC, Gradual and Cyclic settings, where it significantly outperforms baselines that suffer from long term catastrophic collapse. An interesting observation is that our lighter SloMo-Fast variant, which only updates BN parameters, achieves a mean error of 35.9\%, effectively surpassing the full-parameter ROID (36.3\%) and making it a highly efficient alternative. While SloMo-Fast* is the overall leader, ROID remains highly competitive in specific scenarios, maintaining a slight edge in the Mixed IN-C (69.5\%) and Episodic IN-C (52.6\%) tasks. Overall, the framework's consistent performance across continual, mixed, Easy2Hard, and Hard2Easy, transitions confirms its superior adaptability to diverse and evolving distribution shifts.

\begin{table}[ht]
\centering
\small
\setlength{\tabcolsep}{0.4em} 
\resizebox{0.9\linewidth}{!}{%
\begin{tabular}{llccccccc}
\toprule
\textbf{Dataset} & \textbf{Architecture} & \textbf{Source} & \textbf{TENT} & \textbf{RoTTA} & \textbf{CoTTA} & \textbf{ROID} & \textbf{Slomo-Fast} & \textbf{Slomo-Fast*} \\
\midrule

\multirow{3}{*}{IN-C} & ResNet-50   & 82.0 & 62.6 & 65.5 & 66.5 & 54.5 & 53.2 & \textbf{52.0} \\
                      & Swin-b      & 64.0 & 61.4 & 62.7 & 59.2 & 47.0 & 46.2 & \textbf{45.1} \\
                      & ViT-b-16    & 60.2 & 54.5 & 58.3 & 77.0 & 45.0 & 44.1 & \textbf{43.3} \\
\midrule
\multirow{3}{*}{IN-R} & ResNet-50   & 63.8 & 57.5 & 52.9 & 53.0 & 51.2 & 50.3 & \textbf{49.0} \\
                      & Swin-b      & 54.2 & 53.8 & --   & 52.9 & 45.8 & 44.9 & \textbf{43.8} \\
                      & ViT-b-16    & 56.0 & 53.3 & 54.4 & 56.0 & 44.2 & 43.5 & \textbf{42.5} \\
\midrule
\multirow{3}{*}{IN-S} & ResNet-50 & 75.9 & 68.7 & 66.4 & 67.0 & 58.6 & 57.1 & \textbf{56.0} \\
                           & Swin-b    & 68.4 & 68.7 & --   & 69.1 & 58.4 & 57.2 & \textbf{56.3} \\
                           & ViT-b-16  & 70.6 & 59.7 & 68.6 & 69.6 & 58.6 & 57.4 & \textbf{56.1} \\
\bottomrule
\end{tabular}}
\caption{\textbf{Observation:} Across a wide range of datasets and model architectures including convolutional networks (ResNet-50), hierarchical transformers (Swin-b), and plain vision transformers (ViT-b-16), \textbf{Slomo-Fast} and its enhanced variant \textbf{Slomo-Fast*} consistently outperform prior methods.}
\label{tab:final_merged_table}
\end{table}

\subsection{Results for Cyclic TTA Setting}
In the \textbf{Cyclic Setting}, involving recurring domain shifts, SloMo-Fast* shows remarkable stability and performance  across multiple full cycles with variable cycle lengths (\cref{cyclic_results}). As shown in Table~\ref{tab:summary_table}, it achieves the lowest error rates of 14.6\% (C10-C), 27.5\% (C100-C), and 51.9\% (ImageNet-C), consistently outperforming all baselines, including its lighter variant SloMo-Fast. These results confirm its strong adaptability to dynamic distribution shifts. 
The detailed results of the cyclic setting have been shared in \cref{cyclic_results,table:cyclic_tent_cifar10c,table:cyclic_tent_cifar100c,table:cyclic_tent_imagenetc,table:cyclic_cotta_cifa100c,table:cyclic_cotta_imagenetc,table:cyclic_roid_cifar10c,table:cyclic_roid_cifar100c,table:cyclic_roid_imagenetc,table:cyclic_RoTTA_cifar10c,table:cyclic_RoTTA_cifar100c,table:cyclic_RoTTA_imagenetc,table:cyclic_ours_cifar10c,table:cyclic_ours_cifar100c,table:cyclic_ours_imagenetc,table:cyclic_ours*_cifar10c,table:cyclic_ours*_cifar100c,fig:cyclic_roid_pa_cifar10c}.

Table \ref{tab:cytta_error_rates} brings comparison between all the existing Long-term or recurring TTA methods with ours.
\begin{table}[htbp]
\centering
\caption{Comparison of Average Error Rates (\%) across Cyclic-TTA Benchmarks.}
\label{tab:cytta_error_rates}
\resizebox{0.95\linewidth}{!}{%
\begin{tabular}{lcccccc}
\toprule
\textbf{Dataset} & \textbf{ASR} & \textbf{PeTTA} & \textbf{DPCore} & \textbf{ReservoirTTA} & \textbf{LCoTTA} & \textbf{SloMo-Fast} \\
\midrule
\textbf{C10-C}   & 19.4\% & 18.2\% & 15.1\% & 14.6\% & 13.9\% & \textbf{12.4\%} \\
\textbf{C100-C}  & 40.5\% & 38.6\% & 32.9\% & 31.4\% & 29.8\% & \textbf{26.9\%} \\
\textbf{IN-C}    & 56.4\% & 55.2\% & 52.9\% & 52.4\% & 56.1\% & \textbf{51.9\%} \\
\bottomrule
\end{tabular}%
}
\end{table}

\subsection{Ablation Study on Adaptation Rate}
We introduce a metric, Adaptation Rate,  that quantifies how fast a model adapts after each domain shift. Our proposed metric is composed of three components:
\textit{(i) Time to Plateau (\textbf{TTP})}: The number of adaptation steps required for the moving average of accuracy (computed over a window of size $k$) to reach 80\% of its maximum value. A lower TTP indicates faster adaptation.
\textit{(ii) Average Positive Slope (\textbf{APS})}: The mean of all positive differences between consecutive moving averages of accuracy values. Higher APS means a faster upward trend in adaptation.
\textit{(iii) Stability (\textbf{STD})}: Stability quantifies the consistency of the adaptation process and is assessed using the standard deviation of the moving average of accuracy values. A lower STD implies greater adaptation stability. 

We define the composite score for adaptation rate as:
\begin{equation}
\text{Adaptation Rate} = \frac{\text{APS}}{\text{TTP}} - \lambda \cdot \text{STD}
\label{adap}
\end{equation}
where $\lambda$ is a stability regularizer. A higher score indicates a more desirable adaptation behavior: fast, strong, and stable.

\begin{figure}
    \centering
    \includegraphics[width=0.8\linewidth]{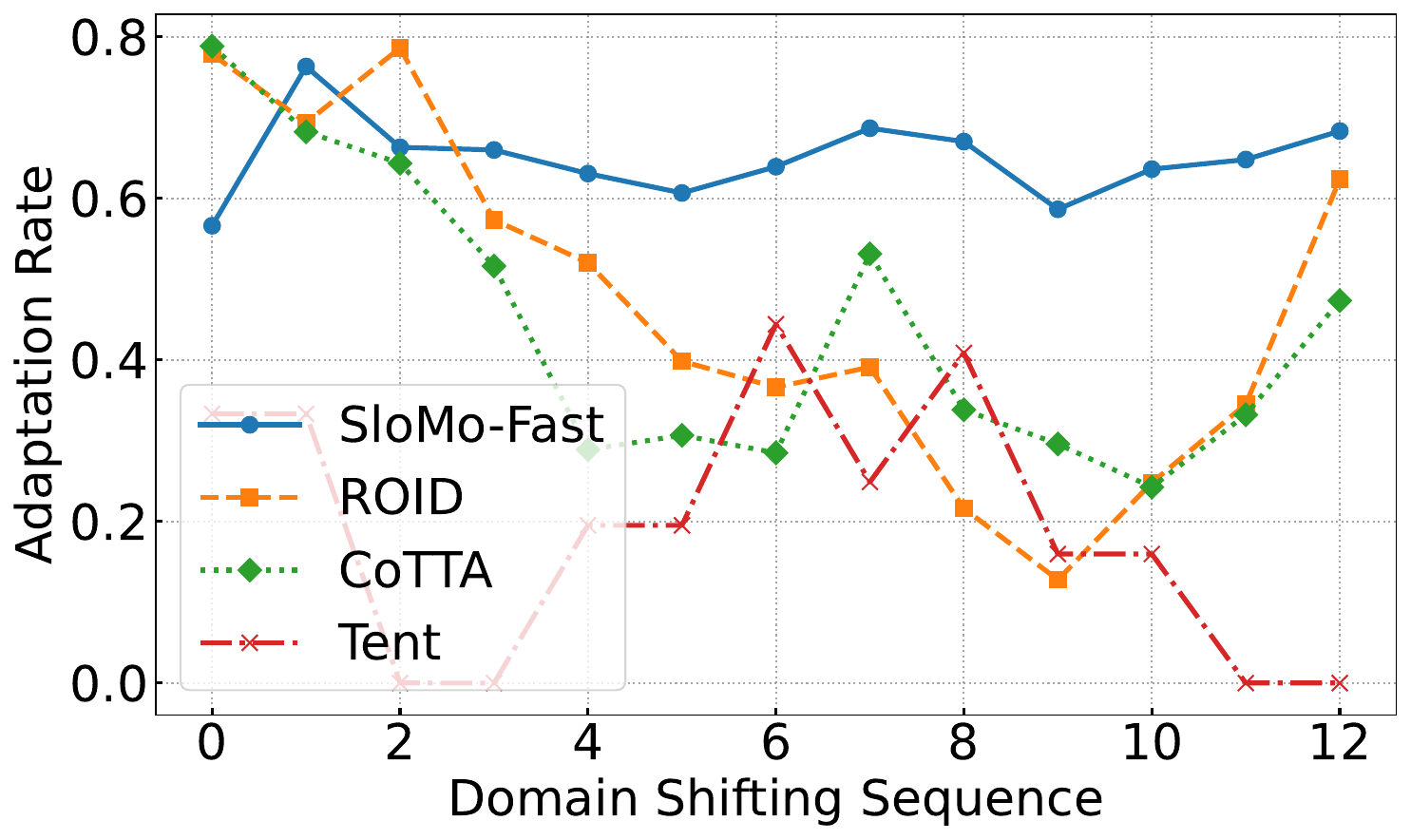}
    \caption{Adaptation rate over a cyclic domain shift. SloMo-Fast shows the best and most stable performance, benefiting from prototype memory and slow-teacher guidance.
}
    \label{fig:domain_shift}

\end{figure}

Using Equation \ref{adap}, we evaluate Slomo-Fast's adaptability under cyclic-TTA. As shown in Figure \ref{fig:domain_shift}, {\pa} achieves the highest adaptation rate, especially when previously seen domains reoccur. This demonstrates its fast adaptability, as it leverages stored knowledge from past domains through prototypes learned by the slow teacher.

\begin{figure}
    \centering
    \includegraphics[width=0.8\linewidth]{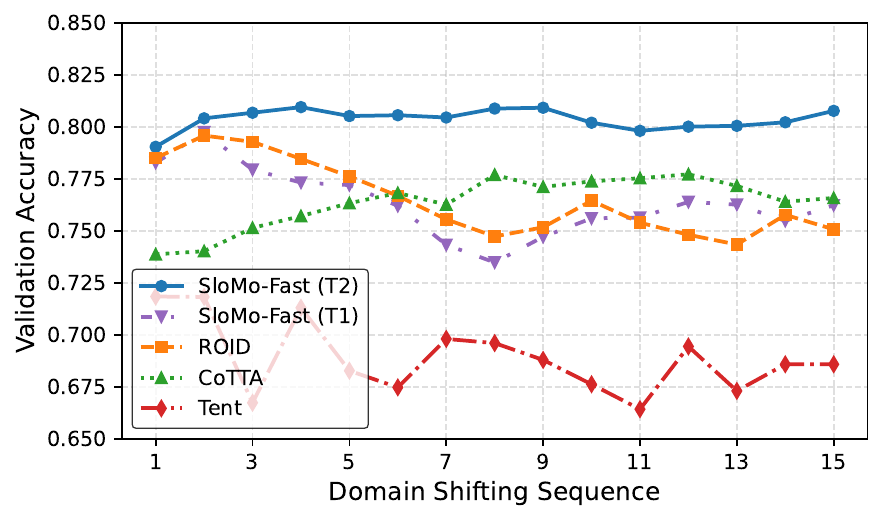}
    \caption{Validation accuracy on the previously encountered domains on the arrival of new domains with sequential adaptation. SloMo-Fast (T2) shows the slowest forgetting,
    highlighting its ability to retain long-term knowledge while adapting to new
    domains.}
    \label{fig:forgetting_main}

\end{figure}

\begin{table}[t]
\centering
\setlength{\tabcolsep}{1.2em} 
\resizebox{0.65\linewidth}{!}{%
\begin{tabular}{lrrr}
\toprule
\multicolumn{4}{c}{\textbf{Mean Error Rate (\%)}} \\
\midrule
\textbf{Components} & \textbf{CF10-C} & \textbf{CF100-C} & \textbf{IN-C} \\
\midrule
w/o $\mathcal{L}_{\text{MSE}}$ & 15.89 & 28.23 & 54.81 \\
w/o $\mathcal{L}_{\text{IM}}$ & 16.17 & 28.35 & 55.42 \\
w/o $\mathcal{L}_{\text{CL}}$ & 16.04 & 28.57 & 55.40 \\
w/o DLC & 15.78 & 28.08 & 54.47 \\
w/o ST & 16.11 & 28.48 & 55.43 \\
\midrule
\textbf{\pa} & \textbf{14.88} & \textbf{27.97} & \textbf{52.80} \\
\bottomrule
\end{tabular}}
\caption{Ablation study of mean classification error rates (\%) for online CTTA. The table shows the impact of removing individual components: $\mathcal{L}_{\text{MSE}}$, $\mathcal{L}_{\text{IM}}$, $\mathcal{L}_{\text{CL}}$, Dynamic Label Calibration (DLC), and Stochastic Restoration (ST).}
\label{tab:component}
\end{table}

\subsection{Various Components of {\pa}}

The ablation in Table \ref{tab:component} shows that removing any loss term or optimization strategy degrades performance on C10-C and CIFAR100-C, confirming each component’s value. Using all components achieves the lowest errors (14.8\% on C10-C, 27.9\% on C100-C, 52.9\% on IN-C), highlighting the effectiveness of combining complementary objectives for robust adaptation. 
(Details and Reasoning found in Section \ref{sup_ablation} in Supplementary Materials)

\begin{figure}[!t]
    \centering

    \includegraphics[width=0.7\linewidth]{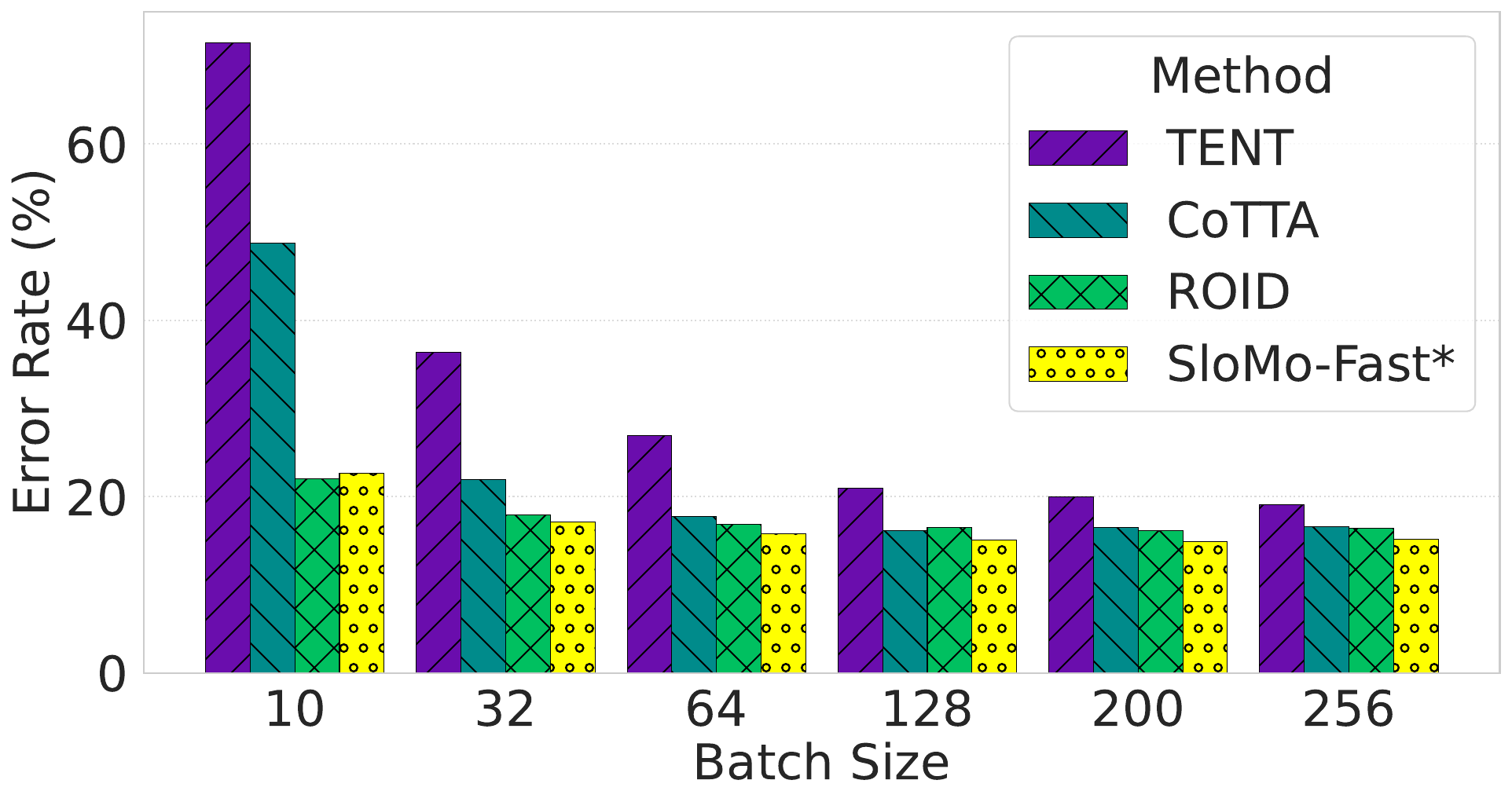}
    \caption{Impact of batch size on error rates (\%) in the CIFAR10C online CTTA setup across different methods. Highlights that increasing batch size improves classification performance, with \pa* outperforming all other methods at each batch size.}
    \label{fig:ablation-batch-size}
\end{figure}

\begin{figure}[ht]
    \centering

    \begin{minipage}[t]{0.45\linewidth}
        \centering
        \includegraphics[width=0.95\linewidth]{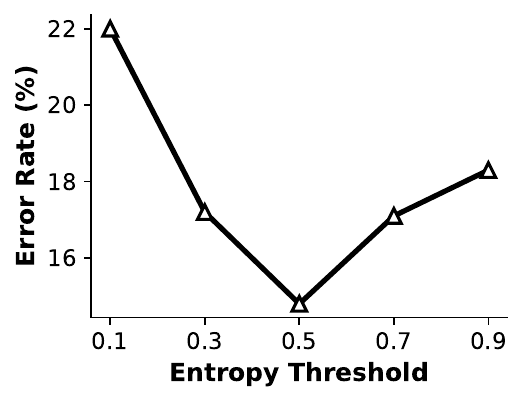}
    \end{minipage}%
    \begin{minipage}[t]{0.45\linewidth}
        \centering
        \includegraphics[width=0.95\linewidth]{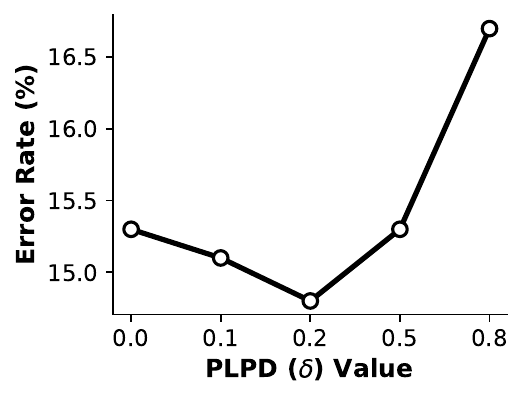}
    \end{minipage}
    \\[1ex]

    \begin{minipage}[t]{0.45\linewidth}
        \centering
        \includegraphics[width=0.95\linewidth]{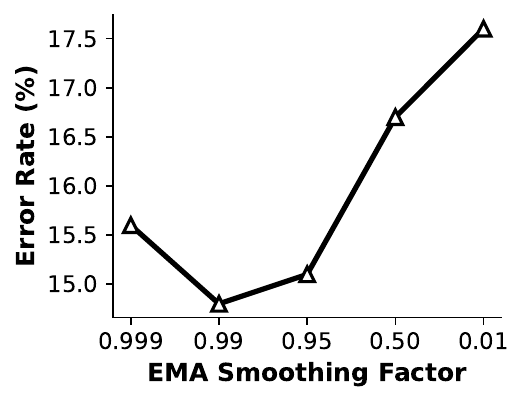}
    \end{minipage}%
    \begin{minipage}[t]{0.45\linewidth}
        \centering
        \includegraphics[width=0.95\linewidth]{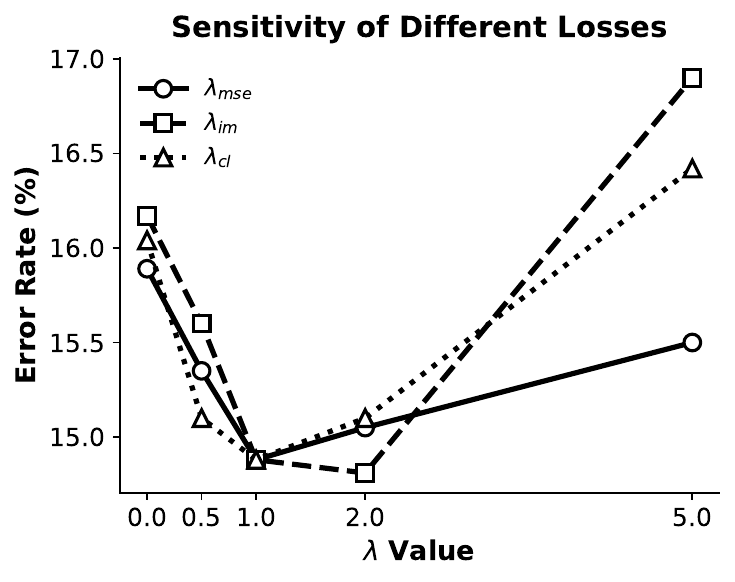}
    \end{minipage}

    \caption{Comparison of loss and accuracy against EMA factor and entropy.}
    \label{fig:hyp_loss}
\end{figure}

\subsection{Threshold, Efficiency, and Qualitative Analysis}
\label{sec:4.6}
\textbf{Threshold and Hyperparameter Selection.} We use optimal hyperparameters: EMA smoothing factor = 0.01, loss weights $\lambda_{cl}=\lambda_{im}=\lambda_{mse}=1$, entropy threshold = 0.5, and PLPD sensitivity threshold = 0.2. The sensitivity of them is shown in Figure~\ref{fig:hyp_loss} where increasing the entropy threshold up to 0.5 improves prototype reliability by including more trustworthy samples, while higher values introduce noisy features. Similarly, the PLPD threshold ensures that only prediction-stable samples contribute to prototype construction, preventing overconfident but unstable features from degrading adaptation.

\begin{table}[t]
\centering
\setlength{\tabcolsep}{0.9em}
\resizebox{0.65\linewidth}{!}{%
\begin{tabular}{lccccc}
\toprule
\multicolumn{6}{c}{\textbf{Parameter Efficiency and Adaptation Cost}} \\
\midrule
\textbf{Method} & \textbf{TP(\%)} & \textbf{CPU} & \textbf{GPU} & \textbf{Time} & \textbf{Err} \\
\midrule
Tent         & 4.9  & 2.5 & 10.4 & 0.01 & 55.8 \\
CoTTA        & 100  & 4.0 & 15.9 & 1.85 & 42.1 \\
ROID         & 4.9  & 3.9 & 11.5 & 0.30 & 36.3 \\

SloMo-Fast   & 4.9  & 6.3 & 12.2 & 0.24 & 35.9 \\
SloMo-Fast*  & 51   & 6.3 & 16.2 & 0.28 & 34.7 \\
\bottomrule
\end{tabular}}
\caption{
Comparison of existing representative methods by following:  
TP(\%) = percentage of trainable parameters;  
CPU = maximum allocated CPU memory (GB);  
GPU = maximum allocated GPU memory (GB);  
Time = average execution time during adaptation (seconds);
Err = classification error rate (\%). 
}
\label{tab:memory}
\end{table}

\noindent
\textbf{Parameter Efficiency and Adaptation Cost.} Table~\ref{tab:memory} shows, SloMo-Fast (BN-only) uses only 4.9\% trainable parameters i.e. comparable to ROID while maintaining similar  efficiency. SloMo-Fast* updates 51\% of parameters yet remains fast (0.28s per batch), outperforms ROID (0.30s), whose sampling, augmentation steps add extra overhead.

\begin{figure}[!t]
    \centering
    \includegraphics[width=0.75\linewidth]{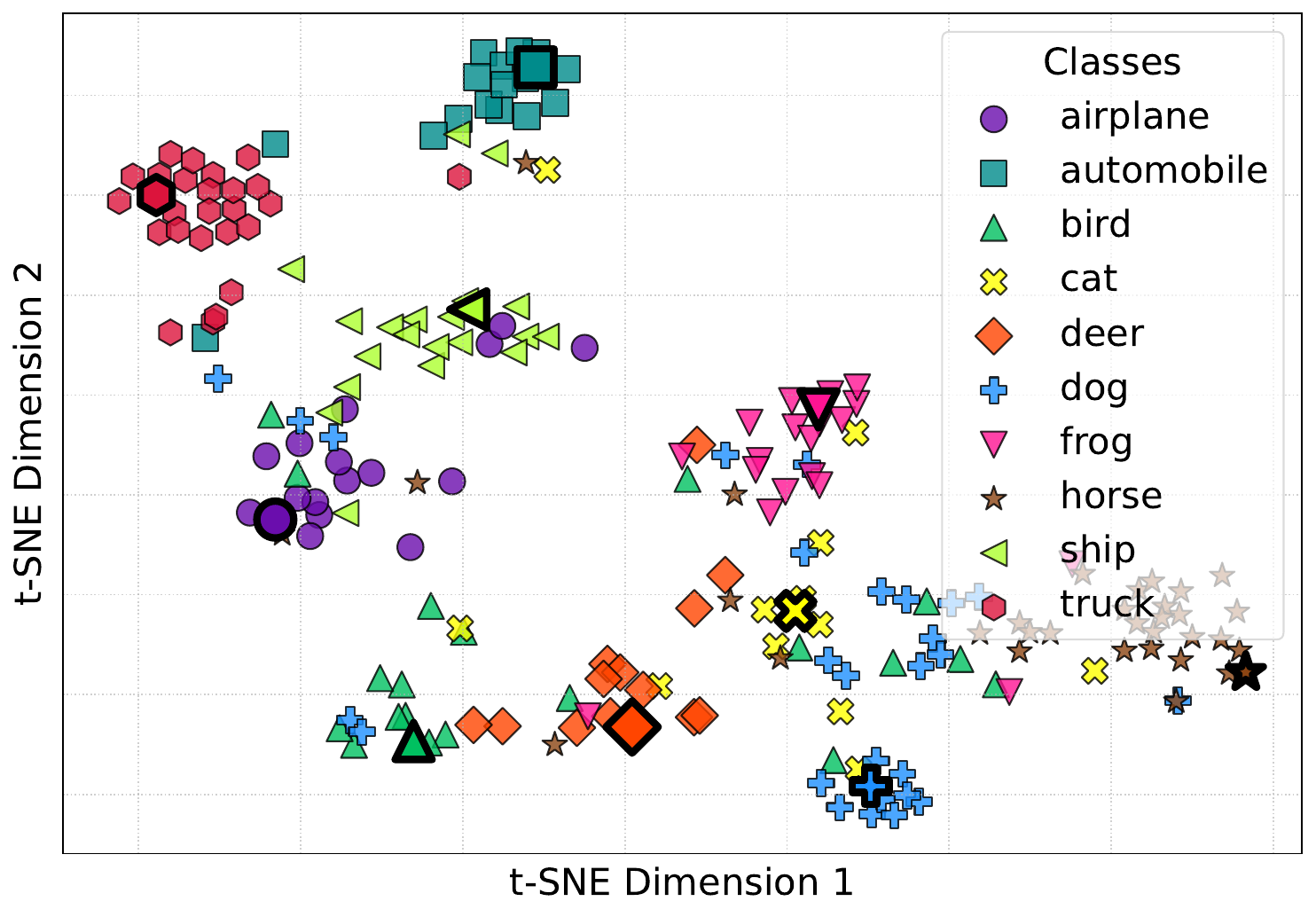}
    \caption{t-SNE visualization of feature representations and class prototypes (bigger shapes with thick black border). The visualization highlights distinct class separation, the model's ability to effectively learn discriminative feature representations.}
    \label{fig:tsne}
\end{figure}

\noindent
\textbf{Qualitative Results.} The t-SNE visualization (Figure~\ref{fig:tsne}) demonstrates that \textbf{SloMo-Fast} produces compact, well-separated class clusters, illustrating strong discriminative and robust representations even under severe corruption.

\section{Conclusion}
\label{sec:conclusion}
We present SloMo-Fast, a dual-teacher framework for CTTA that eliminates the need for source data while significantly enhancing adaptability, generalization, and efficiency. The Fast-Teacher ($T_1$) enables rapid on-the-fly adaptation to shifting domains, while the Slow-Teacher ($T_2$) ensures stable and robust generalization through gradual updates. These class prototypes are dynamically generated during inference, requiring no access to source data. To better reflect real-world deployment, we introduce Cyclic-TTA, a new CTTA setting where domain shifts may repeat over time. Extensive experiments across 11 challenging scenarios on 5 dataset including CIFAR-10C, CIFAR-100C, and ImageNet-C demonstrate that SloMo-Fast consistently outperforms SOTA methods that indicates robustness, generalization, and practicality in privacy-sensitive.

{
    \small
    \bibliographystyle{ieeenat_fullname}
    \bibliography{main}

@String(CVPR= {IEEE Conf. Comput. Vis. Pattern Recog.})

@String(ECCV= {Eur. Conf. Comput. Vis.})

@String(NIPS= {Adv. Neural Inform. Process. Syst.})

@String(ICLR = {Int. Conf. Learn. Represent.})

@String(AAAI = {AAAI})

@String(CVPR  = {CVPR})

@String(ECCV  = {ECCV})

@String(NIPS  = {NeurIPS})

@String(ICLR  = {ICLR})

@inproceedings{marsden2024universal,
  title={Universal test-time adaptation through weight ensembling, diversity weighting, and prior correction},
  author={Marsden, Robert A and D{\"o}bler, Mario and Yang, Bin},
  booktitle={Proceedings of the IEEE/CVF Winter Conference on Applications of Computer Vision},
  pages={2555--2565},
  year={2024}
}

@inproceedings{wang_tent_2021,
	title = {Tent: Fully Test-Time Adaptation by Entropy Minimization},
	booktitle = ICLR,
	author = {Wang, Dequan and Shelhamer, Evan and Liu, Shaoteng and Olshausen, B. and Darrell, Trevor},
	year = 2021,
}

@article{karani2021test,
  title={Test-time adaptable neural networks for robust medical image segmentation},
  author={Karani, Neerav and Erdil, Ertunc and Chaitanya, Krishna and Konukoglu, Ender},
  journal={Medical Image Analysis},
  volume={68},
  pages={101907},
  year={2021},
  publisher={Elsevier}
}

@inproceedings{wang2024continual,
  title={Continual test-time domain adaptation via dynamic sample selection},
  author={Wang, Yanshuo and Hong, Jie and Cheraghian, Ali and Rahman, Shafin and Ahmedt-Aristizabal, David and Petersson, Lars and Harandi, Mehrtash},
  booktitle={Proceedings of the IEEE/CVF Winter Conference on Applications of Computer Vision},
  pages={1701--1710},
  year={2024}
}

@inproceedings{wang_cotta_2022,
        author = {Wang, Qin and Fink, Olga and Van Gool, Luc and Dai, Dengxin},
	title = {Continual Test-Time Domain Adaptation},
	booktitle = CVPR,
        pages = {7191-7201},
	year = 2022
}

@inproceedings{niu_eata_2022,
	title = {Efficient Test-Time Model Adaptation without Forgetting},
	pages = {16888--16905},
        volume = {162},   
	booktitle = {Proceedings of the 39th International Conference on Machine Learning (ICML)},
	author = {Niu, Shuaicheng and Wu, Jiaxiang and Zhang, Yifan and Chen, Yaofo and Zheng, Shijian and Zhao, Peilin and Tan, Mingkui},
        series = 	 {Proceedings of Machine Learning Research},
	year = 2022,
        month = 	 {17--23 Jul},
        publisher =    {PMLR},
}

@inproceedings{zhang_memo_2022,
	title = {{MEMO}: test time robustness via adaptation and augmentation},
	pages = {38629--38642},
	booktitle = NIPS,
	author = {Zhang, Marvin and Levine, Sergey and Finn, Chelsea},
	year = 2022,
}

@inproceedings{song_ecotta_2023,
	title = {{EcoTTA}: Memory-Efficient Continual Test-Time Adaptation via Self-Distilled Regularization},
	pages = {11920--11929},
	booktitle = CVPR,
	author = {Song, Junha and Lee, Jungsoo and Kweon, In So and Choi, Sungha},
	year = 2023,
}

@inproceedings{dobler_rmt_2023,
	location = {Vancouver, {BC}, Canada},
	title = {Robust Mean Teacher for Continual and Gradual Test-Time Adaptation},
	booktitle = CVPR,
        pages={7704-7714},
	author = {Döbler, Mario and Marsden, Robert A. and Yang, Bin},
	year = 2023,
}

@inproceedings{yuan_rotta_2023,
	title = {Robust Test-Time Adaptation in Dynamic Scenarios},
	pages = {15922--15932},
	booktitle = CVPR,
	author = {Yuan, Longhui and Xie, Binhui and Li, Shuang},
	year = 2023,
}

@inproceedings{lee_becotta_2024,
	title = {{BECoTTA}: Input-dependent Online Blending of Experts for Continual Test-time Adaptation},
	pages = {27072--27093},
	booktitle = ICML,
	author = {Lee, Daeun and Yoon, Jaehong and Hwang, Sung Ju},
	year = 2024,
}

@inproceedings{liu_vida_2024,
	title = {{ViDA}: Homeostatic Visual Domain Adapter for Continual Test Time Adaptation},
	author = {Liu, Jiaming and Yang, Senqiao and Jia, Peidong and Zhang, Renrui and Lu, Ming and Guo, Yandong and Xue, Wei and Zhang, Shanghang},
	booktitle = ICLR,
        year = 2024,
}

@article{chakrabarty2023santa,
  title={Santa: Source anchoring network and target alignment for continual test time adaptation},
  author={Chakrabarty, Goirik and Sreenivas, Manogna and Biswas, Soma},
  journal={Transactions on Machine Learning Research},
  year={2023}
}

@article{press2023rdumb,
  title={Rdumb: A simple approach that questions our progress in continual test-time adaptation},
  author={Press, Ori and Schneider, Steffen and K{\"u}mmerer, Matthias and Bethge, Matthias},
  journal={Advances in Neural Information Processing Systems},
  volume={36},
  pages={39915--39935},
  year={2023}
}

@inproceedings{zhu2024reshaping,
  title={Reshaping the online data buffering and organizing mechanism for continual test-time adaptation},
  author={Zhu, Zhilin and Hong, Xiaopeng and Ma, Zhiheng and Zhuang, Weijun and Ma, Yaohui and Dai, Yong and Wang, Yaowei},
  booktitle={European Conference on Computer Vision},
  pages={415--433},
  year={2024},
  organization={Springer}
}

@inproceedings{lee_cmf_2024,
	title = {Continual Momentum Filtering on Parameter Space for Online Test-time Adaptation},
	author = {Lee, Jae-Hong and Chang, Joon-Hyuk},
        booktitle = ICLR,
	year = 2024,
}

@inproceedings{marsden_gtta_2024,
	title = {Introducing Intermediate Domains for Effective Self-Training during Test-Time},
	pages = {1--10},
	booktitle = IJCNN,
	author = {Marsden, Robert A. and Döbler, Mario and Yang, Bin},
	year = 2024,
}

@inproceedings{lee_deyo_2024,
	title = {Entropy is not Enough for Test-Time Adaptation: From the Perspective of Disentangled Factors},
	author = {Lee, Jonghyun and Jung, Dahuin and Lee, Saehyung and Park, Junsung and Shin, Juhyeon and Hwang, Uiwon and Yoon, Sungroh},
        booktitle = ICLR,
	year = 2024,
}

@inproceedings{yu_dplot_2024,
	location = {Seattle, {WA}, {USA}},
	title = {Domain-Specific Block Selection and Paired-View Pseudo-Labeling for Online Test-Time Adaptation},
	pages = {22723--22732},
	booktitle = CVPR,
	author = {Yu, Yeonguk and Shin, Sungho and Back, Seunghyeok and Ko, Minhwan and Noh, Sangjun and Lee, Kyoobin},
	year = 2024,
}

@inproceedings{tian2024parameter,
  title={Parameter-selective continual test-time adaptation},
  author={Tian, Jiaxu and Lyu, Fan},
  booktitle={Proceedings of the Asian Conference on Computer Vision},
  pages={1384--1400},
  year={2024}
}

@inproceedings{hendrycks2021many,
  title={The many faces of robustness: A critical analysis of out-of-distribution generalization},
  author={Hendrycks, Dan and Basart, Steven and Mu, Norman and Kadavath, Saurav and Wang, Frank and Dorundo, Evan and Desai, Rahul and Zhu, Tyler and Parajuli, Samyak and Guo, Mike and others},
  booktitle={Proceedings of the IEEE/CVF international conference on computer vision},
  pages={8340--8349},
  year={2021}
}

@article{hendrycks2019benchmarking,
  title={Benchmarking neural network robustness to common corruptions and perturbations},
  author={Hendrycks, Dan and Dietterich, Thomas},
  journal={arXiv preprint arXiv:1903.12261},
  year={2019}
}

@article{wang2019learning,
  title={Learning robust global representations by penalizing local predictive power},
  author={Wang, Haohan and Ge, Songwei and Lipton, Zachary and Xing, Eric P},
  journal={Advances in Neural Information Processing Systems},
  volume={32},
  year={2019}
}

@inproceedings{niloy2024effective,
  title={Effective restoration of source knowledge in continual test time adaptation},
  author={Niloy, Fahim Faisal and Ahmed, Sk Miraj and Raychaudhuri, Dripta S and Oymak, Samet and Roy-Chowdhury, Amit K},
  booktitle={Proceedings of the IEEE/CVF Winter Conference on Applications of Computer Vision},
  pages={2091--2100},
  year={2024}
}

@inproceedings{sojka2023ar,
  title={Ar-tta: A simple method for real-world continual test-time adaptation},
  author={S{\'o}jka, Damian and Cygert, Sebastian and Twardowski, Bart{\l}omiej and Trzci{\'n}ski, Tomasz},
  booktitle={Proceedings of the IEEE/CVF International Conference on Computer Vision},
  pages={3491--3495},
  year={2023}
}

@inproceedings{liu2024continual,
  title={Continual-mae: Adaptive distribution masked autoencoders for continual test-time adaptation},
  author={Liu, Jiaming and Xu, Ran and Yang, Senqiao and Zhang, Renrui and Zhang, Qizhe and Chen, Zehui and Guo, Yandong and Zhang, Shanghang},
  booktitle={Proceedings of the IEEE/CVF Conference on Computer Vision and Pattern Recognition},
  pages={28653--28663},
  year={2024}
}

@inproceedings{yang2024navigating,
  title={Navigating continual test-time adaptation with symbiosis knowledge},
  author={Yang, Xu and Li, Moqi and Yin, Jie and Wei, Kun and Deng, Cheng},
  booktitle={Proceedings of the Thirty-Third International Joint Conference on Artificial Intelligence},
  pages={5326--5334},
  year={2024}
}

@inproceedings{yang2023exploring,
  title={Exploring safety supervision for continual test-time domain adaptation},
  author={Yang, Xu and Gu, Yanan and Wei, Kun and Deng, Cheng},
  booktitle={Proceedings of the Thirty-Second International Joint Conference on Artificial Intelligence},
  pages={1649--1657},
  year={2023}
}

@inproceedings{maharana2025palm,
  title={Palm: Pushing adaptive learning rate mechanisms for continual test-time adaptation},
  author={Maharana, Sarthak Kumar and Zhang, Baoming and Guo, Yunhui},
  booktitle={Proceedings of the AAAI Conference on Artificial Intelligence},
  volume={39},
  number={18},
  pages={19378--19386},
  year={2025}
}

@inproceedings{choiadaptive,
  title={Adaptive Energy Alignment for Accelerating Test-Time Adaptation},
  author={Choi, Wonjeong and Kim, Do-Yeon and Park, Jungwuk and Lee, Jungmoon and Park, Younghyun and Han, Dong-Jun and Moon, Jaekyun},
  booktitle={The Thirteenth International Conference on Learning Representations}
}

@inproceedings{deng2025learning,
  title={Learning to Generate Gradients for Test-Time Adaptation via Test-Time Training Layers},
  author={Deng, Qi and Niu, Shuaicheng and Zhang, Ronghao and Chen, Yaofo and Zeng, Runhao and Chen, Jian and Hu, Xiping},
  booktitle={Proceedings of the AAAI Conference on Artificial Intelligence},
  volume={39},
  number={15},
  pages={16235--16243},
  year={2025}
}

@inproceedings{ni2025maintaining,
  title={Maintaining consistent inter-class topology in continual test-time adaptation},
  author={Ni, Chenggong and Lyu, Fan and Tan, Jiayao and Hu, Fuyuan and Yao, Rui and Zhou, Tao},
  booktitle={Proceedings of the Computer Vision and Pattern Recognition Conference},
  pages={15319--15328},
  year={2025}
}

@inproceedings{yuan2024tea,
  title={Tea: Test-time energy adaptation},
  author={Yuan, Yige and Xu, Bingbing and Hou, Liang and Sun, Fei and Shen, Huawei and Cheng, Xueqi},
  booktitle={Proceedings of the IEEE/CVF Conference on Computer Vision and Pattern Recognition},
  pages={23901--23911},
  year={2024}
}

@inproceedings{beery2018recognition,
  title={Recognition in terra incognita},
  author={Beery, Sara and Van Horn, Grant and Perona, Pietro},
  booktitle={Proceedings of the European conference on computer vision (ECCV)},
  pages={456--473},
  year={2018}
}

@article{niu2023towards,
  title={Towards stable test-time adaptation in dynamic wild world},
  author={Niu, Shuaicheng and Wu, Jiaxiang and Zhang, Yifan and Wen, Zhiquan and Chen, Yaofo and Zhao, Peilin and Tan, Mingkui},
  journal={arXiv preprint arXiv:2302.12400},
  year={2023}
}

@inproceedings{kundu2020universal,
  title={Universal source-free domain adaptation},
  author={Kundu, Jogendra Nath and Venkat, Naveen and Babu, R Venkatesh and others},
  booktitle={Proceedings of the IEEE/CVF conference on computer vision and pattern recognition},
  pages={4544--4553},
  year={2020}
}

@inproceedings{xu2024revisiting,
  title={Revisiting Source-Free Domain Adaptation: a New Perspective via Uncertainty Control},
  author={Xu, Gezheng and Guo, Hui and Yi, Li and Ling, Charles and Wang, Boyu and Yi, Grace},
  booktitle={The Thirteenth International Conference on Learning Representations},
  year={2024}
}

@inproceedings{schlachter2025memory,
  title={Memory-Efficient Pseudo-Labeling for Online Source-Free Universal Domain Adaptation using a Gaussian Mixture Model},
  author={Schlachter, Pascal and Wagner, Simon and Yang, Bin},
  booktitle={2025 IEEE/CVF Winter Conference on Applications of Computer Vision (WACV)},
  pages={6425--6434},
  year={2025},
  organization={IEEE}
}

@inproceedings{deng2025multi,
  title={Multi-granularity class prototype topology distillation for class-incremental source-free unsupervised domain adaptation},
  author={Deng, Peihua and Zhang, Jiehua and Sheng, Xichun and Yan, Chenggang and Sun, Yaoqi and Fu, Ying and Li, Liang},
  booktitle={Proceedings of the Computer Vision and Pattern Recognition Conference},
  pages={30566--30576},
  year={2025}
}

@inproceedings{mancini2018kitting,
  title={Kitting in the wild through online domain adaptation},
  author={Mancini, Massimiliano and Karaoguz, Hakan and Ricci, Elisa and Jensfelt, Patric and Caputo, Barbara},
  booktitle={2018 IEEE/RSJ International Conference on Intelligent Robots and Systems (IROS)},
  pages={1103--1109},
  year={2018},
  organization={IEEE}
}

@inproceedings{lee2024stationary,
  title={Stationary latent weight inference for unreliable observations from online test-time adaptation},
  author={Lee, Jae-Hong and Chang, Joon-Hyuk},
  booktitle={Forty-first International Conference on Machine Learning}
}

@inproceedings{chen2023improved,
  title={Improved test-time adaptation for domain generalization},
  author={Chen, Liang and Zhang, Yong and Song, Yibing and Shan, Ying and Liu, Lingqiao},
  booktitle={Proceedings of the IEEE/CVF Conference on Computer Vision and Pattern Recognition},
  pages={24172--24182},
  year={2023}
}

@article{iwasawa2021test,
  title={Test-time classifier adjustment module for model-agnostic domain generalization},
  author={Iwasawa, Yusuke and Matsuo, Yutaka},
  journal={Advances in Neural Information Processing Systems},
  volume={34},
  pages={2427--2440},
  year={2021}
}

@article{cho2024feature,
  title={Feature Augmentation Based Test-Time Adaptation},
  author={Cho, Younggeol and Kim, Youngrae and Yoon, Junho and Hong, Seunghoon and Lee, Dongman},
  booktitle={2025 IEEE/CVF Winter Conference on Applications of Computer Vision (WACV)},
  year={2024}
}

@article{han2025ranked,
  title={Ranked Entropy Minimization for Continual Test-Time Adaptation},
  author={Han, Jisu and Na, Jaemin and Hwang, Wonjun},
  journal={International conference on machine learning},
  year={2025}
}

@article{wu2025multi,
  title={Multi-Label Test-Time Adaptation with Bound Entropy Minimization},
  author={Wu, Xiangyu and Yu, Feng and Chen, Qing-Guo and Yang, Yang and Lu, Jianfeng},
  journal={International Conference on Learning Representations},
  year={2025}
}

@article{zhang2024dpcore,
  title={Dpcore: Dynamic prompt coreset for continual test-time adaptation},
  author={Zhang, Yunbei and Mehra, Akshay and Niu, Shuaicheng and Hamm, Jihun},
  journal={arXiv preprint arXiv:2406.10737},
  year={2024}
}

@article{hoang2024persistent,
  title={Persistent test-time adaptation in recurring testing scenarios},
  author={Hoang, Trung-Hieu and Vo, Duc M and others},
  journal={Advances in Neural Information Processing Systems},
  volume={37},
  pages={123402--123442},
  year={2024}
}

@article{vray2026reservoirtta,
  title={Reservoirtta: Prolonged test-time adaptation for evolving and recurring domains},
  author={Vray, Guillaume and Tomar, Devavrat and Gao, Xufeng and Thiran, Jean-Philippe and Shelhamer, Evan and Bozorgtabar, Behzad},
  journal={Advances in Neural Information Processing Systems},
  volume={38},
  pages={63357--63403},
  year={2026}
}

@article{lim2026and,
  title={When and where to reset matters for long-term test-time adaptation},
  author={Lim, Taejun and Hwang, Joong-Won and Lee, Kibok},
  journal={arXiv preprint arXiv:2603.03796},
  year={2026}
}

@article{duan2026lifelong,
  title={Lifelong test-time adaptation via online learning in tracked low-dimensional subspace},
  author={Duan, Dexin and Xu, Rui and Liu, Peilin and Wen, Fei},
  journal={Advances in Neural Information Processing Systems},
  volume={38},
  pages={19024--19053},
  year={2026}
}
}
\clearpage
\setcounter{page}{1}
\begin{center}

{\Huge \bfseries Supplementary Material \par}

\vspace{0.5em}

{\normalsize \bfseries SloMo-Fast: Slow-Momentum and Fast-Adaptive Teachers for Source-Free Continual Test-Time Adaptation \par}

\end{center}

\appendix
\renewcommand{\thefigure}{A.\arabic{figure}}
\setcounter{figure}{0}
\renewcommand{\thetable}{A.\arabic{table}}
\setcounter{table}{0}

\addcontentsline{toc}{section}{Appendix}














\section{Architecture, Dataset, and Benchmark Overview}
\subsection{Architecture Evolution}
\addcontentsline{toc}{subsection}{A.1\quad Architecture Evolution}
In this section, we have shown how our {\pa} evolves from prior research works and highlights the comparison of methodology with previous knowledge distillation based techniques of TTA.  CoTTA \cite{wang_cotta_2022} employs a teacher-student framework where the student model is updated based on the pseudo-labels generated by the teacher model. The teacher model, in turn, is updated using the Exponential Moving Average (EMA) of the student parameters. While CoTTA demonstrates effective adaptation, it suffers from catastrophic forgetting and lacks the ability to retain long-term domain knowledge.


To address this limitation, RMT \cite{dobler_rmt_2023} introduces source prototypes and utilizes contrastive loss between the source class prototypes and test-time inputs. However, relying on source prototypes is often impractical in real-world scenarios due to their rarity and unavailability in many applications. 

In contrast, our {\pa} framework introduces a second teacher model that is more domain-generalized. Instead of using source prototypes, {\pa} constructs class prototypes from confident test samples. This approach eliminates the dependence on source data while enabling long-term retention of domain knowledge, ensuring robust adaptation and generalization across dynamic and evolving domains.

\subsection{Dataset}
To rigorously evaluate the robustness and generalization capabilities of the proposed model, we utilize several benchmark datasets that represent various types of distribution shifts and corruptions.

\begin{itemize}
    \item \textbf{CIFAR10-C:} This dataset serves as a benchmark for model robustness against common real-world corruptions. It consists of the original CIFAR-10 test set subjected to 15 types of corruptions (e.g., noise, blur, weather, and digital effects) across five levels of severity. Each domain contains 10,000 images, spanning 10 distinct classes.
    
    \item \textbf{CIFAR100-C:} Similar to CIFAR10-C, this dataset expands the complexity to 100 classes, providing a more fine-grained evaluation of robustness. It includes 100 samples per category for each corruption domain, totaling 10,000 images per domain, and is used to assess how increased label complexity affects a model's stability under distribution shift.

    \item \textbf{ImageNet-C:} A large-scale robustness benchmark derived from the ImageNet validation set. It includes 15 corruption types categorized into noise, blur, weather, and digital categories. With 1,000 classes and 50 samples per class per domain, it provides a comprehensive 50,000-image evaluation suite that mimics unpredictable environmental conditions in high-resolution settings.

    \item \textbf{ImageNet-R (Rendition):} This dataset is designed to test a model's ability to generalize to diverse visual styles. It contains 30,000 images across 200 ImageNet classes, featuring various renditions such as paintings, embroidery, cartoons, graffiti, and plush toys. It highlights the texture bias of models by challenging them with non-naturalistic representations.

    \item \textbf{ImageNet-Sketch:} This dataset evaluates structural understanding by providing 50,889 black-and-white sketch drawings corresponding to all 1,000 ImageNet categories. By stripping away color and texture information, it benchmarks the model's capacity to recognize objects based solely on shape and skeletal geometry.
\end{itemize}
\subsection{Benchmarks for Test-Time Adaptation}
\label{benchmarks}
All evaluations are conducted in an \textit{online test-time adaptation (TTA)} setting, where predictions are updated and evaluated immediately. We evaluate our model on  benchmarks for analyzing CTTA:

\begin{itemize}
    \item\textbf{Continual Domains:} Following \cite{marsden2024universal}, the model adapts sequentially across $K$ domains $[D_1, D_2, \dots, D_K]$ without prior knowledge of domain boundaries. For the corruption datasets, the sequence includes all 15 corruption types encountered at severity level 5.

\item\textbf{Mixed Domains:} As in \cite{marsden2024universal}, test data from multiple domains are encountered together in a mixed manner during adaptation, with consecutive samples often coming from different domains.

\item\textbf{Gradual Domains:} Although some domain shifts happen abruptly, many progress gradually over time(severity of domain shifts changes incrementally), making this setting a practical scenario for test-time adaptation.

\item\textbf{Episodic Setting:} This setting considers a single domain shift, where upon encountering a new domain, the adaptation model resets to the source model and starts adaptation from the beginning.

\item\textbf{Continual-Cross Group:} Domains are encountered sequentially in a continual setup, where each domain is sampled one after another from different corruption groups (e.g., Noise, Blur, Weather, Digital, Distortion) like inter group mixing.

\item\textbf{Continual-Hard2Easy:} Domains are encountered sequentially, where corruptions are sorted from high error to low error based on the initial source model's performance at severity level 5.

\item\textbf{Continual-Easy2Hard:} Domains are encountered sequentially, where corruptions are sorted from low error to high error based on the initial source model's performance at severity level 5.

\item\textbf{Mixed after Continual TTA:} Domains are first encountered sequentially, as in the continual setting, followed by data from previously seen domains being encountered in a mixed manner.

\item\textbf{Continual after Mixed TTA:} Domains are first encountered in a mixed manner, where test data from multiple domains come together randomly. After this mixed phase, the domains are encountered sequentially, as in the continual setting.

\item \textbf{Continuously Changing Corruptions (CCC)}

To evaluate continual test-time adaptation in realistic long-sequence settings, 
we conduct experiments on the CCC benchmark \cite{press2023rdumb}, which features smooth 
and prolonged transitions across corruptions, thereby exposing the collapse of 
existing TTA methods over time.

\end{itemize}

\subsection{Cyclic Test-Time Adaptation Benchmark (Cyclic-TTA)}
\label{cyclic}
A new benchmark where the domain sequence is repeated in cycles based on corruption subgroups (e.g., Noise, Blur, Weather, Digital, and Distortion). Subgroups include corruptions such as noise (gaussian, shot, impulse), blur (defocus, motion, glass), weather (snow, fog, frost), digital (brightness, contrast), and distortion (elastic transform, pixelate, jpeg compression).

To model recurring distribution shifts, we introduce Cyclic-TTA, where domains reappear with a periodic but not necessarily fixed interval. Let the corruption groups  be $\mathcal{G}=\{G_1, G_2, \dots, G_K\}$, each containing multiple corruption types. A cyclic domain stream is generated by repeating these groups in order, forming $\mathcal{D} = [G_1, \dots, G_K,\, G_1, \dots, G_K, \dots]$. At each test-time step $t$, the model receives a sample $x_t \sim p(x \mid G_{c(t)})$, where the group index cycles as
\[
c(t) = \big((t-1) \bmod (K \cdot r)\big) \bmod K + 1,
\]
and $r$ denotes the \textit{cycle repetition rate}, allowing variable-length intervals before a domain reappears. Thus, domain recurrences satisfy $G_{c(t)} = G_{c(t + \Delta_t)}$ for some variable interval $\Delta_t$ determined by $r$ and the cycle structure. This formulation captures realistic scenarios where domain patterns repeat irregularly, enabling evaluation of methods that must rapidly recover performance upon re-encountering a previously seen domain while maintaining long-term domain-specific knowledge.

\section{Class-Specific Prototype Creation via Dual-Criterion Filtering}
\label{sec:prototype_creation}
To provide a robust foundation for contrastive learning and feature alignment, we implement a selective prototype generation mechanism. This approach ensures that only the most reliable and discriminative features from the target data are used to represent each class, preventing the noise inherent in test-time predictions from corrupting the model's feature space.

The selection process is governed by two complementary metrics: prediction entropy and Dynamic Label Calibration. While low entropy identifies samples where the model is confident, Dynamic Label Calibration (the difference between the highest and second-highest class probabilities) ensures that the prediction is not an ambiguous choice between two similar classes. By maintaining these features in a fixed-size priority queue, we ensure that the class-specific prototypes which calculated as the centroid of the queue, remain stable yet adaptive. Furthermore, the periodic pruning of the most stable elements prevents the prototypes from becoming biased toward a narrow subset of easy samples, encouraging a more comprehensive representation of the evolving target distribution.

To maintain reliable and discriminative prototypes, we selectively store test features in class-specific queues based on low entropy and high sensitivity (Dynamic Label Calibration). Periodic pruning ensures the queues retain only the most confident and stable representations for each class, preventing representation drift. The detailed selection process is outlined in Algorithm \ref{algo}.

\begin{algorithm}
\caption{Dual-Criterion Reliable Feature Selection}
\label{alg:priority_queue}
\begin{algorithmic}[1]
\State \textbf{Input: }test samples \( X_t \), teacher model \( T_1 \), entropy threshold \( \sigma \), sensitivity threshold \( \delta \), time interval \( p \), max queue size \( K \)
\State \textbf{Output: }Updated priority queues \( Q_c \) for each class \( c \)
\State Initialize priority queue \( Q_c \) for each class \( c \) with size \( K \)
\For{each test sample \( x_t \)}
    \State Predict class \( c \), entropy \( \mathcal{H}_t \), sensitivity \( \Delta p_t \), and feature \( z_t \) from \( T_1(x_t) \)
    \If{\( t \mod p = 0 \)} Remove element with \( \min \mathcal{H} \) from each \( Q_c \)
    \EndIf
    \If{\( \mathcal{H}_t \leq \sigma \) \textbf{and} \( \Delta p_t \geq \delta \)}
        \If{\( Q_c \) is full and \( \mathcal{H}_t < \max \mathcal{H} \) in \( Q_c \)} Replace element with \( \max \mathcal{H} \) by \( (z_t, \mathcal{H}_t) \)
        \ElsIf{\( Q_c \) is not full} Insert \( (z_t, \mathcal{H}_t) \) into \( Q_c \)
        \EndIf
    \EndIf
\EndFor
\end{algorithmic}
\label{algo}
\end{algorithm}

\section{Comprehensive Ablation Study:}
\subsection{Ablation Study on Losses Applied to T2}
\label{sup_ablation}
The results of the ablation study are summarized in Tables \ref{tab:t2_loss_effect_cifar10} and \ref{tab:t2_loss_effect_cifar100}, which evaluate the effect of different loss functions applied to the $T_2$ model on the CIFAR10-to-CIFAR10C and CIFAR100-to-CIFAR100C online continual test-time adaptation tasks, respectively, evaluations use the WideResNet-28 and ResNeXt-29 model under the highest corruption severity level (level 5). The classification error rates (\%) are reported for 15 corruption types, along with the mean error rate as a summary.

\begin{table*}
\centering
\caption{Evaluating the effect of our proposed loss on $T_2$, evaluated on the CIFAR10-to-CIFAR10C online continual test-time adaptation task. Results are reported as classification error rates (\%) using a WideResNet-28 model with corruption severity level 5. Mean squared error (MSE), information maximization (IM), and contrastive loss (CL).}
\resizebox{\textwidth}{!}{
\begin{tabular}{cccccccccccccccccccc}
\toprule
\multicolumn{3}{c}{Design Choices} & \multicolumn{15}{c}{Error Rate (\%)} \\ 
\cmidrule(lr){1-3} \cmidrule(lr){4-19}
MSE & IM & CL & Gaussian & Shot & Impulse & Defocus & Glass & Motion & Zoom & Snow & Frost & Fog & Brightness & Contrast & Elastic & Pixelate & JPEG & Mean \\ 
\midrule
\checkmark &   & \checkmark & 22.5 & 18.4 & 25.1 & 13.3 & 24.8 & 14.0 & 12.5 & 14.5 & 14.3 & 13.4 & 10.0 & 12.4 & 17.2 & 13.1 & 16.5 & 16.1 \\
\checkmark & \checkmark &  & 23.2 & 18.9 & 25.4 & 12.0 & 25.5 & 13.4 & 11.9 & 14.4 & 14.3 & 12.7 & 9.4 & 12.1 & 17.2 & 12.7 & 16.7 & 16.0 \\
 & \checkmark & \checkmark & 22.6 & 18.5 & 24.6 & 13.0 & 24.6 & 13.6 & 12.0 & 14.3 & 14.1 & 13.1 & 9.6 & 12.1 & 17.2 & 12.6 & 15.9 & 15.8 \\
\checkmark & \checkmark & \checkmark & 22.4 & 18.5 & 24.7 & 11.9 & 24.6 & 12.2 & 10.1 & 12.7 & 12.9 & 11.4 & 7.5 & 9.9 & 16.2 & 11.7 & 15.9 & 14.8 \\
\bottomrule
\end{tabular}}
\label{tab:t2_loss_effect_cifar10}
\end{table*}

In the CIFAR10-to-CIFAR10C task (Table \ref{tab:t2_loss_effect_cifar10}), the $T_2$ model trained with all three losses—mean squared error (MSE), information maximization (IM), and contrastive loss (CL)—achieves the lowest mean error rate of 14.88\%. This indicates the strong performance of the full configuration under severe corruption scenarios. Removing the contrastive loss (\(\checkmark\) MSE, \(\checkmark\) IM) slightly increases the mean error rate to 16.04\%, suggesting that CL contributes significantly to robustness. Excluding the information maximization loss (\(\checkmark\) MSE, \(\checkmark\) CL) results in a mean error rate of 16.17\%, highlighting the importance of IM in the adaptation process. When MSE is excluded (\(\checkmark\) IM, \(\checkmark\) CL), the mean error rate is slightly better at 15.89\%, reflecting a strong interaction between IM and CL, even in the absence of MSE.

\begin{table*}
\centering
\caption{Evaluating the effect of our proposed loss on $T_2$, evaluated on the CIFAR100-to-CIFAR100C online continual test-time adaptation task. Results are reported as classification error rates (\%) using a ResNeXt-29 model with corruption severity level 5. Mean squared error (MSE), information maximization (IM), and contrastive loss (CL).}

\resizebox{\textwidth}{!}{
\begin{tabular}{cccccccccccccccccccc}
\toprule
\multicolumn{3}{c}{Design Choices} & \multicolumn{15}{c}{Error Rate (\%)} \\ 
\cmidrule(lr){1-3} \cmidrule(lr){4-19}
MSE & IM & CL & Gaussian & Shot & Impulse & Defocus & Glass & Motion & Zoom & Snow & Frost & Fog & Brightness & Contrast & Elastic & Pixelate & JPEG & Mean\\ 
\midrule
\checkmark & \checkmark &  & 38.1 & 33.0 & 33.9 & 26.7 & 32.4 & 27.6 & 25.0 & 27.4 & 26.6 & 29.4 & 23.8 & 24.8 & 26.5 & 24.9 & 27.8 & 28.5 \\
\checkmark & & \checkmark & 38.9 & 33.2 & 33.4 & 26.6 & 32.0 & 27.3 & 24.8 & 26.7 & 27.0 & 28.3 & 23.6 & 24.2 & 26.7 & 24.7 & 27.2 & 28.3 \\
& \checkmark & \checkmark & 38.0 & 32.6 & 33.1 & 26.8 & 31.5 & 26.9 & 24.8 & 27.0 & 26.9 & 28.2 & 23.7 & 24.6 & 26.6 & 24.8 & 27.2 & 28.2 \\
\checkmark & \checkmark & \checkmark & 37.9 & 32.5 & 33.2 & 26.5 & 31.4 & 26.8 & 24.4 & 26.5 & 26.3 & 28.4 & 23.5 & 24.6 & 26.3 & 24.2 & 27.1 & 28.0 \\
 
\bottomrule
\end{tabular}}
\label{tab:t2_loss_effect_cifar100}
\end{table*}

For the CIFAR100-to-CIFAR100C task (Table \ref{tab:t2_loss_effect_cifar100}), similar trends are observed. The $T_2$ model trained with all three losses achieves the lowest mean error rate of 28.00\%. Removing CL (\(\checkmark\) MSE, \(\checkmark\) IM) increases the mean error rate to 28.57\%, demonstrating the importance of CL in enhancing robustness. Excluding IM (\(\checkmark\) MSE, \(\checkmark\) CL) leads to a mean error rate of 28.35\%, showing the critical role of IM in the adaptation process. Finally, removing MSE (\(\checkmark\) IM, \(\checkmark\) CL) results in a mean error rate of 28.23\%, again underscoring the synergy between IM and CL.

The results from both CIFAR10-to-CIFAR10C and CIFAR100-to-CIFAR100C tasks consistently highlight the benefits of integrating all three losses in the $T_2$ model. This combination achieves the lowest error rates across diverse corruption types, validating the effectiveness of the proposed design for continual test-time adaptation.

\subsection{Ablation Study on Dynamic Label Calibration (DLC)and Stochastic Restoration}
\noindent \textbf{Detailed Formulation of Dynamic Label Calibration}\noindent In test-time adaptation, the class proportions of the target domain often diverge from those of the source domain. This label shift degrades model reliability because the learned posterior $P_s(y|x)$ is biased toward source frequencies. We introduce Dynamic Label Calibration to correct this by re-weighting the source-trained posterior using the ratio of target-to-source priors:\begin{equation}P_t(y|x) = \frac{P_s(y|x) \frac{\pi_t(y)}{\pi_s(y)}}{\sum_{y'} P_s(y'|x) \frac{\pi_t(y')}{\pi_s(y')}}\end{equation}\noindent Given that the source domain prior $\pi_s(y)$ is near-uniform due to the information-loss constraints in our dual-teacher framework, the calibration focus shifts to accurately estimating the target domain prior $\pi_t(y)$.\noindent Since the target distribution is latent, we estimate it online using the model's own predictions. To prevent numerical instability or zero-probability assignments in small-batch ($B$) regimes, we introduce a smoothing factor $\alpha$:\begin{equation}\bar{\pi}t = \frac{\sum{i=1}^{B} \hat{y}_i + \alpha}{B + \alpha \cdot K}\end{equation}\noindent where $K$ is the total number of classes. This formulation acts as a regularizer, ensuring the estimated prior $\bar{\pi}_t$ remains robust to transient outliers in the test stream while allowing the model to favor classes that appear more frequently in the new environment.\\

\noindent \textbf{Results }\noindent The results of the ablation study are presented in Tables \ref{tab:pc_cifar10} and \ref{tab:pc_cifar100}, which evaluate the effect of Dynamic Label Calibration (DLC) applied to the model output and Stochastic Restoration (ST) of the $T_2$ model on the CIFAR10-to-CIFAR10C and CIFAR100-to-CIFAR100C online continual test-time adaptation tasks, respectively. The evaluations are conducted using the WideResNet-28 and ResNeXt-29 model under the highest corruption severity level (level 5). Classification error rates (\%) are reported for 15 corruption types, along with the mean error rate as an overall summary.

\begin{table*}
\centering
\caption{Classification error rate (\%) for the CIFAR10-to-CIFAR10C online continual test-time adaptation task. Results are evaluated using the WideResNet-28 model with corruption severity level 5. Dynamic Label Calibration (DLC) is applied to the model output, and Stochastic Restoration (ST) is applied to the $T_2$ model.}

\resizebox{\textwidth}{!}{
\begin{tabular}{ccccccccccccccccccc}
\toprule
\multicolumn{2}{c}{Design Choices} & \multicolumn{15}{c}{Error Rate (\%)} \\ 
\cmidrule(lr){1-2} \cmidrule(lr){3-18}
DLC & ST & Gaussian & Shot & Impulse & Defocus & Glass & Motion & Zoom & Snow & Frost & Fog & Brightness & Contrast & Elastic & Pixelate & JPEG & Mean \\ 
\midrule
\checkmark &   & 22.6 & 17.8 & 23.8 & 13.8 & 24.7 & 14.9 & 12.4 & 14.4 & 14.3 & 13.7 & 10.3 & 12.5 & 17.3 & 12.8 & 15.7 & 16.1 \\
 & \checkmark & 22.5 & 18.5 & 24.4 & 12.8 & 24.7 & 13.3 & 11.7 & 14.4 & 14.0 & 13.0 & 9.6 & 11.7 & 17.0 & 12.5 & 15.8 & 15.7 \\
\checkmark & \checkmark & 22.4 & 18.5 & 24.7 & 11.9 & 24.6 & 12.2 & 10.1 & 12.7 & 12.9 & 11.4 & 7.5 & 9.9 & 16.2 & 11.7 & 15.9 & 14.8 \\
\bottomrule
\end{tabular}}
\label{tab:pc_cifar10}
\end{table*}

\begin{table*}
\centering
\caption{Classification error rate (\%) for the CIFAR100-to-CIFAR100C online continual test-time adaptation task. Results are evaluated using the ResNeXt-29 model with corruption severity level 5. Dynamic Label Calibration (DLC) is applied to the model output, and Stochastic Restoration (ST) is applied to the $T_2$ model.}

\resizebox{\textwidth}{!}{
\begin{tabular}{ccccccccccccccccccc}
\toprule
\multicolumn{2}{c}{Design Choices} & \multicolumn{15}{c}{Error Rate (\%)} \\ 
\cmidrule(lr){1-2} \cmidrule(lr){3-18}
DLC & ST & Gaussian & Shot & Impulse & Defocus & Glass & Motion & Zoom & Snow & Frost & Fog & Brightness & Contrast & Elastic & Pixelate & JPEG & Mean \\ 
\midrule
\checkmark &  & 37.3 & 32.6 & 33.7 & 27.4 & 32.2 & 27.5 & 25.1 & 27.2 & 26.8 & 29.3 & 24.0 & 24.6 & 27.0 & 24.6 & 27.4 & 28.4 \\
 & \checkmark & 37.9 & 32.5 & 33.0 & 26.7 & 31.5 & 27.2 & 24.7 & 26.6 & 26.4 & 28.3 & 23.4 & 24.5 & 26.4 & 24.5 & 27.0 & 28.0 \\
\checkmark & \checkmark & 37.9 & 32.5 & 33.2 & 26.5 & 31.4 & 26.8 & 24.4 & 26.5 & 26.3 & 28.4 & 23.5 & 24.6 & 26.3 & 24.2 & 27.1 & 28.0 \\
\bottomrule
\end{tabular}}
\label{tab:pc_cifar100}
\end{table*}

In the CIFAR10-to-CIFAR10C task (Table \ref{tab:pc_cifar10}), applying DLC to the output and using Stochastic Restoration of the $T_2$ model achieves the lowest mean error rate of 14.88\%. This result demonstrates the effectiveness of combining these techniques for robust adaptation. When Stochastic Restoration is removed, and only DLC is applied to the output, the mean error rate increases to 16.11\%, indicating the critical role of Stochastic Restoration in enhancing the model’s robustness under severe corruptions. Conversely, removing DLC while retaining Stochastic Restoration results in a mean error rate of 15.78\%, suggesting that Dynamic Label Calibration (DLC)also significantly contributes to improved performance. These findings highlight the complementary roles of DLC and ST in enhancing the adaptation capabilities of the $T_2$ model.

In the CIFAR100-to-CIFAR100C task (Table \ref{tab:pc_cifar100}), a similar trend is observed. Applying DLC to the output alongside Stochastic Restoration of the $T_2$ model achieves the lowest mean error rate of 28.00\%. Removing Stochastic Restoration while retaining DLC increases the mean error rate to 28.48\%, demonstrating the importance of Stochastic Restoration for handling severe corruptions. On the other hand, using only Stochastic Restoration without DLC results in a mean error rate of 28.08\%, highlighting the significant role of Dynamic Label Calibration (DLC)in reducing classification errors.

The results from both CIFAR10-to-CIFAR10C and CIFAR100-to-CIFAR100C tasks consistently demonstrate that the combination of Dynamic Label Calibration (DLC)and Stochastic Restoration leads to the most effective adaptation.

\subsection{Ablation Study on Priority Queue Size}
\label{pq_size}
Table~\ref{tab:abl_queue_size} explores how varying the queue size affects classification error on CIFAR10-C and CIFAR100-C. For CIFAR10-C, performance remains stable, with the best result (14.88\%) at a queue size of 25. Larger sizes offer no clear gains. For CIFAR100-C, the lowest error (28.11\%) occurs at size 100, but differences are minor overall. These findings suggest that the model is relatively insensitive to queue size on both datasets.

\begin{table}
\centering
\resizebox{0.5\columnwidth}{!}{
    \begin{tabular}{c||ccccc}
        \toprule
        \diagbox{Dataset}{Queue\\Size} & 5 & 20 & 25 & 50 & 100 \\
        \midrule\midrule
        CIFAR10-C & 14.97 & 14.92 & 14.88 & 14.93 & 14.98 \\
        CIFAR100-C & 28.19 & 28.24 & 28.16 & 28.21 & 28.11 \\
        ImageNet-C & 54.52 & 53.14 & 56.32 & 59.88 & 63.72 \\
        \bottomrule
    \end{tabular}}
\caption{Ablation study of classification error rates (\%) for CIFAR10-C and CIFAR100-C online continual test-time adaptation tasks. This table examines the impact of different queue sizes on classification error rates.}
\label{tab:abl_queue_size}
\vspace{-10pt}
\end{table}

\subsection{Effect of Consistency Loss}

Tables \ref{tab:ct_cifar10} and \ref{tab:ct_cifar100} present the classification error rates (\%) for the CIFAR10-to-CIFAR10C and CIFAR100-to-CIFAR100C online continual test-time adaptation tasks, respectively. These results evaluate the effect of applying a consistency loss between the student model and teacher: $T_1$, $T_2$, and $T_1$ with data augmentation input($T_1 (aug)$). The evaluations are conducted using WideResNet-28 for CIFAR10C and ResNeXt-29 for CIFAR100C under the largest corruption severity level (level 5). Classification error rates are reported for 15 corruption types, along with the mean error rate as a summary.For CIFAR10-C, the best results are achieved by incorporating the consistency loss between the student predictions and the predictions from both $T_1$ and $T_2$. For CIFAR100-C, the best performance is obtained by using the consistency loss between the student predictions and the predictions from $T_2$ and $T_1$ with augmented samples.

\begin{table*}
\centering
\caption{Classification error rate (\%) for the standard CIFAR10-to-CIFAR10C online continual test-time adaptation task. Results are evaluated on WideResNet-28 with the largest corruption severity level 5. The consistency loss calculated between student and teachers. $T_1$ indicates consistency loss calculated between student and teacher 1, $T_2$ indicates consistency loss calculated between student and teacher 2, $T_1(aug) $ indicates consistency loss calculated between student and teacher 1 where the input of teacher is augementation of input images.   }
\resizebox{\textwidth}{!}{
\begin{tabular}{ccccccccccccccccccc}
\toprule
\multicolumn{3}{c}{Design Choices} & \multicolumn{15}{c}{Error Rate (\%)} \\ 
\cmidrule(lr){1-3} \cmidrule(lr){4-19}
T1 & $T_2$ & T1(aug) & Gaussian & Shot & Impulse & Defocus & Glass & Motion & Zoom & Snow & Frost & Fog & Brightness & Contrast & Elastic & Pixelate & JPEG & Mean \\ 
\midrule
\checkmark & \checkmark & & 22.7 & 18.1 & 24.2 & 12.8 & 25.5 & 13.5 & 11.5 & 15.0 & 14.2 & 13.3 & 9.7 & 12.2 & 17.0 & 13.3 & 15.7 & 15.9 \\
\checkmark & & \checkmark & 22.7 & 18.7 & 25.4 & 12.9 & 25.7 & 14.3 & 12.3 & 15.3 & 15.1 & 13.4 & 10.3 & 13.3 & 17.8 & 13.4 & 17.1 & 16.5 \\
 & \checkmark & \checkmark & 22.6 & 18.2 & 24.8 & 13.2 & 25.1 & 14.6 & 12.2 & 14.5 & 14.6 & 13.1 & 10.2 & 12.3 & 17.6 & 12.9 & 16.4 & 16.1 \\
\bottomrule
\end{tabular}}
\label{tab:ct_cifar10}
\end{table*}

\begin{table*}
\centering
\caption{Classification error rate (\%) for the standard CIFAR100-to-CIFAR100C online continual test-time adaptation task. Results are evaluated on ResNeXt-29 with the largest corruption severity level 5. The consistency loss calculated between student and teachers. $T_1$ indicates consistency loss calculated between student and teacher 1, $T_2$ indicates consistency loss calculated between student and teacher 2, $T_1(aug) $ indicates consistency loss calculated between student and teacher 1 where the input of teacher is augementation of input images. }
\resizebox{\textwidth}{!}{
\begin{tabular}{ccccccccccccccccccc}
\toprule
\multicolumn{3}{c}{Design Choices} & \multicolumn{15}{c}{Error Rate (\%)} \\ 
\cmidrule(lr){1-3} \cmidrule(lr){4-19}
T1 & $T_2$ & T1(aug) & Gaussian & Shot & Impulse & Defocus & Glass & Motion & Zoom & Snow & Frost & Fog & Brightness & Contrast & Elastic & Pixelate & JPEG & Mean \\ 
\midrule
\checkmark & \checkmark  &  & 38.2 & 32.9 & 33.9 & 26.3 & 32.0 & 27.0 & 24.7 & 27.3 & 26.6 & 28.9 & 23.7 & 24.0 & 26.5 & 24.3 & 27.1 & 28.2 \\
\checkmark &   & \checkmark & 38.1 & 33.1 & 33.9 & 27.1 & 32.5 & 27.4 & 25.4 & 27.7 & 27.0 & 29.0 & 24.2 & 25.6 & 27.7 & 25.6 & 28.6 & 28.9 \\
 & \checkmark & \checkmark & 37.3 & 32.7 & 33.0 & 26.3 & 31.6 & 27.2 & 24.7 & 26.9 & 26.3 & 28.3 & 23.6 & 24.7 & 26.7 & 24.6 & 27.1 & 28.1 \\
\bottomrule
\end{tabular}}
\label{tab:ct_cifar100}
\end{table*}

 \subsection{ Catastrophic Fogetting}
 \label{cf}
The figures illustrate the performance of different CTTA methods, including {\pa}, on the CIFAR10-C benchmark, highlighting challenges like catastrophic forgetting and the ability to retain long-term knowledge.

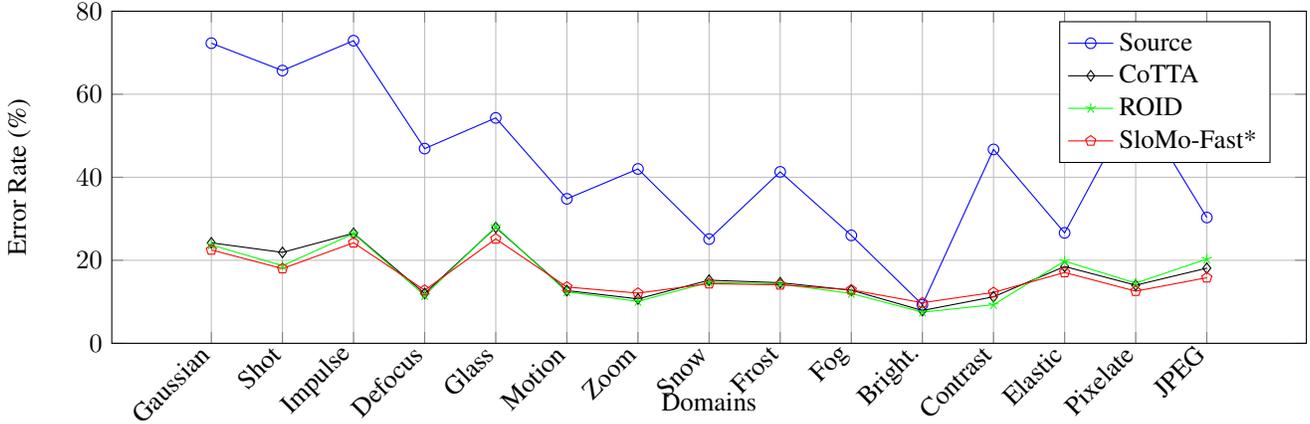
\begin{figure*}
\centering
\begin{tikzpicture}
    \begin{axis}[
        width=\textwidth,
        height=6cm,
        xlabel={Domains},
        ylabel={Error Rate (\%)},
        xtick={1,2,3,4,5,6,7,8,9,10,11,12,13,14,15},
        xticklabels={Gaussian, Shot, Impulse, Defocus, Glass, Motion, Zoom, Snow, Frost, Fog, Bright., Contrast, Elastic, Pixelate, JPEG},
        xticklabel style={rotate=45, anchor=east},
        legend style={at={(1.05,1)}, anchor=north west},
        grid=major,
        ymin=0, ymax=80,
        legend pos=north east,
        legend cell align={left}
    ]

    \addplot+[mark=o, color=blue] coordinates {
        (1,72.3) (2,65.7) (3,72.9) (4,46.9) (5,54.3) (6,34.8) 
        (7,42.0) (8,25.1) (9,41.3) (10,26.0) (11,9.3) (12,46.7)
        (13,26.6) (14,58.4) (15,30.3)
    };
    \addlegendentry{Source}

    \addplot+[mark=diamond, color=black] coordinates {
        (1,24.2) (2,21.9) (3,26.5) (4,12.0) (5,27.9) (6,12.7) 
        (7,10.7) (8,15.2) (9,14.6) (10,12.8) (11,7.9) (12,11.2)
        (13,18.5) (14,14.0) (15,18.1)
    };
    \addlegendentry{CoTTA}

    \addplot+[mark=star, color=green] coordinates {
        (1,23.7) (2,18.7) (3,26.4) (4,11.5) (5,28.1) (6,12.4) 
        (7,10.1) (8,14.7) (9,14.3) (10,12.0) (11,7.5) (12,9.3)
        (13,19.8) (14,14.5) (15,20.3)
    };
    \addlegendentry{ROID}

    \addplot+[mark=pentagon, color=red] coordinates {
        (1,22.49) (2,18.00) (3,24.22) (4,12.79) (5,25.15) (6,13.56) 
        (7,12.09) (8,14.37) (9,14.08) (10,12.87) (11,9.77) (12,12.25)
        (13,17.06) (14,12.56) (15,15.80)
    };
    \addlegendentry{\pa*}

    \end{axis}
    \end{tikzpicture}
    \caption{CTTA Error rates (\%) for Source (blue), CoTTA (black), ROID (green), and PA (red) across domains in the CIFAR10-C benchmark.}
    \label{fig:error_rates}
\end{figure*}

In the standard CTTA setting, as shown in Figure \ref{fig:error_rates}, the {\pa} method achieves consistently low error rates, with a mean error of \textbf{15.79\%}, outperforming CoTTA (\textbf{16.5\%}) and ROID (\textbf{16.2\%}). This demonstrates {\pa}'s superior adaptability while avoiding performance degradation seen in other methods.

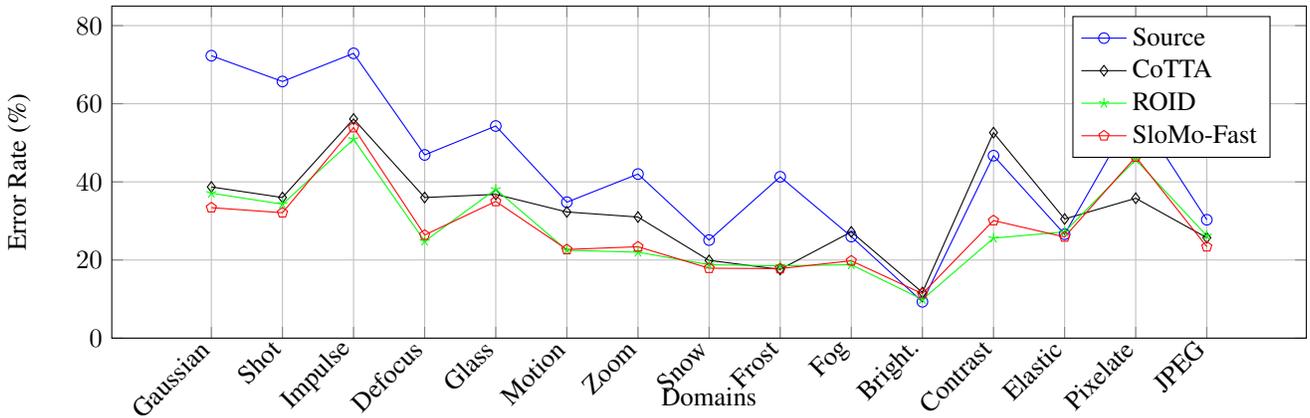
\begin{figure*}
\centering
\begin{tikzpicture}
    \begin{axis}[
        width=\textwidth,
        height=6cm,
        xlabel={Domains},
        ylabel={Error Rate (\%)},
        xtick={1,2,3,4,5,6,7,8,9,10,11,12,13,14,15},
        xticklabels={Gaussian, Shot, Impulse, Defocus, Glass, Motion, Zoom, Snow, Frost, Fog, Bright., Contrast, Elastic, Pixelate, JPEG},
        xticklabel style={rotate=45, anchor=east},
        legend style={at={(1.05,1)}, anchor=north west},
        grid=major,
        ymin=0, ymax=85,
        legend pos=north east,
        legend cell align={left}
    ]

    \addplot+[mark=o, color=blue] coordinates {
        (1,72.3) (2,65.7) (3,72.9) (4,46.9) (5,54.3) (6,34.8) 
        (7,42.0) (8,25.1) (9,41.3) (10,26.0) (11,9.3) (12,46.7)
        (13,26.6) (14,58.4) (15,30.3)
    };
    \addlegendentry{Source}

    \addplot+[mark=diamond, color=black] coordinates {
        (1,38.7) (2,36.0) (3,56.1) (4,36.0) (5,36.8) (6,32.3) 
        (7,31.0) (8,19.9) (9,17.6) (10,27.2) (11,11.7) (12,52.6)
        (13,30.5) (14,35.8) (15,25.7)
    };
    \addlegendentry{CoTTA}

    \addplot+[mark=star, color=green] coordinates {
        (1,37.1) (2,34.3) (3,50.9) (4,24.8) (5,38.1) (6,22.5) 
        (7,22.0) (8,18.8) (9,18.5) (10,18.8) (11,9.9) (12,25.6)
        (13,27.2) (14,45.7) (15,26.2)
    };
    \addlegendentry{ROID}

    \addplot+[mark=pentagon, color=red] coordinates {
        (1,33.4) (2,32.1) (3,53.9) (4,26.4) (5,35.0) (6,22.7) 
        (7,23.4) (8,17.9) (9,17.8) (10,19.8) (11,11.4) (12,30.1)
        (13,25.9) (14,46.4) (15,23.4)
    };
    \addlegendentry{{\pa}}

    \end{axis}
\end{tikzpicture}
\caption{Mixed TTA Error rates (\%) for Source (blue), CoTTA (black), ROID (green), and PA (red) methods across domains in the CIFAR10-C benchmark for mixed domains.}
\label{fig:mixed_domains_error_rates}
\end{figure*}

For mixed domain settings, as shown in Figure \ref{fig:mixed_domains_error_rates}, {\pa} maintains the best mean error rate of \textbf{28.0\%}, compared to CoTTA (\textbf{32.5\%}) and ROID (\textbf{28.0\%}). This highlights {\pa}'s ability to handle mixed corruption scenarios effectively.

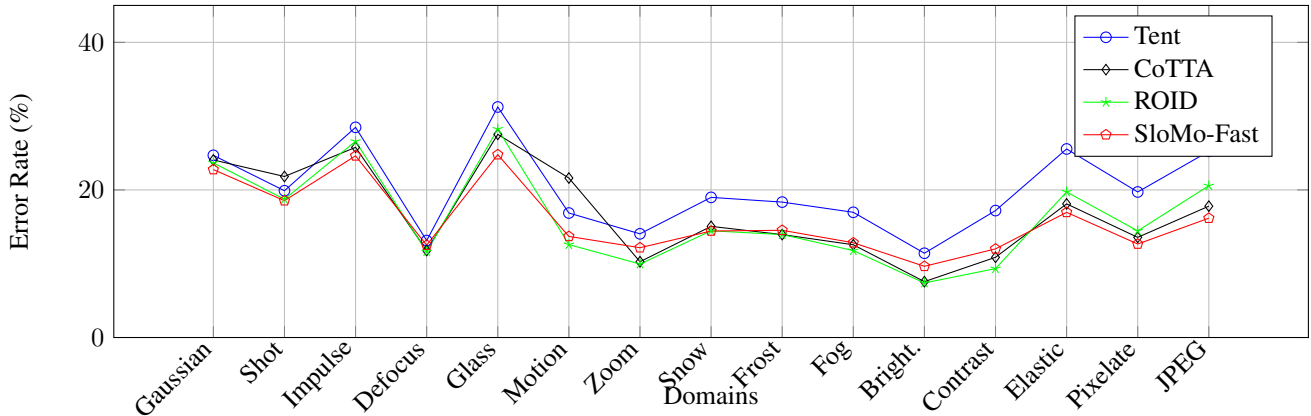
\begin{figure*}
\centering
\begin{tikzpicture}
    \begin{axis}[
        width=\textwidth,
        height=6cm,
        xlabel={Domains},
        ylabel={Error Rate (\%)},
        xtick={1,2,3,4,5,6,7,8,9,10,11,12,13,14,15},
        xticklabels={Gaussian, Shot, Impulse, Defocus, Glass, Motion, Zoom, Snow, Frost, Fog, Bright., Contrast, Elastic, Pixelate, JPEG},
        xticklabel style={rotate=45, anchor=east},
        legend style={at={(1.05,1)}, anchor=north west},
        grid=major,
        ymin=0, ymax=45,
        legend pos=north east,
        legend cell align={left}
    ]

    \addplot+[mark=o, color=blue] coordinates {
        (1,24.69) (2,19.89) (3,28.48) (4,13.14) (5,31.24) (6,16.87) 
        (7,14.05) (8,18.99) (9,18.35) (10,16.97) (11,11.42) (12,17.20)
        (13,25.55) (14,19.72) (15,25.23)
    };
    \addlegendentry{Tent}

    \addplot+[mark=diamond, color=black] coordinates {
        (1,24.06) (2,21.82) (3,25.77) (4,11.78) (5,27.53) (6,21.61) 
        (7,10.27) (8,15.06) (9,13.95) (10,12.57) (11,7.57) (12,10.87)
        (13,18.11) (14,13.57) (15,17.80)
    };
    \addlegendentry{CoTTA}

    \addplot+[mark=star, color=green] coordinates {
        (1,23.66) (2,18.74) (3,26.59) (4,11.63) (5,28.25) (6,12.59) 
        (7,9.96) (8,14.46) (9,13.95) (10,11.78) (11,7.39) (12,9.34)
        (13,19.73) (14,14.45) (15,20.57)
    };
    \addlegendentry{ROID}

    \addplot+[mark=pentagon, color=red] coordinates {
        (1,22.76) (2,18.54) (3,24.62) (4,12.51) (5,24.78) (6,13.70) 
        (7,12.17) (8,14.43) (9,14.53) (10,12.84) (11,9.65) (12,12.01)
        (13,16.96) (14,12.67) (15,16.17)
    };
    \addlegendentry{\pa}

    \end{axis}
\end{tikzpicture}
\caption{Mixed after Continual TTA Error rates (\%) for Tent (blue), CoTTA (black), ROID (green), and PA (red) methods across domains in the CIFAR10-C benchmark for mixed domains after continual learning.}
\label{fig:mixed_after_continual_error_rates}
\end{figure*}

When evaluating performance in a mixed-after-continual setting, as in Figure \ref{fig:mixed_after_continual_error_rates}, {\pa} achieves the lowest mean error rate of \textbf{21.34\%}, significantly outperforming ROID (\textbf{27.37\%}) and CoTTA (\textbf{26.76\%}), showcasing its resilience to catastrophic forgetting.

\begin{figure*}
\centering
\resizebox{\textwidth}{!}{
\begin{tikzpicture}
    \begin{axis}[
        width=\textwidth,
        height=6cm,
        xlabel={Domains},
        ylabel={Error Rate (\%)},
        xtick={1,2,3,4,5,6,7,8,9,10,11,12,13,14,15},
        xticklabels={gaussian, shot, impulse, defocus, glass, motion, zoom, snow, frost, fog, brightness, contrast, elastic, pixel, jpeg},
        xticklabel style={rotate=45, anchor=east},
        ymin=0, ymax=30,
        legend style={at={(1.05,1)}, anchor=north west},
        grid=major,
        legend pos=north east,
        legend cell align={left}
    ]

    \addplot+[mark=star, color=green, line width=2pt] coordinates {
        (1,20.62) (2,21.00) (3,24.87) (4,10.52) (5,28.20) (6,12.06) 
        (7,10.06) (8,14.01) (9,13.79) (10,11.92) (11,7.37) (12,9.17)
        (13,19.16) (14,14.51) (15,20.02)
    };
    \addlegendentry{ROID Cycle 2}

    \addplot+[mark=pentagon, color=black, line width=2pt] coordinates {
        (1,20.34) (2,21.12) (3,20.40) (4,14.34) (5,15.92) (6,16.27) 
        (7,13.90) (8,13.10) (9,13.23) (10,12.58) (11,9.17) (12,11.26)
        (13,15.64) (14,11.98) (15,14.61)
    };
    \addlegendentry{\pa{ }Cycle 2}

    \end{axis}
\end{tikzpicture}}
\caption{Cyclic TTA Error rates (\%) for ROID and PA methods across domains with subgroup boundaries (Cycle 2 only). Here, Existing best ROID is fluctuating and indicates catestrophic forgetting where {\pa} is stable}
\label{fig:cyclic_roid_pa_cifar10c}
\end{figure*}
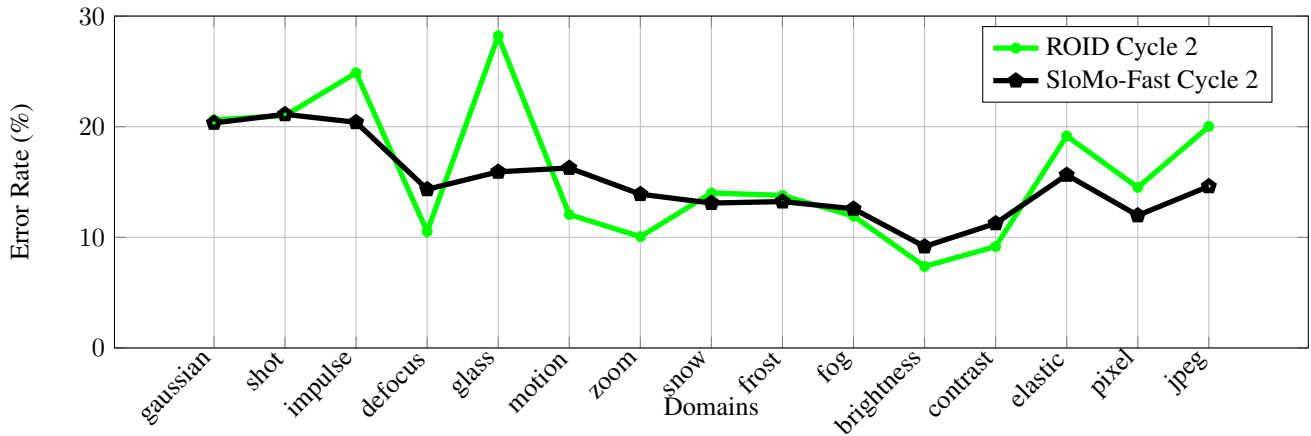

In the cyclic domain adaptation scenario, as shown in Figure \ref{fig:cyclic_roid_pa_cifar10c}, {\pa} exhibits stable performance, maintaining an average error rate of \textbf{14.63\%} across repeated domains, compared to ROID's \textbf{15.63\%}. This demonstrates {\pa}'s ability to retain previously learned knowledge without succumbing to forgetting, a common issue in ROID and CoTTA.

Overall, the results validate {\pa} as a robust solution for CTTA, capable of preserving long-term domain knowledge while achieving state-of-the-art performance.

\subsection{Forgetting Analysis}
\label{sec:forgetting}

\subsubsection{Forgetting Ratio}

The Forgetting Ratio ($F$) is a quantitative measure used to evaluate catastrophic forgetting in continual learning environments. It calculates the performance decline on previously encountered domains after the model has adapted to a new target domain. Mathematically, let $a_{i,j}$ denote the accuracy on domain $j$ after the model has completed adaptation to domain $i$ (where $i > j$). The forgetting for a specific domain $j$ at step $k$ is defined as the difference between the maximum accuracy achieved on that domain up to that point and its current accuracy: $f_{k,j} = \max_{l \in \{1, \dots, k-1\}} (a_{l,j} - a_{k,j})$. The overall Forgetting Ratio at sequence $k$ is then derived by averaging these individual drops across all preceding domains: $F_k = \frac{1}{k-1} \sum_{j=1}^{k-1} f_{k,j}$. A lower ratio indicates that the model is effectively preserving knowledge, while a higher ratio suggests that new adaptations are overwriting previously learned features.

In the context of Continual Test-Time Adaptation (CTTA), the forgetting ratio serves as a critical indicator of the stability-plasticity trade-off. As the model parameters shift to minimize error on an evolving target domain, they often drift away from the optimal weights required for the original source domain and earlier target domains. This drift is particularly problematic in source-free settings, where the absence of the original data makes it difficult to anchor the model's representations. High forgetting ratios often lead to error accumulation, where the loss of generalized features causes the model to rely on noisy pseudo-labels, eventually resulting in model collapse. To mitigate this, robust frameworks utilize stability mechanisms, such as a slow-moving teacher model, to maintain a low forgetting ratio. By preserving long-term knowledge, these systems ensure that the model can handle recurring shifts and maintain reliable performance across the entire deployment lifecycle.

To investigate how different test-time adaptation methods retain knowledge of the
\emph{initial} domain while adapting to a sequence of new domains, we track their
validation accuracy on the initial-domain validation set after each adaptation step.
Figure~\ref{fig:forgetting_main} presents the forgetting behavior across 15 consecutive
domain shifts (D1–D15).

As shown in the figure, the SloMo-Fast (T2) model demonstrates the slowest
forgetting, maintaining high validation accuracy on the original domain even after
multiple adaptations. This indicates that T2 effectively preserves long-term memory
of previously learned distributions. In contrast, Tent, CoTTA, Roid, and
SloMo-Fast (T1) exhibit progressively larger accuracy drops, reflecting a more
pronounced forgetting of the initial domain knowledge.

\subsection{Sensitivity Analysis of Teacher 2}
Each loss weight $\lambda$ governs the relative influence of a component in our adaptation framework:
{$\mathcal{L}_{MSE}$ stabilizes student updates through teacher consistency, 
$\mathcal{L}_{IM}$ encourages confident yet diverse predictions, 
and $\mathcal{L}_{CL}$ enforces prototype alignment for discriminative features. 
Theoretically, removing any of these components impairs performance, 
moderate weighting maintains a balance among their roles, 
and overly large weights risk dominating the adaptation process and disturbing this balance.
The trends in Figure ~\ref{fig:sensitivity_all} confirm this intuition across CIFAR10-C, CIFAR100-C, and ImageNet-C. Removing any component ($\lambda=0$) causes a sharp error increase, confirming the necessity of each term. 
At $\lambda=0.5$, error remains higher than baseline but better than full removal, showing weakened influence. 
The best performance is consistently achieved near $\lambda=1$, validating our balanced default setting. 
Increasing to $\lambda=2$ slightly hurts but remains close to baseline, 
whereas $\lambda=5$ leads to clear degradation: for instance, 
ImageNet-C error rises to $63.9\%$ under $\lambda_{MSE}=5$, reflecting over-regularization of the student. 
These observations match our theoretical expectation that balance among $\mathcal{L}_{CL}$, $\mathcal{L}_{MSE}$, and $\mathcal{L}_{IM}$ 
is essential, and extreme weighting disturbs the synergy among them.

\begin{figure*}[htbp]
    \centering
    
    \begin{subfigure}[b]{0.32\textwidth}
        \centering
        \includegraphics[width=\textwidth]{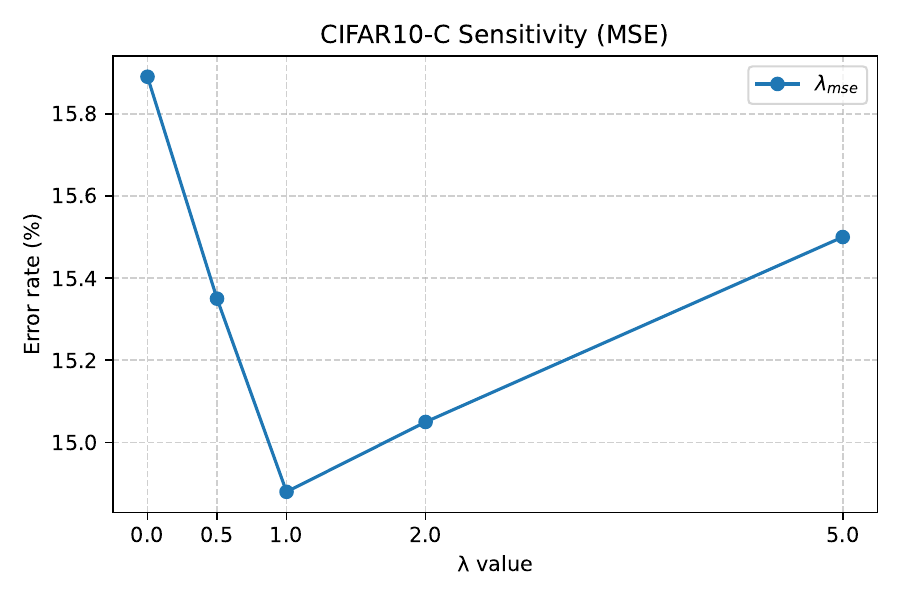}
        \caption{$\lambda_{mse}$}
        \label{fig:cifar10_mse}
    \end{subfigure}
    \hfill
    \begin{subfigure}[b]{0.32\textwidth}
        \centering
        \includegraphics[width=\textwidth]{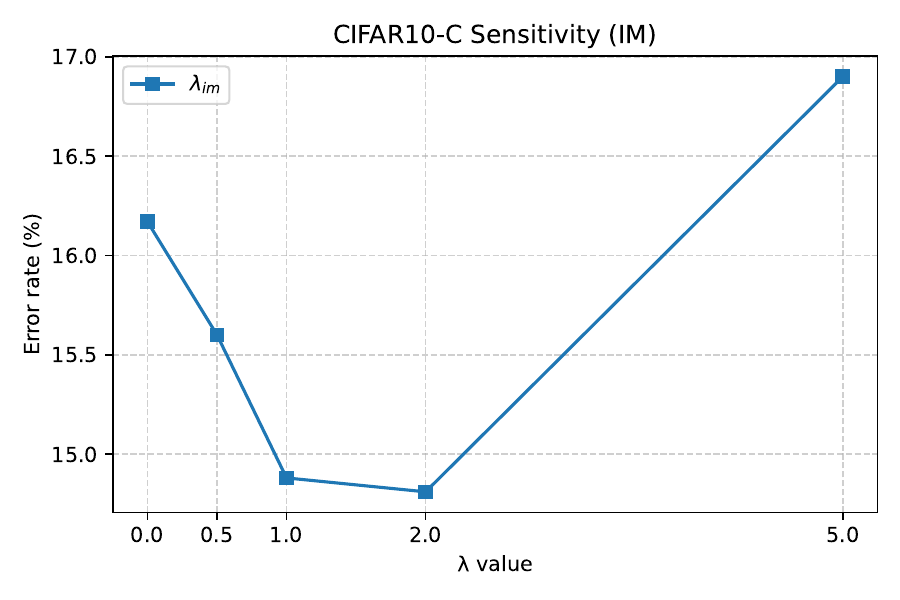}
        \caption{$\lambda_{im}$}
        \label{fig:cifar10_im}
    \end{subfigure}
    \hfill
    \begin{subfigure}[b]{0.32\textwidth}
        \centering
        \includegraphics[width=\textwidth]{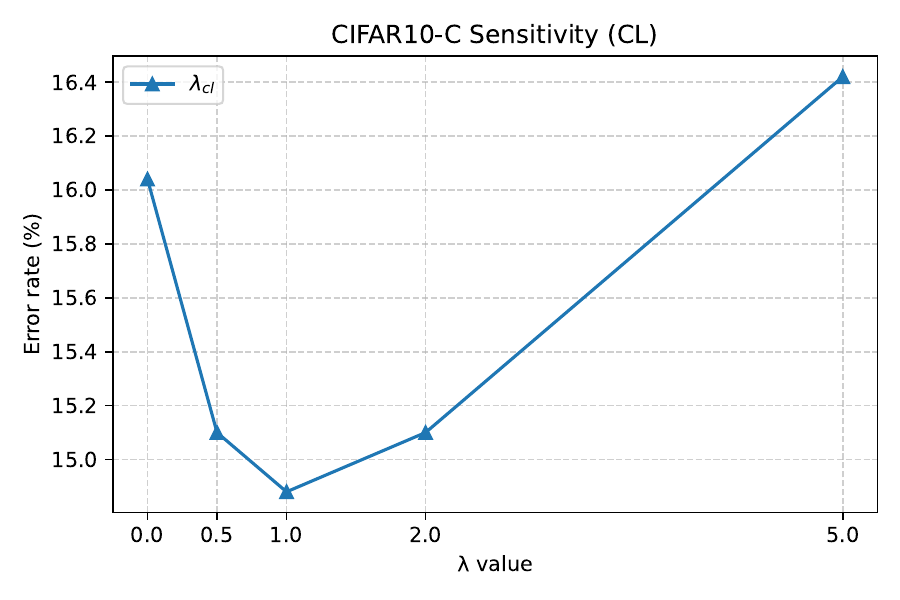}
        \caption{$\lambda_{cl}$}
        \label{fig:cifar10_cl}
    \end{subfigure}
    
    \begin{subfigure}[b]{0.32\textwidth}
        \centering
        \includegraphics[width=\textwidth]{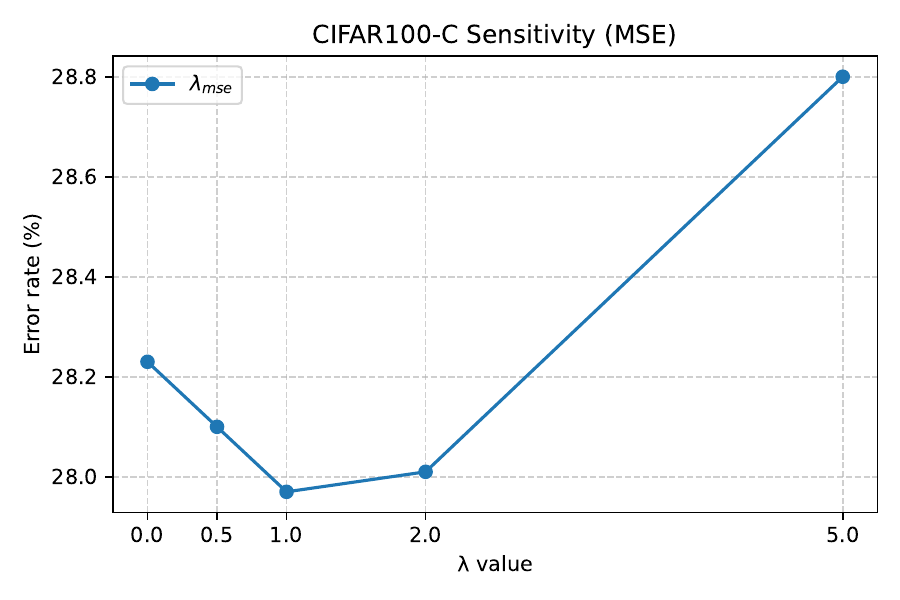}
        \caption{$\lambda_{mse}$}
        \label{fig:cifar100_mse}
    \end{subfigure}
    \hfill
    \begin{subfigure}[b]{0.32\textwidth}
        \centering
        \includegraphics[width=\textwidth]{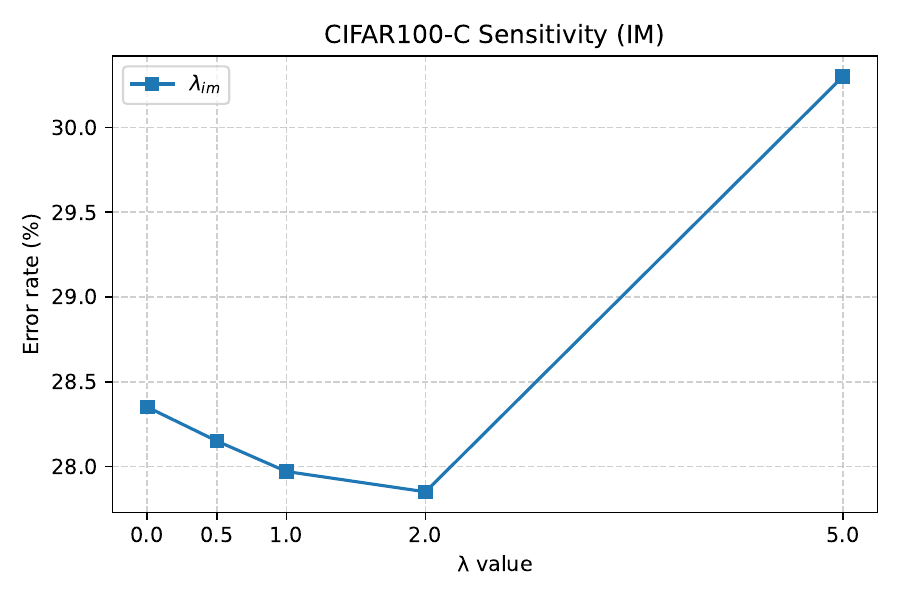}
        \caption{$\lambda_{im}$}
        \label{fig:cifar100_im}
    \end{subfigure}
    \hfill
    \begin{subfigure}[b]{0.32\textwidth}
        \centering
        \includegraphics[width=\textwidth]{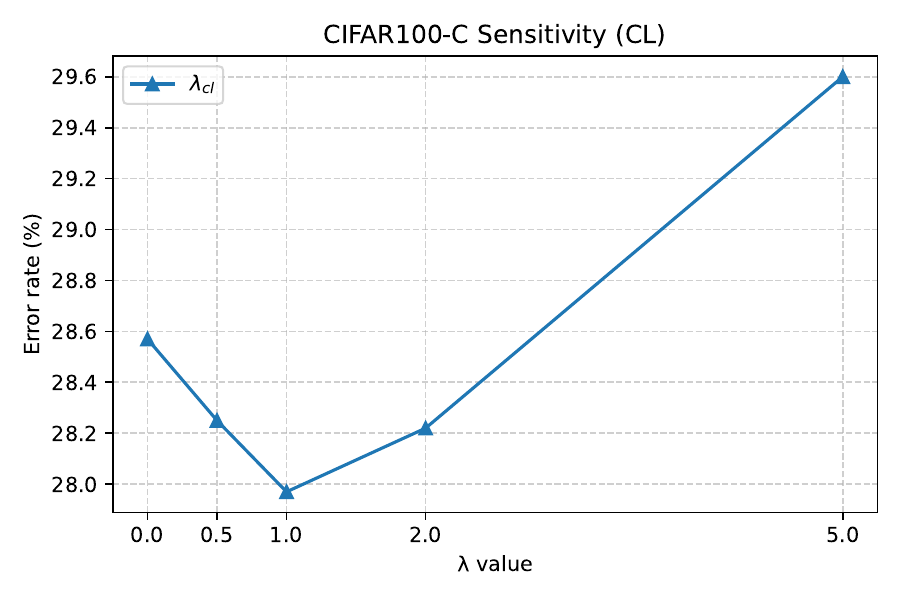}
        \caption{$\lambda_{cl}$}
        \label{fig:cifar100_cl}
    \end{subfigure}
    
    \begin{subfigure}[b]{0.32\textwidth}
        \centering
        \includegraphics[width=\textwidth]{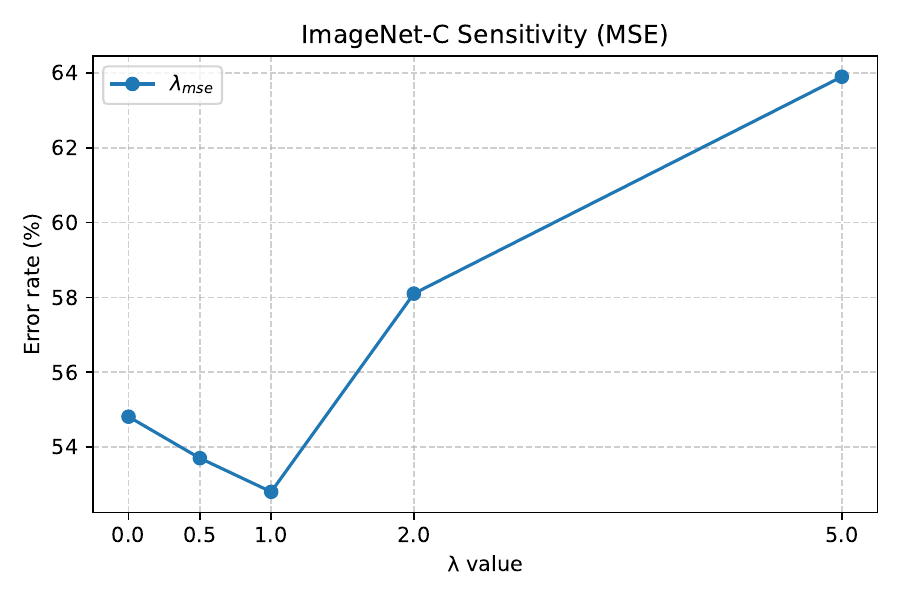}
        \caption{$\lambda_{mse}$}
        \label{fig:imagenet_mse}
    \end{subfigure}
    \hfill
    \begin{subfigure}[b]{0.32\textwidth}
        \centering
        \includegraphics[width=\textwidth]{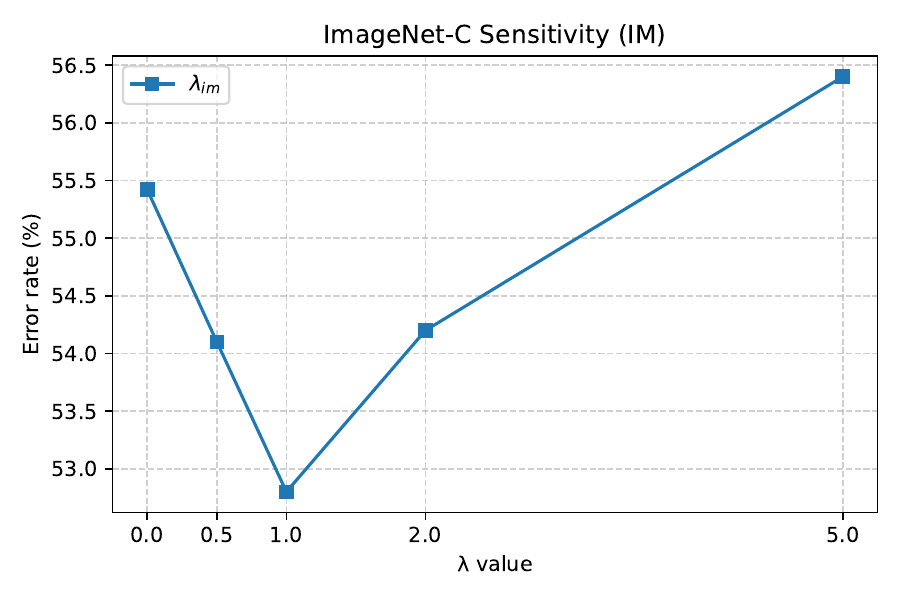}
        \caption{$\lambda_{im}$}
        \label{fig:imagenet_im}
    \end{subfigure}
    \hfill
    \begin{subfigure}[b]{0.32\textwidth}
        \centering
        \includegraphics[width=\textwidth]{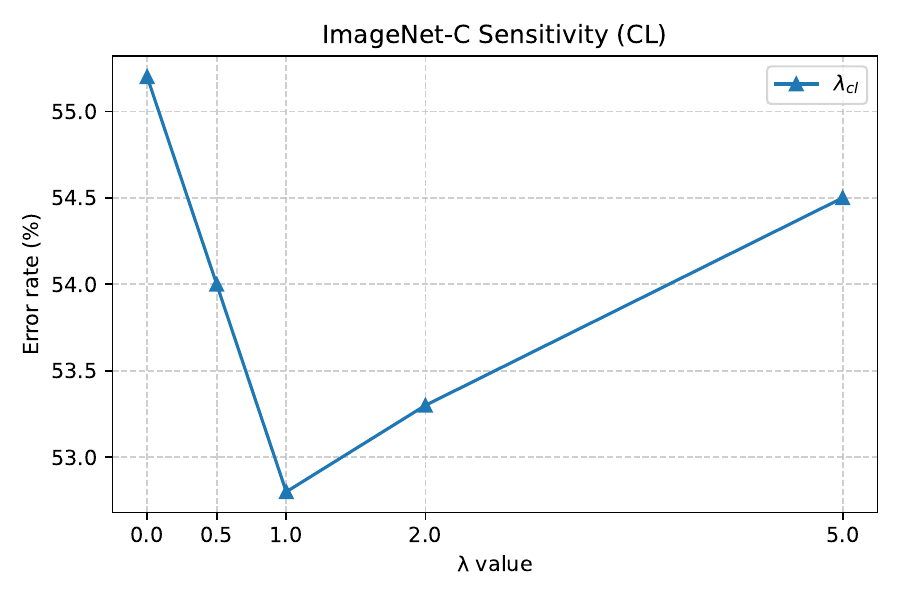}
        \caption{$\lambda_{cl}$}
        \label{fig:imagenet_cl}
    \end{subfigure}
    
    \caption{Sensitivity of error rates to different $\lambda$ values for corruption benchmarks. 
    Top row: CIFAR10-C, middle row: CIFAR100-C, bottom row: ImageNet-C. 
    Each column corresponds to the effect of $\lambda_{mse}$, $\lambda_{im}$, and $\lambda_{cl}$, respectively.}
    \label{fig:sensitivity_all}
\end{figure*}

\begin{figure}[t]
    \centering
    \includegraphics[width=0.95\linewidth]{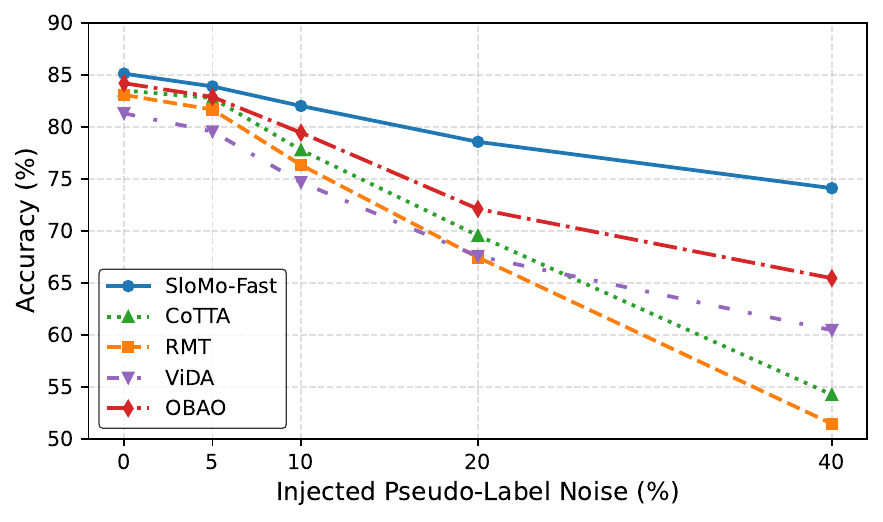}
    \caption{
        Accuracy under injected pseudo-label noise. SloMo-Fast degrades much more slowly 
        than competing TTA methods, confirming stronger resistance to noisy or incorrect 
        pseudo-label propagation.
    }
    \label{fig:noisy_pl_results}
\end{figure}

\subsection{Robustness to Noisy and Incorrect Pseudo-Labels}
\label{subsec:noisy_pl_experiment}
To address the weakness arising from the method’s reliance on Fast-Teacher (T1) pseudo-labels which can introduce confirmation bias and propagate errors into the feature queues, we conduct a robustness analysis by injecting controlled pseudo-label noise into the T1 outputs. Specifically, for a noise ratio $r \in \{0,5,10,20,40\}\%$, we randomly flip the pseudo-labels of incoming samples before they participate in queue updates or prototype formation. As shown in Fig.~\ref{fig:noisy_pl_results}, SloMo-Fast demonstrates substantially slower performance degradation compared to CoTTA, TENT, RMT, ViDA, and OBAO, indicating stronger resistance to pseudo-label noise. This stability arises from the PLPD-based reliability filtering and entropy-aware selection, which together suppress the impact of corrupted labels.  

\begin{figure*}
    \centering
    \includegraphics[width=0.85\linewidth]{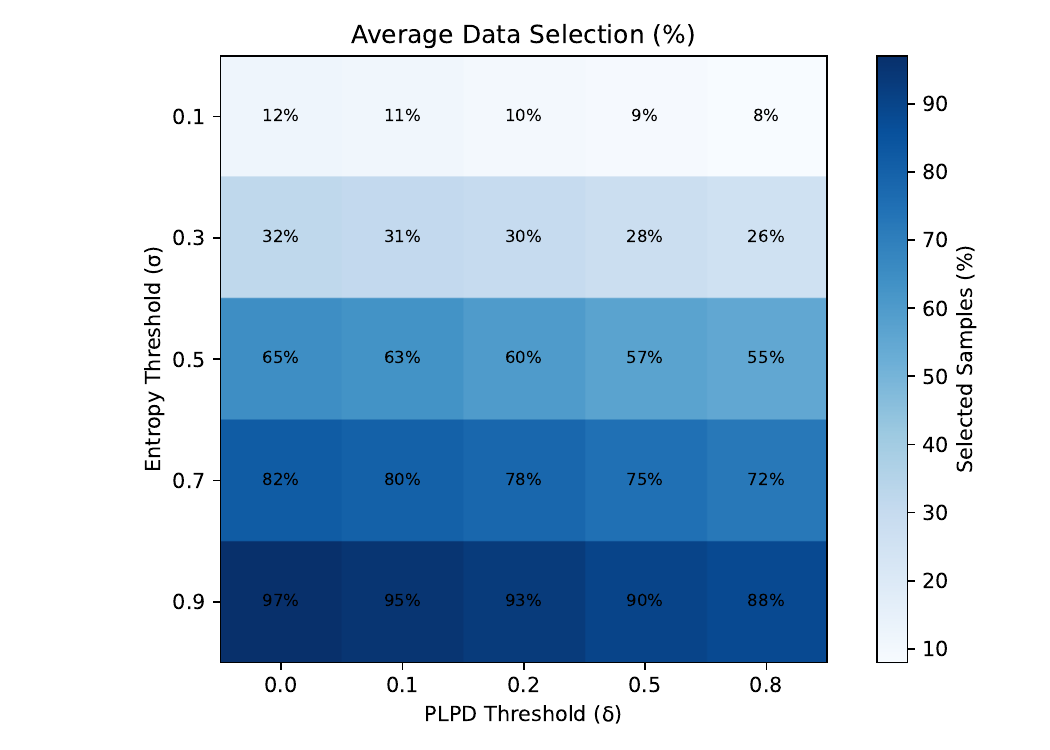}
    \caption{The data selection ratio for prototype generation from T1 with different Entropy and PLPD threshold values.}
    \label{fig:selection-matrix}
    \vspace{-10pt}
\end{figure*}

Fig.~\ref{fig:selection-matrix} illustrates the data selection ratio for prototype generation by evaluating the sensitivity of the entropy threshold ($\sigma$) and the Pseudo-Label Probability Difference (PLPD) threshold ($\delta$). The analysis demonstrates a direct correlation between threshold strictness and the volume of retained features, where decreasing the entropy limit or increasing the PLPD requirement leads to a more selective filtration process. This mechanism is critical for ensuring that the class-specific queues used to generate prototypes are populated only with high-confidence, stable representations, thereby effectively mitigating the risk of noisy features causing representation drift during continual adaptation. By visualizing these trade-offs, the figure highlights how the dual-criterion filtering balances the need for a sufficient sample size with the necessity of prediction reliability, ensuring the prototypes remain discriminative and stable anchors for the Slow-Teacher's alignment across evolving domains.

Overall, this experiment highlights that our improved mechanism effectively mitigates the pseudo-label confirmation-bias weakness of the original approach. By maintaining higher accuracy and more stable prototypes under increasing noise levels, SloMo-Fast proves better suited for long-term, real-world TTA scenarios where teacher predictions may be unreliable.

\subsection{Effect of Cycle Duration and Domain Reappearance Frequency in Cyclic-TTA}

Since cyclic domain shifts are common in real-world test-time adaptation (e.g., recurrent weather patterns, repeated scenes, or periodic operational environments), we further analyze how different cycle durations affect model stability. We vary the cycle length $L \in \{3, 5, 10, 15\}$, where each value controls how frequently previously-seen domains reappear in the test stream. Shorter cycles imply frequent domain recurrence, while longer cycles simulate long delays before returning to earlier distributions. As shown in Fig.~\ref{fig:cycle_duration_results}, SloMo-Fast maintains significantly higher stability across all cycle durations compared to CoTTA, ROID, TENT, and other baselines. When the cycle is short, SloMo-Fast quickly re-aligns to earlier domains due to its prototype memory and slow-teacher smoothing; when cycles are long, the method still shows lower forgetting and faster recovery upon reappearance.

\begin{figure}[t]
    \centering
    \includegraphics[width=0.95\linewidth]{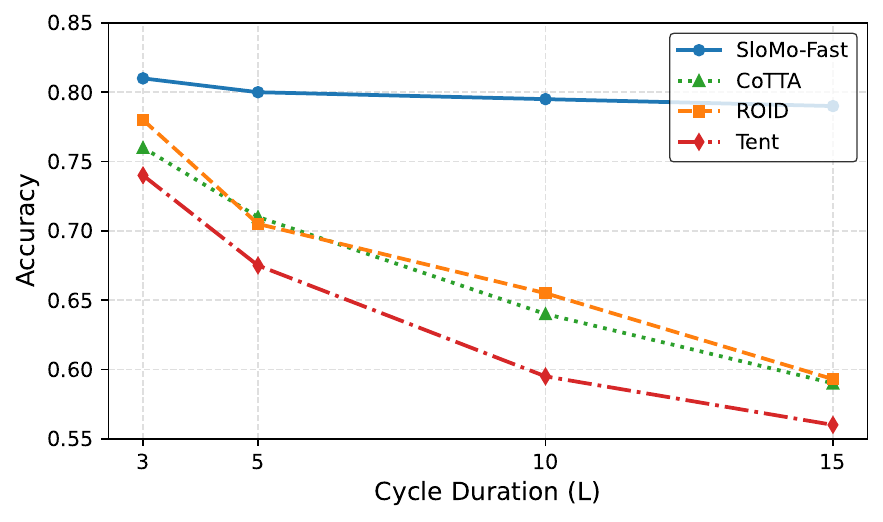}
    \caption{
        Accuracy across different cycle durations ($L = 3, 5, 10, 15$). 
        SloMo-Fast maintains stable performance for both short and long cycles 
        and recovers faster when domains reappear, demonstrating robustness to 
        varying domain-shift frequency.
    }
    \label{fig:cycle_duration_results}
\end{figure}

These results demonstrate that SloMo-Fast addresses the weakness of unstable long-term adaptation in cyclic settings. By leveraging prototype queues and a slow teacher for temporal consistency, the model adapts smoothly across repeated domains while minimizing catastrophic forgetting, even under long cycle durations. This makes the approach well-suited for real-world streams where domains fluctuate periodically or non-uniformly.

\begin{figure}[t]
    \centering
    \includegraphics[width=0.85\linewidth]{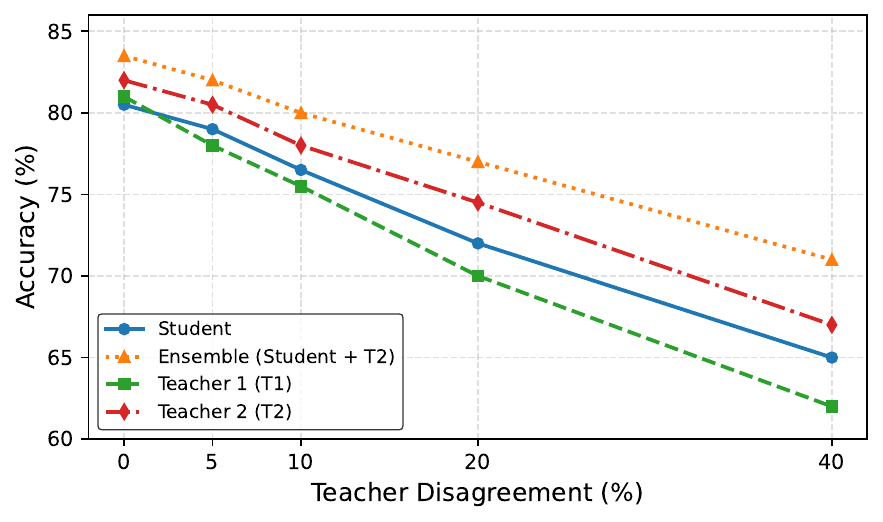}
    \caption{
        Accuracy under increasing disagreement between T1 and T2. 
        The ensemble demonstrates significantly stronger robustness than either 
        teacher individually, indicating that combining fast adaptation (student) 
        with slow temporal smoothing (T2) mitigates harmful effects of inconsistent 
        pseudo-label sources.
    }
    \label{fig:teacher_disagreement}
\end{figure}

\subsection{Robustness Under Teacher Disagreement}

To further analyze the stability of our method, we study how the system behaves when the two teachers: Fast-Teacher (T1) and Slow-Teacher (T2) produce conflicting predictions for the same input. Such disagreement commonly arises under sharp distribution shifts, degraded image quality, or teacher drift. We simulate this setting by injecting a controlled disagreement ratio $d \in \{0, 5, 10, 20, 40\}\%$, where for a fraction of samples, T1's prediction is replaced with an incorrect random class while T2 remains unchanged. We measure accuracy for the student model, the default ensemble (student + T2), and both teachers independently.

As shown in Fig.~\ref{fig:teacher_disagreement}, SloMo-Fast’s ensemble remains consistently more robust than either teacher alone as disagreement increases. While both T1 and T2 degrade with higher disagreement, the ensemble benefits from the temporal stability encoded by T2 and the immediate corrective feedback learned by the student. This complementary effect produces a slower accuracy drop and prevents the catastrophic failure seen in T1-heavy update strategies. These results highlight that explicitly modeling teacher disagreement is critical for mitigating confirmation bias, maintaining prototype integrity, and ensuring stable predictions under severe distribution inconsistencies.

\subsection{Efficiency Comparison of Adaptation Methods}
Table~\ref{tab:memory} evaluates the trade-offs between computational overhead and adaptation performance. While methods like CoTTA achieve adaptation through full-parameter updates (100\% TP), they incur a prohibitive execution time of 1.85s, making them less suitable for real-time applications. In contrast, the lightweight variant of SloMo-Fast updates only 4.9\% of parameters (primarily batch normalization layers) yet outperforms ROID in both classification accuracy (35.9\% vs. 36.3\% error) and inference speed (0.24s vs. 0.30s). This suggests that the dual-teacher architecture and dual-criterion filtering provide a more effective adaptation signal than traditional gradient-based updates on similar parameter budgets.

Further analysis reveals that SloMo-Fast* (51\% TP) establishes a superior Pareto frontier for high-performance CTTA. By expanding the trainable parameter set to include the student and teacher backbones, it achieves the lowest error rate of 34.7\% while remaining approximately 6.6 times faster than CoTTA. Although our framework requires slightly higher CPU memory to manage the class-specific prototype queues and Dynamic Label Calibration, the GPU overhead remains competitive. These results demonstrate that SloMo-Fast offers a highly scalable solution: the 4.9\% variant provides extreme efficiency for resource-constrained edge devices, while the 51\% variant offers state-of-the-art accuracy with minimal latency penalties.

\section{Supplementary Experimental Results}
\label{sec:details_result}

\subsection{Detailed result}

Across a wide range of datasets and model architectures including convolutional networks (WRN-28, ResNet-50), hierarchical transformers (Swin-b), and plain vision transformers (ViT-b-16), Slomo-Fast and its enhanced variant Slomo-Fast* demonstrate consistently superior performance in continual test-time adaptation. Unlike existing methods, whose effectiveness often varies significantly with the backbone or data distribution, our methods remain robust and stable across both lightweight and large-scale architectures. The improvements are especially pronounced on complex benchmarks like IN-C, IN-R, and IN-Sketch, confirming the generalizability of Slomo-Fast across both domain shifts and backbone types.

\begin{table*}
    \centering
    \begin{adjustbox}{max width=\textwidth}
    \begin{tabular}{l|ccccccccccccccc|c}
        \toprule
        \textbf{Method} & \rot{\textbf{Gaussian}} & \rot{\textbf{shot}} & \rot{\textbf{impulse}} & \rot{\textbf{defocus}} & \rot{\textbf{glass}} & \rot{\textbf{motion}} & \rot{\textbf{zoom}} & \rot{\textbf{snow}} & \rot{\textbf{frost}} & \rot{\textbf{fog}} & \rot{\textbf{bright.}} & \rot{\textbf{contrast}} & \rot{\textbf{elastic}} & \rot{\textbf{pixelate}} & \rot{\textbf{jpeg}} & \textbf{Mean} \\
        \midrule
        \multicolumn{17}{c}{\textbf{CIFAR10-C}} \\
        \midrule
        Source            & 72.3 & 65.7 & 72.9 & 46.9 & 54.3 & 34.8 & 42.0 & 25.1 & 41.3 & 26.0 & 9.3 & 46.7 & 26.6 & 58.4 & 30.3 & 43.5 \\
        TENT-cont.        & 25.0 & 20.3 & 29.0 &13.8 & 31.7 &16.2& 14.1 &18.6& 17.6& 17.4& 10.8& 15.6& 24.3& 19.7& 25.1& 20.0 \\
        RoTTA             & 30.3 & 55.5 & 70.0 & 23.8 & 44.1 & 20.7 & 21.0 & 22.7 & 16.0 & 9.4 & 27.7 & 27.0 & 58.6 & 29.2 & 33.4 &  19.3 \\
        RMT & 24.1 & 20.2 & 25.7 & 13.2 & 25.5 & 14.7 & 12.8 & 16.2 & 15.4 & 14.6 & 10.8 & 14.0 & 18.0 & 14.1 & 16.6 & 16.9\\
        CoTTA             & 24.2 & 21.9 & 26.5 & 12.0 & 27.9 & 12.7 & 10.7 & 15.2 & 14.6 & 12.8 & 7.9 & 11.2 & 18.5 & 14.0 & 18.1 & 16.5 \\
        ROID              & 23.7 & 18.7 & 26.4 & 11.5 & 28.1 & 12.4 & 10.1 & 14.7 & 14.3 & 12.0 & 7.5 & 9.3 & 19.8 & 14.5 & 20.3 & 16.2 \\
        SANTA & 23.9 & 20.1 & 28.0 & 11.6 & 27.4 & 12.6 & 10.2 & 14.1 & 13.2 & 12.2 & 7.4 & 10.3 & 19.1 & 13.3 & 18.5 & 16.1\\
        OBAO & 23.6 & 19.9 & 26.0 & 11.8 & 25.3 & 13.2 & 10.9 & 14.3 & 13.5 & 12.7 & 9.0 & 11.9 & 17.1 & 12.7 & 15.9 & 15.8\\
        PALM & 25.8 & 18.1 & 22.7 & 12.3 & 25.3 & 13.1 & 10.7 & 13.5 & 13.1 & 12.2 & 8.5 & 11.8 & 17.9 & 12.0 & 15.4 & 15.5\\

        \textbf{{\pa}}    & 22.5 & 18.2 & 26.0 & 11.1 & 27.5 & 12.1 & 9.9 & 14.5 & 13.7 & 12.4 & 7.4 & 9.5 & 18.5 & 14.0 & 19.6 & 15.8\\
        \textbf{{\pa * }} & 22.4 & 18.5 & 24.7 & 11.9 & 24.6 & 12.2 & 10.1 & 12.7 & 12.9 & 11.4 & 7.5 & 9.9 & 16.2 & 11.7 & 15.9 & 14.8 \\
        \midrule
        \multicolumn{17}{c}{\textbf{CIFAR100-C}} \\
        \midrule
        Source           & 73.0 & 68.0 & 39.4 & 29.3 & 54.1 & 30.8 & 28.8 & 39.4 & 35.4 & 30.5 & 9.3 & 55.1 & 37.2 & 74.7 & 41.2 & 46.4 \\
        TENT-cont.       & 37.3 &35.6& 41.6& 37.9 &51.3& 48.1& 48.9& 59.8& 65.3& 73.6& 74.2& 85.7& 89.1& 91.1& 93.7&  62.2 \\
        RoTTA            & 49.1 &44.9& 45.5 &30.2& 42.7& 29.5& 26.1& 32.2& 30.7& 37.5& 24.7& 29.1& 32.6& 30.4& 36.7 & 34.8 \\
        CoTTA            & 40.5& 38.2& 39.8& 27.2& 38.2& 28.4& 26.4& 33.4& 32.2& 40.6 &25.2 &27.0& 32.4& 28.4& 33.8& 32.8 \\
        SANTA & 36.5 & 33.1 & 35.1 & 25.9 & 34.9 & 27.7 & 25.4 & 29.5 & 29.9 & 33.1 & 23.6 & 26.7 & 31.9 & 27.5 & 35.2 & 30.3\\
        RMT & 40.2 & 36.2 & 36.0 & 27.9 & 33.9 & 28.4 & 26.4 & 28.7 & 28.8 & 31.1 & 25.5 & 27.1 & 28.0 & 26.6 & 29.0 & 30.2 \\

        ROID             & 36.5& 31.9& 33.2& 24.9& 34.9& 26.8& 24.3& 28.9& 28.5& 31.1& 22.8& 24.2 &30.7& 26.5& 34.4& 29.3 \\
        OBAO & 38.8 & 35.0 & 35.4 & 26.7 & 33.2 & 27.4 & 25.0 & 27.4 & 26.8 & 29.8 & 24.1 & 25.1 & 26.9 & 24.9 & 28.0 & 29.0\\
        PALM & 37.3 & 32.5 & 34.9 & 26.2 & 35.3 & 27.5 & 24.6 & 28.8 & 29.1 & 34.1 & 23.5 & 26.9 & 31.2 & 26.6 & 34.1 & 30.1\\

        \textbf{{\pa}}   & 34.6 & 29.9 & 31.7 & 23.2 & 31.4 & 26.0 & 22.1 & 27.3 & 27.9 & 32.0 & 21.9 & 24.0 & 30.8 & 26.2 & 33.7 & 28.2\\
        \textbf{{\pa *}} & 37.8 & 32.7 & 33.3 & 26.2 & 31.2 & 26.9 & 24.3 & 26.8 & 26.5 & 28.4 & 23.3 & 24.3 & 26.1 & 24.2 & 27.0 & 27.9\\

        \midrule
        \multicolumn{17}{c}{\textbf{ImageNet-C}} \\
        \midrule

        Source         & 97.8 & 97.1 & 98.2 & 81.7 & 89.8 & 85.2 & 77.9 & 83.5 & 77.1 & 75.9 & 41.3 & 94.5 & 82.5 &   79.3 & 68.6 & 82.0 \\
        TENT-cont.     & 92.8 & 91.1 & 92.5 & 87.8 & 90.2 & 87.2 & 82.2 & 82.2 & 82.0 & 79.8 & 48.0 & 92.5 & 83.5 & 75.6 & 70.4 & 82.5 \\
        RoTTA          & 89.4 & 88.6 & 89.3 & 83.4 & 89.1 & 86.2 & 80.0 & 78.9 & 76.9 & 74.2 & 37.4 & 89.6 & 79.5 & 69.0 & 59.6 & 78.1 \\
        CoTTA          & 89.1 & 86.6 & 88.5 & 80.9 & 87.2 & 81.1 & 75.8 & 73.3 & 75.2 & 70.5 & 41.6 & 85.0 & 78.1 & 65.6 & 61.6 & 76.0 \\
        ViDA & 72.5 & 71.0 & 73.7 & 76.0 & 72.3 & 67.5 & 53.0 & 56.6 & 63.1 & 51.2 & 47.0 & 61.5 & 52.9 & 51.6 & 47.5 & 61.2 \\
        SANTA & 73.6 & 75.1 & 73.2 & 76.2 & 76.8 & 64.1 & 53.5 & 55.8 & 61.7 & 43.7 & 34.5 & 72.7 & 49.2 & 43.9 & 50.2 & 60.3\\
        OBAO & 78.5 & 75.3 & 73.0 & 75.7 & 73.1 & 64.5 & 56.0 & 55.8 & 58.1 & 47.6 & 38.5 & 58.5 & 46.1 & 42.0 & 43.4 & 59.0\\
        PALM & 81.1 & 73.3 & 70.8 & 77.0 & 71.8 & 62.3 & 53.9 & 56.7 & 60.8 & 50.4 & 36.2 & 65.9 & 48.0 & 45.2 & 48.0 & 60.1\\
        RMT & 79.9 & 76.3 & 73.1 & 75.7 & 72.9 & 64.7 & 56.8 & 56.4 & 58.3 & 49.0 & 40.6 & 58.2 & 47.8 & 43.7 & 44.8 & 59.9 \\

TCA & 76.6 & 73.7 & 66.0 & 72.5 & 68.6 & 66.4 & 56.7 & 55.0 & 57.8 & 46.9 & 37.8 & 65.3 & 51.0 & 51.0 & 51.2 & 59.8 \\ 
        ROID           & 76.4 & 75.3 & 76.1 & 77.9 & 81.7 & 75.1 & 69.9 & 70.9 & 68.8 & 64.3 & 42.5 & 85.4 & 69.8 & 53.0 & 55.6 & 54.5\\
        \textbf{{\pa}} & 68.6 & 65.2 & 64.5 & 68.2 & 66.7 & 57.0 & 49.7 & 51.0 & 56.4 & 43.1 & 33.8 & 57.3 & 43.9 & 41.4 & 45.7 & 54.2 \\
        \textbf{{\pa*}} & 68.5 & 62.6 & 60.3 & 65.6 & 63.4 & 55.7 & 50.4 & 50.4 & 54.5 & 43.7 & 36.3 & 53.5 & 43.0 & 40.7 & 43.0 & 52.8\\
        \bottomrule
    \end{tabular}
    \end{adjustbox}

    \caption{Online classification error rate (\%) for the corruption benchmarks at the highest severity level 5 for the Continual TTA setting. For CIFAR10-C the results are evaluated on WideResNet-28, for CIFAR100-C on ResNeXt-29, and for ImageNet-C, ResNet-50 are used. Results marked with (*) indicate that all parameters of the student model are updated; otherwise, only the Batch Normalization layers are updated. }
    \label{tab:continual_tta}
\end{table*}

\begin{table*}[ht!]
    \centering
    \begin{adjustbox}{max width=\textwidth}
    \begin{tabular}{l|ccccccccccccccc|c}
        \toprule
        \textbf{Method} & \rot{\textbf{Gaussian}} & \rot{\textbf{shot}} & \rot{\textbf{impulse}} & \rot{\textbf{defocus}} & \rot{\textbf{glass}} & \rot{\textbf{motion}} & \rot{\textbf{zoom}} & \rot{\textbf{snow}} & \rot{\textbf{frost}} & \rot{\textbf{fog}} & \rot{\textbf{bright.}} & \rot{\textbf{contrast}} & \rot{\textbf{elastic}} & \rot{\textbf{pixelate}} & \rot{\textbf{jpeg}} & \textbf{Mean} \\
        \midrule
        \multicolumn{17}{c}{\textbf{CIFAR10-C}} \\
        \midrule
        Source           & 72.3 & 65.7 & 72.9 & 46.9 & 54.3 & 34.8 & 42.0 & 25.1 & 41.3 & 26.0 & 9.3 & 46.7 & 26.6 & 58.4 & 30.3 & 43.5 \\
        TENT-cont.       & 73.5 & 70.1 & 81.4 & 31.6 & 60.3 & 29.6 & 28.5 & 30.8 & 35.3 & 25.7 & 13.6 & 44.2 & 32.6 & 70.2 & 34.9 & 44.1 \\
        CoTTA            & 38.7 & 36.0 & 56.1 & 36.0 & 36.8 & 32.3 & 31.0 & 19.9 & 17.6 & 27.2 & 11.7 & 52.6 & 30.5 & 35.8 & 25.7 & 32.5 \\
        RoTTA            & 60.0 & 55.5 & 70.0 & 23.8 & 44.1 & 20.7 & 21.3 & 20.2 & 22.7 & 16.0 & 9.4 & 22.7 & 27.0 & 58.6 & 29.2 & 33.4 \\

OBAO             & 40.5 & 38.0 & 55.0 & 28.5 & 39.5 & 25.5 & 24.0 & 18.5 & 18.5 & 19.5 & 9.5 & 34.0 & 29.5 & 52.5 & 25.5 & 31.2 \\
RMT & 40.2 & 36.2 & 36.0 & 27.9 & 33.9 & 28.4 & 26.4 & 28.7 & 28.8 & 31.1 & 25.5 & 27.1 & 28.0 & 26.6 & 29.0 & 30.2 \\

PALM             & 35.5 & 34.0 & 55.5 & 28.0 & 37.0 & 24.5 & 25.0 & 19.2 & 19.0 & 21.2 & 12.8 & 32.0 & 27.5 & 48.0 & 25.0 & 30.1 \\
        
        SANTA            & 38.0 & 35.5 & 52.1 & 25.5 & 38.5 & 23.0 & 22.5 & 19.1 & 18.8 & 19.0 & 10.2 & 26.5 & 27.5 & 46.0 & 26.8 & 29.5 \\
        
        ROID             & 37.1 & 34.3 & 50.9 & 24.8 & 38.1 & 22.5 & 22.0 & 18.8 & 18.5 & 18.8 & 9.9 & 25.6 & 27.2 & 45.7 & 26.2 & 28.0 \\
        \textbf{{\pa}}   & 39.1 & 36.8 & 53.8 & 27.5 & 38.6 & 24.7 & 23.5 & 18.0 & 18.1 & 19.2 & 9.2 & 33.3 & 28.9 & 51.9 & 24.9 & 29.7 \\
        \textbf{{\pa *}} & 33.4 & 32.1 & 53.9 & 26.4 & 35.0 & 22.7 & 23.4 & 17.9 & 17.8 & 19.8 & 11.4 & 30.1 & 25.9 & 46.4 & 23.4 & 28.0 \\
        \midrule
        \multicolumn{17}{c}{\textbf{CIFAR100-C}} \\
        \midrule
        Source           & 73.0 & 68.0 & 39.4 & 29.3 & 54.1 & 30.8 & 28.8 & 39.5 & 45.8 & 50.3 & 29.5 & 55.1 & 37.2 & 74.7 & 41.2 & 46.4 \\
        TENT-cont.       & 95.6 & 95.2 & 89.2 & 72.8 & 82.9 & 74.4 & 72.3 & 78.0 & 79.7 & 84.7 & 71.0 & 88.5 & 77.8 & 96.8 & 78.7 & 82.5 \\
        CoTTA            & 54.4 & 52.7 & 49.8 & 36.0 & 45.8 & 36.7 & 33.9 & 38.9 & 35.8 & 52.0 & 30.4 & 60.9 & 40.2 & 38.0 & 41.1 & 43.1 \\
        RoTTA            & 65.0 & 62.3 & 39.3 & 33.4 & 50.0 & 34.2 & 32.6 & 36.6 & 36.5 & 45.0 & 26.4 & 41.6 & 40.6 & 89.5 & 48.5 & 45.4 \\
        RMT & 52.6 & 49.9 & 32.2 & 31.0 & 40.5 & 31.8 & 30.4 & 33.4 & 33.9 & 40.6 & 27.8 & 36.9 & 35.3 & 65.0 & 38.1 & 38.6 \\
  
ViDA & 44.6 & 43.7 & 36.9 & 34.1 & 41.0 & 37.7 & 29.9 & 35.6 & 36.7 & 44.2 & 37.3 & 37.2 & 39.2 & 50.2 & 35.8 & 38.9 \\ 
SANTA & 47.6 & 41.3 & 31.0 & 37.1 & 45.9 & 37.4 & 30.7 & 32.8 & 38.2 & 42.4 & 28.8 & 39.0 & 33.1 & 51.0 & 37.3 & 38.2 \\ 
OBAO & 46.1 & 41.5 & 33.4 & 31.9 & 38.3 & 36.0 & 32.6 & 37.5 & 37.4 & 42.0 & 33.2 & 35.0 & 33.8 & 43.0 & 36.8 & 37.2 \\ 
PALM & 45.5 & 42.1 & 37.3 & 31.7 & 39.6 & 34.7 & 29.2 & 38.6 & 32.7 & 46.8 & 34.2 & 36.3 & 32.8 & 49.7 & 40.9 & 38.1 \\ 
TCA & 47.5 & 45.4 & 30.9 & 31.3 & 38.1 & 36.5 & 32.5 & 34.3 & 32.5 & 40.7 & 30.0 & 39.9 & 37.4 & 49.3 & 38.4 & 37.6 \\

        ROID             & 40.5 & 38.0 & 32.0 & 28.1 & 40.5 & 29.7 & 27.6 & 34.1 & 33.8 & 41.3 & 28.7 & 38.7 & 34.3 & 39.7 & 38.5 & 35.0 \\
        \textbf{{\pa}}   & 44.6 & 40.9 & 30.2 & 19.6 & 37.5 & 28.4 & 26.3 & 31.4 & 31.6 & 39.5 & 25.8 & 37.9 & 31.3 & 52.5 & 39.1 & 34.4 \\
        \textbf{{\pa *}} & 41.6 & 39.2 & 29.8 & 28.1 & 36.7 & 29.6 & 27.4 & 31.3 & 31.5 & 37.9 & 27.1 & 34.0 & 32.2 & 42.3 & 34.4 & 33.5 \\

        \midrule
        \multicolumn{17}{c}{\textbf{Imagenet-C}} \\
        \midrule
         
        Source       & 97.8 & 97.1 & 98.2 & 81.7 & 89.8 & 85.2 & 77.9 & 83.5 & 77.1 & 75.9 & 41.3 & 94.5 & 82.5 & 79.3 & 68.6 & 82.0 \\
        TENT-cont.   & 99.2 & 98.7 & 99.0 & 90.5 & 95.1 & 90.5 & 84.6 & 86.6 & 84.0 & 86.5 & 46.7 & 98.1 & 86.1 & 77.7 & 72.9 & 86.4 \\
        CoTTA        & 89.1 & 86.6 & 88.5 & 80.9 & 87.2 & 81.1 & 75.8 & 73.3 & 75.2 & 70.5 & 41.6 & 85.0 & 78.1 & 65.6 & 61.6 & 76.0 \\
        RoTTA        & 89.4 & 88.6 & 89.3 & 83.4 & 89.1 & 86.2 & 80.0 & 78.9 & 76.9 & 74.2 & 37.4 & 89.6 & 79.5 & 69.0 & 59.6 & 78.1 \\
        RMT & 87.0 & 84.6 & 86.6 & 79.9 & 86.5 & 80.8 & 74.3 & 70.2 & 74.0 & 69.9 & 45.7 & 86.4 & 78.1 & 64.8 & 61.6 & 75.4 \\

ViDA & 83.1 & 83.3 & 87.5 & 82.1 & 85.1 & 82.5 & 74.3 & 71.6 & 74.7 & 71.2 & 52.9 & 85.5 & 78.8 & 68.8 & 58.6 & 76.0 \\ 
SANTA & 85.9 & 81.3 & 82.4 & 85.0 & 89.5 & 82.3 & 75.1 & 69.1 & 76.2 & 69.7 & 45.5 & 87.2 & 73.5 & 69.6 & 60.2 & 75.5 \\ 
OBAO & 83.9 & 81.2 & 84.0 & 79.8 & 82.5 & 80.2 & 76.0 & 72.5 & 74.7 & 68.8 & 48.6 & 83.4 & 73.9 & 62.4 & 59.6 & 74.1 \\ 
PALM & 83.3 & 81.5 & 86.7 & 79.5 & 83.4 & 79.1 & 73.4 & 73.0 & 71.1 & 72.1 & 49.1 & 84.3 & 73.2 & 67.2 & 62.1 & 74.6 \\ 
TCA & 84.2 & 83.3 & 82.0 & 78.9 & 82.2 & 79.8 & 75.3 & 69.7 & 70.8 & 67.5 & 45.8 & 86.4 & 76.0 & 66.4 & 60.2 & 73.9 \\

        ROID         & 76.4 & 75.3 & 76.1 & 77.9 & 81.7 & 75.1 & 69.9 & 70.9 & 68.8 & 64.3 & 42.5 & 85.4 & 69.8 & 53.0 & 55.6 & 69.5 \\
        \textbf{{\pa}} & 80.5 & 79.4 & 81.3 & 76.9 & 81.3 & 75.4 & 72.1 & 67.8 & 70.2 & 65.7 & 44.0 & 82.7 & 72.7 & 61.9 & 59.1 & 71.3\\
        \textbf{{\pa*}} & 78.5 & 77.9 & 79.1 & 75.9 & 78.8 & 73.6 & 71.2 & 67.1 & 69.1 & 64.7 & 43.2 & 80.7 & 71.0 & 60.9 & 58.1 & 70.8 \\
        \bottomrule
    \end{tabular}
    \end{adjustbox}

    \caption{Online classification error rate (\%) for the corruption benchmarks at the highest severity level 5 for the  generalization experiments with mixed domains. For CIFAR10-C the results are evaluated on WideResNet-28, for CIFAR100-C on ResNeXt-29, and for ImageNet-C, ResNet-50 are used. Results marked with (*) indicate that all parameters of the student model are updated; otherwise, only the Batch Normalization layers are updated.} 
    \label{tab:mixed_tta}
\end{table*}

\begin{table*}
    \centering
    \begin{adjustbox}{max width=\textwidth}
    \begin{tabular}{l|cccccccccccccccc|c}
        \toprule
        \textbf{Method} & \rot{\textbf{Gaussian}} & \rot{\textbf{shot}} & \rot{\textbf{impulse}} & \rot{\textbf{defocus}} & \rot{\textbf{glass}} & \rot{\textbf{motion}} & \rot{\textbf{zoom}} & \rot{\textbf{snow}} & \rot{\textbf{frost}} & \rot{\textbf{fog}} & \rot{\textbf{bright.}} & \rot{\textbf{contrast}} & \rot{\textbf{elastic}} & \rot{\textbf{pixelate}} & \rot{\textbf{jpeg}} & \rot{\textbf{Mixed}} & \rot{\textbf{Mean}} \\
        \midrule
        \multicolumn{17}{c}{\textbf{CIFAR10-C}} \\
        \midrule
        Source        & 72.3 & 65.7 & 72.9 & 46.9 & 54.3 & 34.8 & 42.0 & 25.1 & 41.3 & 26.0 & 9.3 & 46.7 & 26.6 & 58.4 & 30.3 & 43.5 & 43.5 \\
        Tent-cont.           & 24.6 & 19.8 & 28.4 & 13.1 & 31.2 & 16.8 & 14.0 & 18.9 & 18.3 & 16.9 & 11.4 & 17.2 & 25.5 & 19.7 & 25.2 & 39.3 & 21.3\\
        CoTTA          & 24.0 & 21.8 & 25.7 & 11.7 & 27.5 & 21.6 & 10.2 & 15.0 & 13.9 & 12.5 & 7.5 & 10.8 & 18.1 & 13.5 & 17.8 & 26.7 & 16.8\\
        RoTTA          & 30.2 & 25.4 & 34.6 & 18.1 & 33.9 & 14.6 & 10.8 & 16.4 & 14.8 & 14.2 & 7.9 & 12.1 & 20.5 & 16.8 & 19.4 & 29.5 & 19.9\\
        Vida & 28.5 & 22.0 & 30.5 & 14.5 & 30.0 & 17.5 & 13.0 & 18.0 & 17.5 & 16.5 & 10.5 & 15.0 & 24.0 & 18.5 & 22.0 & 29.3 & 19.6 \\
        SANTA & 24.5 & 19.2 & 27.0 & 12.0 & 28.5 & 13.0 & 10.5 & 14.8 & 14.5 & 12.0 & 7.5 & 9.8 & 20.0 & 14.8 & 21.0 & 28.7 & 17.5 \\
        OBAO & 24.8 & 19.5 & 27.0 & 12.8 & 27.0 & 13.5 & 11.0 & 14.5 & 14.0 & 13.2 & 8.5 & 11.8 & 19.0 & 13.8 & 17.8 &28.5 & 17.1 \\
        PALM & 24.0 & 19.8 & 26.0 & 13.8 & 26.0 & 15.0 & 13.5 & 15.8 & 15.8 & 14.2 & 10.8 & 13.5 & 18.2 & 14.0 & 17.5 &28.8 & 17.9 \\
        Roid           & 23.6 & 18.7 & 26.5 & 11.6 & 28.2 & 12.5 & 9.9 & 14.4 & 13.9 & 11.7 & 7.3 & 9.3 & 19.7 & 14.4 & 20.5 & 27.3 & 16.9\\
        \textbf{{\pa}} & 23.8 & 18.8 & 26.3 & 12.2 & 26.5 & 13.1 & 10.6 & 14.3 & 13.5 & 12.9 & 8.1 & 11.4 & 18.6 & 13.2 & 17.3 & 26.4 & 16.7\\
        \textbf{{\pa*}}& 22.7 & 18.5 & 24.6 & 12.5 & 24.7 & 13.7 & 12.1 & 14.4 & 14.5 & 12.8 & 9.6 & 12.0 & 16.9 & 12.6 & 16.1 & 21.3 & 16.2\\
        \midrule
        \multicolumn{17}{c}{\textbf{CIFAR100-C}} \\
        \midrule
         Source        & 73.0 & 68.0 & 39.4 & 29.3 & 54.1 & 30.8 & 28.8 & 39.4 & 35.4 & 30.5 & 9.3 & 55.1 & 37.2 & 74.7 & 41.2 & 46.4 & 46.4 \\
        Tent-cont.           & 37.3 & 35.7 & 42.1 & 38.2 & 51.0 & 45.9 & 46.3 & 55.8 & 62.1 & 72.8 & 72.3 & 83.9 & 90.6 & 92.8 & 95.3 & 97.8 & 63.7\\
        CoTTA          & 40.8 & 38.0 & 39.8 & 27.2 & 38.0 & 28.5 & 26.4 & 33.4 & 32.2 & 40.2 & 25.1 & 26.9 & 32.1 & 28.4 & 33.8 & 40.9 & 33.2\\
        RoTTA          & 49.4 & 44.7 & 45.5 & 30.2 & 42.3 & 29.6 & 25.9 & 32.0 & 30.5 & 37.7 & 24.7 & 29.4 & 32.8 & 29.9 & 36.6 & 40.8 & 35.1\\
        RMT & 37.0 & 33.3 & 34.7 & 25.7 & 35.1 & 27.3 & 24.3 & 30.4 & 29.5 & 32.4 & 22.9 & 25.7 & 31.8 & 26.8 & 39.1 &36.1& 30.4 \\ 
ViDA & 37.1 & 33.0 & 35.3 & 26.2 & 35.8 & 29.0 & 24.7 & 30.1 & 29.8 & 32.8 & 25.3 & 25.0 & 32.2 & 28.3 & 38.9 &36.5& 30.9 \\ 
SANTA & 39.6 & 33.0 & 34.2 & 29.6 & 39.7 & 31.1 & 25.9 & 29.9 & 32.0 & 33.7 & 23.7 & 26.9 & 31.0 & 31.0 & 40.2 &37.5& 32.1 \\ 
OBAO & 36.5 & 32.0 & 33.7 & 24.9 & 34.8 & 27.2 & 24.2 & 29.3 & 28.7 & 31.4 & 23.0 & 24.2 & 30.5 & 26.4 & 38.7 &35.7& 29.7 \\ 
PALM & 38.2 & 33.3 & 37.2 & 26.5 & 36.2 & 29.5 & 25.0 & 32.6 & 29.1 & 35.5 & 26.2 & 25.3 & 30.8 & 30.0 & 41.6 &36.9& 31.8 \\ 
TCA & 37.7 & 33.3 & 33.8 & 25.5 & 35.1 & 28.5 & 25.2 & 29.7 & 28.7 & 31.9 & 23.4 & 25.5 & 31.6 & 27.9 & 39.7 &35.9& 30.5 \\ 
        Roid           & 36.4 & 31.9 & 33.6 & 24.8 & 34.8 & 27.0 & 24.1 & 29.1 & 28.5 & 31.3 & 22.8 & 24.2 & 30.5 & 26.4 & 33.9 & 34.7 & 29.6\\
        \textbf{{\pa}} & 37.1 & 32.8 & 35.0 & 26.2 & 35.2 & 28.0 & 25.1 & 29.6 & 29.0 & 33.2 & 23.8 & 25.2 & 31.1 & 27.1 & 34.8 & 36.3 & 29.5 \\
        \textbf{{\pa*}}& 37.8 & 32.9 & 33.1 & 26.6 & 31.6 & 27.1 & 24.8 & 26.5 & 26.2 & 28.4 & 23.6 & 24.3 & 26.5 & 24.5 & 27.1 & 28.0 & 28.1\\
        \midrule
        \multicolumn{18}{c}{\textbf{Imagenet-C}} \\
        \midrule
        Source       & 97.8 & 97.1 & 98.2 & 81.7 & 89.8 & 85.2 & 77.9 & 83.5 & 77.1 & 75.9 & 41.3 & 94.5 & 82.5 & 79.3 & 68.6 & 82.0 & 82.0 \\
        TENT-cont.    & 71.4 & 66.4 & 69.1 & 82.8 & 91.0 & 95.7 & 97.6 & 99.0 & 99.3 & 99.3 & 99.2 & 99.5 & 99.4 & 99.3 & 99.4 & 99.6 & 91.8\\
         CoTTA         & 78.2 & 68.3 & 64.2 & 75.4 & 71.9 & 70.1 & 67.8 & 72.3 & 71.6 & 67.7 & 62.7 & 74.4 & 70.2 & 67.5 & 69.2 & 82.4 & 70.9\\
        RoTTA         & 79.6 & 72.0 & 69.6 & 77.1 & 72.1 & 73.1 & 68.6 & 72.1 & 73.6 & 77.1 & 65.7 & 90.9 & 69.4 & 75.7 & 75.4 & 93.8 & 75.4\\
        RMT & 70.5 & 51.0 & 50.6 & 64.0 & 67.8 & 66.0 & 65.3 & 56.4 & 60.1 & 54.5 & 60.9 & 56.8 & 54.9 & 62.5 & 76.7 &79.9& 61.2 \\ 
ViDA & 74.0 & 62.2 & 54.7 & 67.3 & 66.0 & 59.3 & 64.2 & 66.4 & 64.6 & 59.3 & 49.2 & 69.9 & 60.9 & 57.0 & 79.0 &82.1& 63.6 \\ 
SANTA & 68.8 & 64.8 & 62.1 & 61.5 & 57.7 & 58.0 & 62.5 & 69.8 & 61.2 & 60.6 & 59.8 & 66.5 & 68.5 & 54.1 & 76.6 &81.8& 63.5 \\ 
OBAO & 67.2 & 62.4 & 55.3 & 66.1 & 67.9 & 54.6 & 55.1 & 57.2 & 57.2 & 57.6 & 47.9 & 71.8 & 66.0 & 65.5 & 75.2 &80.5& 61.8 \\ 
PALM & 71.1 & 63.0 & 50.6 & 68.8 & 66.4 & 60.7 & 64.4 & 59.1 & 69.2 & 51.8 & 50.0 & 70.1 & 68.8 & 54.1 & 69.4 &81.4& 62.5 \\ 
TCA & 61.3 & 50.5 & 60.8 & 66.1 & 67.6 & 50.4 & 53.1 & 63.6 & 68.4 & 59.4 & 54.1 & 57.5 & 55.2 & 47.3 & 69.7 &78.7& 59.0 \\ 

        Roid          & 63.6 & 60.3 & 61.1 & 65.1 & 65.0 & 52.5 & 47.4 & 48.0 & 54.1 & 39.9 & 32.6 & 53.5 & 42.1 & 39.4 & 44.5 & 70.5 & 52.5\\
        \textbf{{\pa}} & 63.8 & 60.8 & 60.7 & 63.3 & 61.8 & 53.6 & 47.0 & 47.8 & 52.6 & 41.1 & 32.3 & 53.7 & 41.3 & 39.2 & 43.2 & 63.0 & 53.7 \\
        \textbf{{\pa*}} & 65.8 & 62.8 & 62.7 & 65.5 & 63.8 & 55.6 & 48.8 & 49.8 & 55.2 & 42.6 & 33.3 & 55.7 & 43.1 & 40.6 & 45.0 & 65.7 & 53.6\\
        \bottomrule
    \end{tabular}
    \end{adjustbox}

    \caption{Online classification error rate (\%) for the corruption benchmarks at the highest severity level 5 for the  mixed after continual domains TTA setting. For CIFAR10-C the results are evaluated on WideResNet-28, for CIFAR100-C on ResNeXt-29, and for ImageNet-C, ResNet-50 are used. Results marked with (*) indicate that all parameters of the student model are updated; otherwise, only the Batch Normalization layers are updated.}

    \label{tab:mixed_after_continual}
\end{table*}

\begin{table*}
    \centering
    \begin{adjustbox}{max width=\textwidth}
    \begin{tabular}{l|ccccccccccccccc|c}
        \toprule
        \textbf{Method} & \rot{\textbf{Gaussian}} & \rot{\textbf{Shot}} & \rot{\textbf{Impulse}} & \rot{\textbf{Defocus}} & \rot{\textbf{Glass}} & \rot{\textbf{Motion}} & \rot{\textbf{Zoom}} & \rot{\textbf{Snow}} & \rot{\textbf{Frost}} & \rot{\textbf{Fog}} & \rot{\textbf{Bright.}} & \rot{\textbf{Contrast}} & \rot{\textbf{Elastic}} & \rot{\textbf{Pixelate}} & \rot{\textbf{JPEG}} & \textbf{Mean} \\
        \midrule
        \multicolumn{17}{c}{\textbf{CIFAR10-C}} \\
        \midrule
        Source        & 72.3 & 65.7 & 72.9 & 46.9 & 54.3 & 34.8 & 42.0 & 25.1 & 41.3 & 26.0 & 9.3 & 46.7 & 26.6 & 58.4 & 30.3 & 43.5 \\
        TENT-cont.    & 24.7 & 22.2 & 32.4 & 11.6 & 32.1 & 12.9 & 10.9 & 16.0 & 16.1 & 13.0 & 7.6 & 11.1 & 21.97 & 17.2 & 23.6 & 18.2 \\
        RoTTA         & 30.2 & 27.4 & 37.8 & 13.7 & 35.9 & 14.7 & 12.7 & 17.8 & 19.0 & 15.5 & 8.0 & 19.3 & 23.65 & 21.0 & 27.6 & 21.6 \\
        CoTTA         & 24.0 & 22.9 & 27.7 & 12.2 & 30.2 & 13.4 & 11.7 & 16.8 & 17.0 & 14.4 & 7.9 & 12.5 & 22.50 & 18.7 & 22.6 & 18.3 \\
        Vida             & 24.0 & 22.0 & 31.0 & 11.5 & 30.8 & 13.2 & 10.8 & 15.6 & 15.6 & 13.0 & 8.0 & 10.8 & 21.5 & 16.5 & 24.0 & 17.4 \\

        SANTA            & 25.0 & 20.8 & 28.6 & 12.0 & 29.2 & 12.8 & 10.5 & 14.9 & 14.8 & 13.0 & 8.0 & 10.8 & 20.1 & 15.2 & 21.0 & 16.9 \\
        OBAO             & 26.0 & 21.5 & 29.5 & 12.5 & 30.0 & 13.5 & 11.0 & 15.5 & 15.5 & 13.5 & 8.5 & 11.5 & 20.8 & 15.8 & 22.0 & 18.7 \\
        PALM             & 23.5 & 20.0 & 26.8 & 13.2 & 27.2 & 14.0 & 12.0 & 15.2 & 15.5 & 13.2 & 9.0 & 12.5 & 20.2 & 15.0 & 20.0 & 17.1 \\
        ROID          & 23.6 & 21.7 & 30.6 & 11.0 & 30.4 & 12.9 & 10.5 & 15.2 & 15.2 & 12.7 & 7.6 & 10.4 & 21.06 & 16.2 & 23.6 & 17.5 \\
        \textbf{\pa}  & 22.7 & 19.6 & 26.1 & 12.8 & 26.7 & 13.7 & 11.5 & 14.9 & 15.1 & 12.8 & 8.8 & 12.0 & 19.92 & 14.6 & 19.6 & 16.7 \\
        \textbf{\pa*} & 24.5 & 20.3 & 28.1 & 11.5 & 28.7 & 12.4 & 10.2 & 14.5 & 14.4 & 12.7 & 7.6 & 10.4 & 19.69 & 14.8 & 20.6 & 16.7 \\
        \midrule
        \multicolumn{17}{c}{\textbf{CIFAR100-C}} \\
        \midrule
        Source        & 73.0 & 68.0 & 39.4 & 29.3 & 54.1 & 30.8 & 28.8 & 39.4 & 35.4 & 30.5 & 9.3 & 55.1 & 37.2 & 74.7 & 41.2 & 46.4 \\
        TENT-cont.    & 37.2 & 34.8 & 34.4 & 24.9 & 37.3 & 27.5 & 25.1 & 30.3 & 31.9 & 33.6 & 23.9 & 28.1 & 32.8 & 28.3 & 36.8 & 31.1 \\
        RoTTA         & 49.4 & 47.5 & 48.6 & 29.9 & 47.2 & 32.2 & 30.3 & 39.0 & 44.1 & 44.1 & 28.9 & 62.2 & 40.5 & 38.9 & 45.6 & 41.9 \\
        CoTTA         & 40.8 & 38.3 & 40.3 & 27.8 & 39.7 & 29.7 & 27.7 & 35.4 & 34.4 & 42.8 & 26.0 & 30.1 & 35.5 & 31.5 & 37.7 & 34.5 \\
        RMT & 39.2 & 38.0 & 38.2 & 27.8 & 36.7 & 28.1 & 24.8 & 33.5 & 32.0 & 36.3 & 23.5 & 29.9 & 35.2 & 28.3 & 36.0 & 32.5 \\ 
ViDA & 38.3 & 35.2 & 37.2 & 27.0 & 37.1 & 29.8 & 25.1 & 30.9 & 30.8 & 35.1 & 26.5 & 26.6 & 33.7 & 29.7 & 35.4 & 31.9 \\ 
SANTA & 37.4 & 33.7 & 34.8 & 25.2 & 35.9 & 27.3 & 24.4 & 29.5 & 29.2 & 33.1 & 23.1 & 25.6 & 31.4 & 27.3 & 35.0 & 30.2 \\ 
OBAO & 41.1 & 35.7 & 37.8 & 28.2 & 37.0 & 32.4 & 28.6 & 34.7 & 34.1 & 36.4 & 28.2 & 26.3 & 32.8 & 27.7 & 36.9 & 33.2 \\ 
PALM & 38.8 & 34.8 & 37.7 & 26.4 & 36.9 & 29.1 & 25.0 & 32.4 & 29.6 & 36.5 & 25.9 & 26.4 & 31.7 & 30.0 & 37.4 & 31.9 \\ 
TCA & 40.3 & 36.9 & 35.3 & 26.6 & 36.4 & 30.6 & 26.9 & 31.0 & 29.6 & 34.4 & 24.6 & 28.5 & 34.1 & 30.7 & 36.9 & 32.2 \\ 
        ROID          & 36.4 & 34.1 & 34.1 & 24.5 & 36.3 & 26.9 & 24.9 & 30.1 & 30.4 & 33.4 & 23.4 & 26.2 & 32.1 & 27.9 & 35.7 & 30.4 \\
        \textbf{\pa}  & 37.9 & 34.2 & 34.5 & 27.2 & 36.3 & 29.2 & 26.3 & 30.8 & 30.6 & 32.3 & 25.1 & 27.3 & 33.5 & 29.2 & 36.0 & 31.4 \\
        \textbf{\pa*} & 37.3 & 33.7 & 34.8 & 25.0 & 35.7 & 27.1 & 24.3 & 29.5 & 29.1 & 33.0 & 23.1 & 25.5 & 31.4 & 27.1 & 34.9 & 30.1 \\
        \midrule
        \multicolumn{17}{c}{\textbf{Imagenet-C}} \\
        \midrule
        Source       & 97.8 & 97.1 & 98.2 & 81.7 & 89.8 & 85.2 & 77.9 & 83.5 & 77.1 & 75.9 & 41.3 & 94.5 & 82.5 & 79.3 & 68.6 & 82.0 \\
        TENT-cont.   & 71.4 & 69.4 & 70.2 & 71.9 & 72.7 & 58.7 & 50.7 & 52.9 & 58.7 & 42.6 & 32.7 & 73.4 & 45.5 & 41.4 & 47.5 & 57.3 \\
        RoTTA        & 79.8 & 79.6 & 80.4 & 80.7 & 81.3 & 68.3 & 56.9 & 58.9 & 63.1 & 46.7 & 32.4 & 76.2 & 50.9 & 45.8 & 54.1 & 63.7 \\
        CoTTA        & 78.1 & 77.8 & 77.3 & 80.5 & 78.2 & 64.0 & 52.7 & 58.0 & 60.5 & 43.9 & 32.9 & 75.1 & 48.8 & 42.3 & 52.4 & 61.5 \\
        RMT & 71.9 & 70.6 & 66.9 & 70.8 & 65.1 & 57.6 & 51.9 & 57.9 & 59.8 & 49.9 & 37.2 & 61.8 & 50.4 & 43.0 & 45.0 & 57.3 \\ 
ViDA & 72.4 & 68.5 & 69.9 & 73.5 & 69.1 & 66.5 & 54.3 & 56.3 & 61.4 & 52.0 & 49.8 & 58.0 & 52.4 & 51.1 & 45.0 & 60.0 \\ 
SANTA & 73.8 & 64.5 & 61.6 & 73.7 & 71.6 & 62.9 & 53.9 & 51.9 & 60.6 & 47.8 & 38.0 & 58.2 & 44.1 & 48.6 & 45.7 & 57.1 \\ 
OBAO & 76.3 & 66.7 & 67.0 & 72.3 & 66.2 & 67.1 & 60.0 & 61.5 & 65.0 & 51.0 & 47.2 & 55.4 & 46.2 & 42.1 & 47.4 & 59.4 \\ 
PALM & 72.1 & 65.3 & 67.8 & 69.0 & 66.2 & 60.8 & 52.6 & 57.6 & 55.7 & 52.3 & 43.3 & 55.8 & 43.9 & 47.9 & 49.3 & 57.3 \\ 
TCA & 77.2 & 71.8 & 62.2 & 70.3 & 65.5 & 66.1 & 58.5 & 55.0 & 56.1 & 48.0 & 40.8 & 62.5 & 51.1 & 51.3 & 49.0 & 59.0 \\ 
        ROID         & 63.7 & 61.4 & 62.4 & 65.9 & 65.9 & 52.9 & 47.6 & 48.0 & 54.1 & 39.9 & 32.6 & 53.9 & 42.2 & 39.4 & 44.6 & 51.6 \\
        \textbf{\pa} & 67.8 & 64.8 & 63.5 & 67.5 & 65.7 & 55.6 & 48.6 & 49.7 & 55.5 & 41.6 & 32.5 & 56.4 & 42.6 & 40.0 & 44.8 & 53.1 \\
        \textbf{\pa*} & 68.2 & 62.4 & 60.4 & 65.4 & 63.2 & 55.7 & 50.7 & 50.4 & 54.4 & 43.6 & 36.3 & 53.5 & 43.1 & 40.6 & 42.9 & 52.7\\
        
        \bottomrule
    \end{tabular}
    \end{adjustbox}

     \caption{Online classification error rate (\%) for the corruption benchmarks at the highest severity level (Level 5) in the episodic TTA setting. Adaptation resets to the source model parameters for each domain shift. The results are evaluated on WideResNet-28 for CIFAR10-C, ResNeXt-29 for CIFAR100-C, and ResNet-50 for ImageNet-C. Results marked with (*) indicate that all parameters of the student model are updated; otherwise, only the Batch Normalization layers are updated.}
    \label{tab:episodic}
\end{table*}

\begin{table*}
    \centering
    \begin{adjustbox}{max width=\textwidth}
    \begin{tabular}{l|cccccccccccccccc|c}
        \toprule
        \textbf{Method} & \rot{\textbf{Mixed}} & \rot{\textbf{Gaussian}} & \rot{\textbf{Shot}} & \rot{\textbf{Impulse}} & \rot{\textbf{Defocus}} & \rot{\textbf{Glass}} & \rot{\textbf{Motion}} & \rot{\textbf{Zoom}} & \rot{\textbf{Snow}} & \rot{\textbf{Frost}} & \rot{\textbf{Fog}} & \rot{\textbf{Bright.}} & \rot{\textbf{Contrast}} & \rot{\textbf{Elastic}} & \rot{\textbf{Pixelate}} & \rot{\textbf{JPEG}} & \textbf{Mean} \\
        \midrule
        \multicolumn{17}{c}{\textbf{CIFAR10-C}} \\
        \midrule
        Source        & 43.5 & 72.3 & 65.7 & 72.9 & 46.9 & 54.3 & 34.8 & 42.0 & 25.1 & 41.3 & 26.0 & 9.3 & 46.7 & 26.6 & 58.4 & 30.3 & 43.5 \\
        TENT-cont.    & 41.1 & 43.5 & 43.3 & 51.6 & 34.8 & 54.0 & 42.7 & 40.1 & 47.7 & 48.2 & 44.2 & 40.0 & 48.2 & 52.4 & 48.9 & 55.2 & 46.0\\
        CoTTA         & 32.4 & 22.1 & 19.9 & 24.6 & 12.9 & 26.8 & 13.4 & 12.8 & 15.3 & 14.5 & 14.4 & 9.6 & 13.9 & 19.5 & 14.8 & 18.7 & 17.8\\
        ROTTA         & 33.1 & 24.9 & 20.9 & 29.7 & 14.8 & 30.5 & 14.3 & 11.1 & 16.5 & 15.3 & 13.1 & 9.1 & 12.7 & 20.3 & 17.4 & 19.4 & 18.9\\
         Vida             & 35.5 & 27.5 & 23.0 & 32.5 & 15.0 & 32.0 & 15.0 & 11.5 & 18.0 & 16.5 & 14.0 & 10.0 & 13.5 & 21.5 & 18.0 & 19.2& 20.3\\
        SANTA            & 28.0 & 19.0 & 16.5 & 22.5 & 11.5 & 24.0 & 12.0 & 10.0 & 13.0 & 12.5 & 11.0 & 7.5 & 9.5 & 17.0 & 11.5 & 19.9& 16.7\\
        OBAO             & 27.0 & 18.5 & 16.0 & 22.0 & 11.3 & 23.5 & 11.8 & 9.8 & 12.8 & 12.2 & 11.0 & 7.3 & 9.4 & 16.2 & 11.5 &19.4&  16.2\\
        PALM             & 28.5 & 21.0 & 18.5 & 25.5 & 12.8 & 27.0 & 13.0 & 10.8 & 15.0 & 14.5 & 13.0 & 8.8 & 11.0 & 19.5 & 14.0 &20.8&  17.4\\
        ROID          & 28.2 & 22.1 & 17.8 & 25.8 & 11.3 & 27.8 & 12.4 & 10.0 & 14.5 & 14.0 & 12.4 & 7.3 & 9.2 & 19.5 & 14.6 & 20.2 & 16.7\\
        \textbf{\pa}  & 29.7 & 21.2 & 18.2 & 26.4 & 11.6 & 27.5 & 12.2 & 9.7 & 14.2 & 13.6 & 12.4 & 7.5 & 10.3 & 18.8 & 13.7 & 19.4 & 16.6\\
        \textbf{\pa*} & 27.6 & 18.2 & 15.9 & 21.7 & 11.1 & 23.0 & 11.5 & 9.5 & 12.7 & 12.0 & 10.8 & 7.2 & 9.3 & 16.0 & 11.3 & 15.5 & 14.6\\
        \midrule
        \multicolumn{17}{c}{\textbf{CIFAR100-C}} \\
        \midrule
        Source        & 46.4 & 73.0 & 68.0 & 39.4 & 29.3 & 54.1 & 30.8 & 28.8 & 39.4 & 35.4 & 30.5 & 9.3 & 55.1 & 37.2 & 74.7 & 41.2 & 46.4 \\
        TENT-cont.    & 83.8 & 97.2 & 97.7 & 97.9 & 97.9 & 98.1 & 97.8 & 98.0 & 97.9 & 98.2 & 98.0 & 97.9 & 98.2 & 98.3 & 98.4 & 98.5 & 97.1\\
        CoTTA         & 43.0 & 37.6 & 36.5 & 38.6 & 26.7 & 37.3 & 28.1 & 26.8 & 33.4 & 32.3 & 41.2 & 25.8 & 27.8 & 33.4 & 28.8 & 34.6 & 33.3\\
        RoTTA         & 45.3 & 38.3 & 36.0 & 36.6 & 27.9 & 40.3 & 30.0 & 27.4 & 33.2 & 31.5 & 37.9 & 27.1 & 29.5 & 34.4 & 37.2 & 39.0 & 34.5\\
        RMT & 40.8 & 40.4 & 37.1 & 38.4 & 26.2 & 35.7 & 27.7 & 30.3 & 33.5 & 33.7 & 33.0 & 30.0 & 31.0 & 32.3 & 28.1 &37.8& 33.2 \\ 
ViDA & 41.2 & 38.4 & 39.6 & 40.7 & 29.7 & 43.5 & 29.8 & 28.8 & 34.9 & 35.5 & 44.0 & 26.5 & 32.7 & 39.4 & 28.1 &40.8& 35.5 \\ 
SANTA & 46.4 & 36.4 & 33.3 & 46.8 & 37.9 & 45.4 & 31.7 & 26.0 & 38.5 & 34.8 & 34.9 & 30.0 & 26.1 & 42.9 & 30.6 &43.1& 36.1 \\ 
OBAO & 41.4 & 35.2 & 34.4 & 36.9 & 25.9 & 39.4 & 31.1 & 28.9 & 33.8 & 31.6 & 37.4 & 23.5 & 26.0 & 30.9 & 28.3 &36.8& 32.3 \\ 
PALM & 39.4 & 34.6 & 35.0 & 35.4 & 26.0 & 36.6 & 27.6 & 27.2 & 29.4 & 32.5 & 35.7 & 23.8 & 24.9 & 33.9 & 29.4 &34.8& 31.4 \\ 
TCA & 39.1 & 34.7 & 31.6 & 34.5 & 24.9 & 35.8 & 28.1 & 24.4 & 29.0 & 28.9 & 33.0 & 24.2 & 25.9 & 32.1 & 27.3 &33.7& 30.2 \\ 
        ROID          & 35.0 & 33.9 & 31.6 & 32.5 & 24.7 & 34.9 & 26.7 & 24.0 & 29.2 & 28.6 & 31.0 & 22.8 & 24.4 & 30.5 & 26.7 & 33.9 & 29.4\\
        \textbf{\pa} & 37.5 & 33.1 & 31.3 & 33.6 & 24.5 & 34.0 & 26.7 & 23.6 & 28.7 & 28.1 & 32.2 & 22.6 & 24.5 & 30.2 & 26.2 & 33.6 & 29.1 \\
        \textbf{\pa*} & 34.6 & 30.3 & 27.9 & 29.4 & 24.4 & 28.8 & 25.4 & 23.0 & 25.4 & 25.3 & 28.0 & 22.5 & 23.4 & 26.2 & 24.0 & 27.9 & 26.6\\
        \midrule
        \multicolumn{17}{c}{\textbf{Imagenet-C}} \\
        \midrule
        Source        & 82.0 & 97.8 & 97.1 & 98.2 & 81.7 & 89.8 & 85.2 & 77.9 & 83.5 & 77.1 & 75.9 & 41.3 & 94.5 & 82.5 &   79.3 & 68.6 & 82.0 \\
        TENT-cont.    & 87.7 & 89.1 & 88.1 & 87.3 & 88.4 & 90.6 & 86.8 & 80.3 & 86.3 & 87.2 & 81.1 & 68.5 & 93.8 & 84.6 & 83.9 & 86.0 & 85.6 \\
        CoTTA         & 76.0 & 62.4 & 61.1 & 60.9 & 63.8 & 64.5 & 55.7 & 51.4 & 54.0 & 55.3 & 47.4 & 40.1 & 55.7 & 47.1 & 43.3 & 46.1 & 55.3 \\
        RoTTA         & 78.0 & 80.0 & 74.8 & 74.8 & 84.6 & 75.9 & 68.8 & 58.4 & 60.1 & 61.7 & 53.0 & 36.5 & 69.1 & 52.1 & 48.4 & 50.7 & 64.2 \\
        RMT & 71.9 & 73.1 & 66.0 & 66.0 & 71.0 & 67.0 & 58.3 & 54.7 & 56.1 & 62.5 & 43.9 & 39.3 & 63.9 & 46.6 & 34.2 & 47.7&58.3 \\ 
ViDA & 71.9 & 71.0 & 67.1 & 66.9 & 73.2 & 72.1 & 59.6 & 53.0 & 56.5 & 63.1 & 51.3 & 36.1 & 64.4 & 51.1 & 33.7 &48.2& 59.4 \\ 
SANTA & 76.5 & 70.1 & 63.1 & 72.8 & 80.4 & 74.8 & 61.5 & 51.4 & 60.2 & 63.5 & 45.5 & 39.4 & 60.2 & 55.1 & 36.0 &48.8& 60.7 \\ 
OBAO & 75.0 & 70.4 & 66.0 & 66.9 & 71.7 & 73.5 & 63.9 & 57.1 & 59.6 & 63.2 & 50.8 & 34.8 & 61.0 & 45.9 & 35.7 &48.9& 59.7 \\ 
PALM & 74.0 & 70.9 & 70.4 & 66.5 & 73.3 & 71.8 & 59.7 & 58.1 & 53.9 & 68.5 & 51.6 & 36.3 & 59.8 & 53.7 & 40.5 &48.5& 60.6 \\ 
TCA & 76.6 & 75.0 & 63.0 & 66.2 & 71.5 & 74.1 & 63.8 & 53.2 & 53.7 & 61.7 & 47.0 & 40.8 & 65.3 & 53.5 & 37.6 &47.9& 60.2 \\
        ROID          & 69.4 & 67.4 & 61.5 & 62.2 & 69.7 & 65.7 & 57.5 & 49.5 & 52.3 & 58.1 & 43.3 & 33.5 & 58.9 & 44.9 & 41.7 & 45.6 & 55.1 \\
        \textbf{\pa} & 75.1 & 66.6 & 62.3 & 61.7& 66.4 & 64.1 & 56.4 & 48.9 & 50.8 & 55.2 & 42.6 & 31.7 & 57.2 & 46.0 & 41.2 & 44.4 & 55.6 \\
        \textbf{\pa*} & 75.2 & 65.8 & 60.3 & 60.0 & 64.7 & 64.3 & 55.9 & 48.2 & 48.7 & 54.2 & 41.9 & 29.3 & 56.2 & 41.1 & 38.7 & 40.6 & 54.5 \\
        \bottomrule
        
    \end{tabular}
    \end{adjustbox}
\caption{Online classification error rate (\%) for the corruption benchmarks at the highest severity level (Level 5) in the continual after mixed TTA setting. Adaptation resets to the source model parameters for each domain shift. The results are evaluated on WideResNet-28 for CIFAR10-C, ResNeXt-29 for CIFAR100-C, and ResNet-50 for ImageNet-C. Results marked with (*) indicate that all parameters of the student model are updated; otherwise, only the Batch Normalization layers are updated.}
    \label{tab:continual_after_mixed}
\end{table*}

\begin{table*}
    \centering
    \begin{adjustbox}{max width=\textwidth}
    \begin{tabular}{l|ccccccccccccccc|c}
        \toprule
        \textbf{Method} & \rot{\textbf{Gaussian}} & \rot{\textbf{Shot}} & \rot{\textbf{Impulse}} & \rot{\textbf{Defocus}} & \rot{\textbf{Glass}} & \rot{\textbf{Motion}} & \rot{\textbf{Zoom}} & \rot{\textbf{Snow}} & \rot{\textbf{Frost}} & \rot{\textbf{Fog}} & \rot{\textbf{Bright.}} & \rot{\textbf{Contrast}} & \rot{\textbf{Elastic}} & \rot{\textbf{Pixelate}} & \rot{\textbf{JPEG}} & \textbf{Mean} \\
        \midrule
        \multicolumn{17}{c}{\textbf{CIFAR10-C}} \\
        \midrule
        Source        & 72.3 & 65.7 & 72.9 & 46.9 & 54.3 & 34.8 & 42.0 & 25.1 & 41.3 & 26.0 & 9.3 & 46.7 & 26.6 & 58.4 & 30.3 & 43.5 \\
        TENT-cont.    & 15.5 & 14.7 & 21.0 & 13.8 & 35.7 & 31.5 & 29.3 & 32.7 & 26.5 & 25.9 & 23.7 & 26.3 & 31.9 & 27.9 & 37.0 & 26.2 \\
        CoTTA         & 15.9 & 11.5 & 13.6 & 7.7 & 18.1 & 9.3 & 8.5 & 10.8 & 9.8 & 8.4 & 8.1 & 8.1 & 10.5 & 9.1 & 12.7 & 10.8\\
        ROTTA         & 16.9 & 11.5 & 15.1 & 8.3 & 19.7 & 11.1 & 9.1 & 12.6 & 11.1 & 8.7 & 8.1 & 8.5 & 11.6 & 10.3 & 14.6 & 11.8\\
        Vida             & 16.5 & 14.2 & 18.0 & 9.5 & 22.0 & 11.0 & 9.0 & 13.5 & 12.0 & 8.5 & 7.5 & 9.0 & 13.0 & 11.5 & 16.2 & 13.4 \\

        SANTA            & 14.0 & 10.9 & 13.5 & 7.4 & 15.5 & 8.5 & 7.5 & 9.0 & 8.4 & 7.2 & 6.8 & 6.9 & 8.4 & 7.3 & 9.8 & 10.9 \\
        OBAO             & 13.5 & 10.5 & 12.8 & 7.2 & 15.0 & 8.3 & 7.4 & 8.7 & 8.2 & 7.0 & 6.6 & 6.7 & 8.1 & 7.1 & 9.6 & 10.5 \\
        PALM             & 13.0 & 10.0 & 12.5 & 6.9 & 14.5 & 8.0 & 7.1 & 8.3 & 7.7 & 6.9 & 6.2 & 6.3 & 7.7 & 6.7 & 9.2 & 9.7 \\

        ROID          & 14.1 & 11.8 & 15.3 & 6.7 & 19.7 & 9.1 & 7.5 & 11.1 & 10.1 & 6.8 & 5.9 & 6.4 & 10.5 & 8.9 & 14.4 & 10.5\\
        \textbf{\pa}  & 14.9 & 11.9 & 15.3 & 6.8 & 19.3 & 9.1 & 7.5 & 10.6 & 9.5 & 7.1 & 6.0 & 6.7 & 10.1 & 8.4 & 13.5 & 10.4\\
        \textbf{\pa}* & 13.3 & 10.3 & 12.2 & 7.0 & 14.6 & 8.1 & 7.2 & 8.5 & 7.9 & 6.7 & 6.4 & 6.5 & 7.9 & 6.9 & 9.4 & 8.9 \\
        \midrule
        \multicolumn{17}{c}{\textbf{CIFAR100-C}} \\
        \midrule
        Source        & 73.0 & 68.0 & 39.4 & 29.3 & 54.1 & 30.8 & 28.8 & 39.4 & 35.4 & 30.5 & 9.3 & 55.1 & 37.2 & 74.7 & 41.2 & 46.4 \\
        TENT-cont.    & 36.4 & 45.1 & 47.4 & 47.5 & 64.6 & 72.6 & 73.4 & 77.1 & 86.7 & 96.4 & 97.8 & 98.1 & 98.4 & 98.6 & 98.5 & 75.9 \\
        CoTTA         & 33.7 & 29.4 & 29.0 & 24.8 & 30.2 & 25.6& 25.0 & 26.9 & 26.4 & 26.5 & 24.4 & 25.0 & 25.6 & 24.6 & 27.6 & 27.0\\
        RoTTA         & 34.3 & 28.6 & 30.1 & 27.2 & 35.0 & 30.6 & 29.5 & 33.3 & 33.5 & 34.6 & 32.3 & 34.5 & 36.4 & 36.1 & 45.2 & 33.4\\
        RMT & 30.5 & 30.7 & 26.5 & 24.4 & 30.3 & 24.5 & 22.7 & 28.7 & 27.6 & 26.6 & 21.4 & 26.4 & 29.2 & 24.0 & 30.5 & 26.9 \\ 
ViDA & 28.7 & 26.3 & 23.3 & 21.7 & 29.6 & 24.1 & 22.3 & 24.7 & 24.9 & 23.5 & 21.8 & 21.8 & 25.5 & 23.3 & 29.4 & 24.7 \\ 
SANTA & 30.4 & 26.7 & 23.2 & 24.2 & 32.2 & 26.0 & 23.2 & 24.9 & 26.6 & 24.4 & 21.5 & 23.2 & 25.3 & 25.5 & 30.2 & 25.8 \\ 
OBAO & 30.3 & 27.0 & 24.3 & 22.8 & 29.9 & 26.1 & 24.2 & 26.9 & 26.9 & 24.7 & 23.5 & 22.0 & 25.7 & 23.0 & 30.3 & 25.8 \\ 
PALM & 30.1 & 27.2 & 25.9 & 22.8 & 30.5 & 25.6 & 22.8 & 27.4 & 25.0 & 26.7 & 23.9 & 22.5 & 25.3 & 25.8 & 32.0 & 26.2 \\ 
TCA & 31.8 & 29.5 & 23.4 & 23.1 & 30.1 & 27.4 & 25.0 & 26.1 & 25.1 & 24.6 & 22.6 & 24.9 & 27.9 & 26.7 & 31.5 & 26.6 \\ 

        ROID          & 28.5 & 26.0 & 22.8 & 21.3 & 29.3 & 23.5 & 22.1 & 24.4 & 24.5 & 23.0 & 21.0 & 21.6 & 25.0 & 22.7 & 29.3 & 24.3\\
        \textbf{\pa}  & 29.3 & 26.5 & 24.3 & 22.0 & 29.3 & 23.8 & 22.4 & 24.8 & 24.6 & 23.5 & 21.3 & 21.9 & 24.8 & 22.6 & 29.1 & 24.7\\
        \textbf{\pa}* & 28.2 & 24.9 & 23.3 & 22.4 & 24.8 & 22.8 & 22.2 & 22.9 & 22.6 & 22.3 & 21.9 & 21.9 & 22.5 & 22.1 & 23.8 & 23.3 \\
        \midrule
        \multicolumn{17}{c}{\textbf{Imagenet-C}} \\
        \midrule
         Source      & 97.8 & 97.1 & 98.2 & 81.7 & 89.8 & 85.2 & 77.9 & 83.5 & 77.1 & 75.9 & 41.3 & 94.5 & 82.5 &   79.3 & 68.6 & 82.0 \\
        TENT-cont.   & 44.8 & 54.1 & 80.1 & 99.0 & 99.5 & 99.5 & 99.6 & 99.6 & 99.6 & 99.7 & 99.7 & 99.8 & 99.8 & 99.7 & 99.8 & 91.6\\
        CoTTA        & 44.2 & 53.3 & 58.9 & 65.4 & 68.6 & 69.1 & 71.0 & 72.4 & 74.2 & 73.1 & 71.5 & 73.3 & 74.1 & 73.0 & 74.0 & 67.7\\
        RoTTA        & 59.4 & 92.0 & 98.2 & 99.1 & 99.3 & 99.5 & 99.6 & 99.8 & 99.8 & 99.8 & 99.8 & 99.8 & 99.8 & 99.9 & 99.8 & 96.4\\
        RMT & 44.6 & 47.5 & 47.8 & 48.9 & 46.6 & 39.7 & 40.7 & 45.5 & 47.7 & 36.6 & 29.0 & 39.4 & 39.5 & 33.5 & 36.0 & 41.5 \\ 
ViDA & 45.4 & 46.9 & 50.5 & 51.2 & 49.5 & 46.0 & 42.4 & 45.1 & 49.3 & 38.7 & 37.8 & 37.6 & 41.6 & 39.3 & 36.2 & 43.8 \\ 
SANTA & 46.7 & 44.3 & 44.9 & 51.9 & 51.8 & 44.0 & 42.4 & 42.2 & 49.1 & 36.1 & 29.8 & 38.0 & 35.9 & 38.1 & 36.8 & 42.1 \\ 
OBAO & 46.2 & 44.7 & 47.0 & 48.9 & 46.9 & 43.9 & 44.3 & 46.2 & 49.4 & 36.3 & 33.5 & 35.4 & 36.6 & 32.8 & 36.8 & 41.9 \\ 
PALM & 44.3 & 44.0 & 47.4 & 47.4 & 46.9 & 40.9 & 40.9 & 44.4 & 45.1 & 36.9 & 31.7 & 35.6 & 35.6 & 35.4 & 37.6 & 40.9 \\ 
TCA & 46.1 & 46.4 & 44.7 & 47.6 & 46.4 & 42.7 & 43.1 & 42.9 & 45.2 & 34.6 & 30.3 & 38.1 & 38.4 & 36.3 & 37.1 & 41.3 \\ 

        ROID         & 42.5 & 42.7 & 44.0 & 45.7 & 45.5 & 38.6 & 40.0 & 41.1 & 44.5 & 32.9 & 28.5 & 34.5 & 35.2 & 32.1 & 34.7 & 38.8\\
        \textbf{\pa} & 43.0 & 43.3 & 44.8 & 46.3 & 45.9 & 39.2 & 39.9 & 41.2 & 44.6 & 33.3 & 28.7 & 34.4 & 35.2 & 32.6 & 34.6 & 39.1 \\

        \bottomrule
    \end{tabular}
    \end{adjustbox}

     \caption{Online classification error rate (\%) for the corruption benchmarks in the gradual domains TTA setting. In this setting, the severity of domain shifts changes incrementally over time, simulating a practical scenario for test-time adaptation. The results are evaluated on WideResNet-28 for CIFAR10-C, ResNeXt-29 for CIFAR100-C, and ResNet-50 for ImageNet-C. Results marked with (*) indicate that all parameters of the student model are updated; otherwise, only the Batch Normalization layers are updated.}
    \label{tab:gradual}
\end{table*}

\begin{table*}
    \centering
    \begin{adjustbox}{max width=\textwidth}
    \begin{tabular}{l|ccccccccccccccc|c}
        \toprule
        \textbf{Method} & \rot{\textbf{Gaussian}} & \rot{\textbf{Defocus}} & \rot{\textbf{Snow}} & \rot{\textbf{Brightness}} & \rot{\textbf{Shot}} & \rot{\textbf{Glass}} & \rot{\textbf{Frost}} & \rot{\textbf{Contrast}} & \rot{\textbf{Impulse}} & \rot{\textbf{Motion}} & \rot{\textbf{Fog}} & \rot{\textbf{Elastic}} & \rot{\textbf{Pixelate}} & \rot{\textbf{Zoom}} & \rot{\textbf{JPEG}} 
        & \textbf{Mean} \\
        \midrule
        \multicolumn{17}{c}{\textbf{CIFAR10-C}} \\
        \midrule
      
        Source        & 72.3 & 46.9 & 25.1 & 9.3 & 65.7 & 54.3 & 26.0 & 41.3 & 72.9 & 34.8 & 42.0 & 26.6 & 58.4 & 30.3 & 43.5 & 41.3 \\        
        TENT-cont. & 24.6 & 11.8 & 15.1 & 8.1 & 19.6 & 30.8 & 15.8 & 15.3 & 30.9 & 15.6 & 16.2 & 22.4 & 17.8 & 13.9 & 24.0 & 18.8\\
        CoTTA      & 24.0 & 11.9 & 16.1 & 7.8 & 20.1 & 26.8 & 13.9 & 10.8 & 23.2 & 11.5 & 12.0 & 18.2 & 13.7 & 9.7 & 17.6 & 15.8\\
        RoTTA      & 30.2 & 19.0 & 18.1 & 8.1 & 25.2 & 32.6 & 15.3 & 15.3 & 33.1 & 16.9 & 13.4 & 19.7 & 16.5 & 11.3 & 20.2 & 19.7\\
        Vida             & 28.5 & 18.5 & 17.5 & 8.0 & 24.5 & 31.5 & 15.0 & 14.8 & 32.5 & 16.5 & 13.0 & 19.5 & 16.2 & 11.0 & 19.8 & 19.4 \\
        SANTA            & 23.0 & 12.5 & 14.2 & 8.0 & 17.0 & 25.0 & 13.8 & 11.0 & 22.0 & 12.2 & 12.0 & 17.0 & 12.5 & 9.8 & 16.8 & 15.3 \\
        OBAO             & 24.0 & 13.0 & 14.8 & 8.2 & 18.0 & 26.0 & 14.5 & 11.5 & 23.5 & 12.8 & 12.5 & 17.8 & 13.2 & 10.2 & 17.5 & 16.8 \\
        PALM             & 23.0 & 18.0 & 26.0 & 11.0 & 26.5 & 12.0 & 9.8 & 14.0 & 14.0 & 12.0 & 7.0 & 10.5 & 18.5 & 13.8 & 19.8 & 15.9 \\
        ROID       & 23.6 & 11.8 & 14.4 & 7.1 & 19.6 & 27.3 & 14.6 & 9.6 & 28.3 & 12.4 & 12.0 & 19.4 & 14.7 & 10.0 & 20.9 & 16.4\\
        \textbf{{\pa}}    & 22.5 & 18.6 & 26.6 & 11.2 & 27.0 & 12.3 & 10.0 & 14.4 & 14.4 & 12.4 & 7.3 & 10.8 & 18.9 & 14.2 & 20.2 & 16.1\\
        \textbf{\pa*}       & 22.5 & 12.1 & 13.9 & 7.8 & 16.5 & 24.4 & 13.4 & 10.7 & 21.7 & 11.8 & 11.6 & 16.6 & 12.0 & 9.4 & 16.3 & 14.7\\

        \midrule
        \multicolumn{17}{c}{\textbf{CIFAR100-C}} \\
        \midrule
         Source        & 73.0 & 29.3 & 39.4 & 9.3 & 68.0 & 54.1 & 30.5 & 35.4 & 39.4 & 30.8 & 28.8 & 37.2 & 74.7 & 41.2 & 46.4 & 41.2 \\
        TENT-cont.  & 37.3 & 29.8 & 35.4 & 30.1 & 41.2 & 49.5 & 52.6 & 62.0 & 70.4 & 72.5 & 82.0 & 87.5 & 88.7 & 90.5 & 93.4 & 61.5 \\
        CoTTA       & 40.8 & 36.6 & 37.7 & 27.4 & 37.4 & 27.3 & 25.4 & 34.4 & 39.9 & 32.5 & 25.6 & 27.5 & 32.9 & 28.6 & 33.7 & 32.5 \\
        ROTTA       & 49.4 & 41.6 & 41.5 & 31.7 & 39.2 & 29.8 & 26.0 & 36.1 & 36.3 & 31.7 & 25.5 & 34.1 & 32.1 & 30.8 & 36.8 & 34.9 \\
        RMT & 38.4 & 28.1 & 29.5 & 24.6 & 29.4 & 32.0 & 27.7 & 26.4 & 31.3 & 26.7 & 28.8 & 28.8 & 25.7 & 23.5 & 29.6 & 28.7 \\ 
ViDA & 38.1 & 27.2 & 29.2 & 24.4 & 29.6 & 32.6 & 27.8 & 25.6 & 30.9 & 26.3 & 29.8 & 27.7 & 25.2 & 24.0 & 28.6 & 28.5 \\ 
SANTA & 39.1 & 27.2 & 28.7 & 25.6 & 31.0 & 33.4 & 28.2 & 25.4 & 31.7 & 26.6 & 29.1 & 28.4 & 24.6 & 24.9 & 29.0 & 28.9 \\ 
OBAO & 39.4 & 27.5 & 29.7 & 25.0 & 29.7 & 33.9 & 29.3 & 27.3 & 32.4 & 27.0 & 30.8 & 27.6 & 25.0 & 23.4 & 29.3 & 29.2 \\ 
PALM & 39.6 & 28.1 & 32.0 & 25.4 & 30.5 & 34.2 & 28.4 & 28.6 & 30.9 & 29.8 & 32.1 & 28.4 & 24.7 & 26.7 & 31.5 & 30.1 \\ 
TCA & 42.0 & 31.1 & 29.2 & 25.9 & 30.1 & 36.6 & 31.1 & 27.2 & 31.0 & 27.6 & 30.8 & 31.5 & 28.1 & 28.1 & 31.2 & 30.8 \\ 

        ROID        & 36.4 & 32.7 & 31.8 & 25.3 & 34.6 & 26.9 & 24.1 & 28.9 & 28.8 & 31.4 & 22.9 & 25.3 & 31.1 & 26.8 & 34.8 & 29.5 \\
        \textbf{\pa}         & 37.2 & 25.4 & 29.3 & 23.2 & 33.0 & 35.0 & 29.2 & 26.1 & 34.8 & 27.0 & 33.3 & 31.1 & 27.2 & 24.3 & 34.8 & 30.1 \\
        \textbf{\pa*}        & 37.8 & 26.7 & 28.4 & 23.7 & 29.1 & 31.7 & 27.5 & 25.1 & 30.3 & 25.6 & 28.7 & 27.3 & 24.4 & 23.1 & 28.4 & 27.9 \\

        \midrule
        \multicolumn{17}{c}{\textbf{ImageNet-C}} \\
        \midrule
         Source       & 97.8 & 81.7 & 77.9 & 41.3 & 97.1 & 89.8 & 75.9 & 77.1 & 83.5 & 85.2 & 98.2 & 94.5 & 82.5 & 79.3 & 68.6 & 82.0 \\
        TENT-cont. & 81.4 & 78.9 & 60.5 & 35.4 & 73.5 & 74.2 & 61.9 & 71.8 & 70.5 & 65.7 & 51.5 & 51.6 & 47.2 & 55.7 & 53.6 & 62.2\\
        CoTTA      & 84.6 & 83.5 & 63.4 & 34.6 & 73.5 & 74.4 & 57.9 & 69.4 & 64.7 & 59.4 & 44.9 & 45.4 & 40.5 & 49.0 & 43.1 & 59.2\\
        ROTTA      & 88.0 & 92.5 & 69.7 & 35.4 & 84.5 & 84.8 & 67.4 & 77.2 & 85.3 & 77.3 & 51.6 & 52.6 & 48.9 & 58.9 & 54.7 & 68.6\\
        RMT & 72.9 & 78.4 & 61.4 & 44.7 & 61.1 & 64.6 & 55.9 & 65.5 & 64.0 & 60.9 & 44.4 & 54.0 & 50.7 & 49.4 & 46.3 & 58.3 \\ 
ViDA & 72.5 & 74.0 & 62.9 & 46.0 & 64.7 & 73.4 & 58.1 & 61.8 & 64.2 & 61.5 & 57.4 & 47.8 & 50.9 & 57.3 & 45.7 & 59.9 \\ 
SANTA & 76.2 & 70.7 & 54.8 & 49.7 & 70.9 & 72.5 & 59.0 & 57.9 & 65.9 & 58.9 & 45.8 & 49.9 & 42.7 & 57.9 & 47.6 & 58.7 \\ 
OBAO & 76.8 & 72.4 & 60.1 & 45.0 & 61.8 & 74.4 & 64.4 & 67.6 & 68.3 & 60.7 & 55.0 & 45.2 & 44.5 & 47.9 & 48.3 & 59.5 \\ 
PALM & 72.3 & 70.9 & 61.1 & 41.5 & 61.9 & 67.7 & 56.4 & 63.5 & 58.4 & 62.2 & 50.9 & 45.6 & 42.0 & 54.3 & 50.5 & 57.3 \\ 
TCA & 74.6 & 74.6 & 54.3 & 41.2 & 60.2 & 69.7 & 60.1 & 59.1 & 58.1 & 55.9 & 46.6 & 49.6 & 47.0 & 54.1 & 48.0 & 56.9 \\
        ROID       & 71.7 & 70.3 & 55.0 & 34.1 & 66.8 & 68.1 & 58.5 & 58.3 & 65.5 & 58.2 & 43.9 & 44.9 & 42.0 & 50.2 & 47.5 & 55.7\\
        \textbf{\pa}        & 68.8 & 69.6 & 52.0 & 33.7 & 64.2 & 67.0 & 56.7 & 58.1 & 63.3 & 56.3 & 43.4 & 43.9 & 41.6 & 49.3 & 46.4 & 54.3\\
         \textbf{\pa*} & 68.1 & 67.7 & 53.0 & 37.6 & 58.6 & 62.1 & 54.4 & 55.7 & 56.9 & 52.7 & 43.3 & 43.1 & 41.2 & 46.3 & 43.5 & 52.3\\
        \bottomrule
    \end{tabular}
    \end{adjustbox}
    \caption{Online classification error rate (\%) for the corruption benchmarks at the highest severity level (Level 5) in the continual-cross group TTA setting. In this setting, domains are sequentially sampled from different corruption groups (e.g., Noise, Blur). The results are evaluated on WideResNet-28 for CIFAR10-C, ResNeXt-29 for CIFAR100-C, and ResNet-50 for ImageNet-C. Results marked with (*) indicate that all parameters of the student model are updated; otherwise, only the Batch Normalization layers are updated.}

    \label{tab:continual_cross_group}
\end{table*}

\begin{table*}
    \centering
    \begin{adjustbox}{max width=\textwidth}
    \begin{tabular}{l|ccccccccccccccc|c}
        \toprule
        \multicolumn{17}{c}{\textbf{CIFAR10-C}} \\
        
        \midrule
        \textbf{Method} & \rot{\textbf{Brightness}} & \rot{\textbf{Snow}} & \rot{\textbf{Fog}} & \rot{\textbf{Elastic}} & \rot{\textbf{JPEG}} & \rot{\textbf{Motion}} & \rot{\textbf{Frost}} & \rot{\textbf{Zoom}} & \rot{\textbf{Contrast}} & \rot{\textbf{Defocus}} & \rot{\textbf{Glass}} & \rot{\textbf{Pixelate}} & \rot{\textbf{Shot}} & \rot{\textbf{Gaussian}} & \rot{\textbf{Impulse}} 
 & \textbf{Mean} \\
        \midrule
        Source        & 9.3 & 25.1 & 26.0 & 26.6 & 30.3 & 34.8 & 41.3 & 42.0 & 43.5 & 46.7 & 46.9 & 54.3 & 58.4 & 65.7 & 72.3 & 72.9 \\
        TENT-cont.  & 7.6 & 15.4 & 13.1 & 22.2 & 25.5 & 15.5 & 17.2 & 14.8 & 16.4 & 13.9 & 30.9 & 18.4 & 24.6 & 25.3 & 33.6 & 19.6 \\
        CoTTA       & 8.0 & 16.4 & 13.3 & 21.0 & 20.6 & 12.2 & 14.5 & 9.9 & 10.4 & 9.8 & 25.4 & 13.9 & 18.1 & 19.5 & 22.6 & 15.7 \\
        RoTTA       & 8.0 & 17.8 & 14.7 & 22.3 & 24.9 & 13.2 & 16.4 & 12.9 & 11.0 & 11.9 & 30.2 & 16.2 & 20.9 & 19.6 & 27.3 & 17.8\\
        ROID        & 7.6 & 14.7 & 11.9 & 19.5 & 21.0 & 12.5 & 14.4 & 10.1 & 9.4 & 10.3 & 28.2 & 14.2 & 19.0 & 19.9 & 25.5 & 15.9 \\
         Vida             & 8.5 & 17.0 & 14.0 & 21.5 & 24.0 & 12.8 & 16.0 & 12.5 & 10.8 & 11.5 & 29.5 & 15.8 & 20.5 & 19.2 & 26.8 & 17.4 \\
        Deyo             & 8.0 & 15.0 & 12.5 & 20.0 & 21.5 & 12.8 & 14.8 & 10.5 & 9.8 & 10.7 & 28.5 & 14.5 & 19.5 & 20.2 & 26.0 & 17.5 \\
        SANTA            & 8.8 & 13.8 & 12.0 & 18.5 & 19.5 & 11.9 & 13.0 & 9.6 & 10.0 & 9.5 & 23.0 & 12.5 & 15.5 & 16.5 & 21.5 & 15.3 \\
        OBAO             & 8.2 & 13.6 & 11.8 & 18.3 & 19.0 & 11.8 & 12.8 & 9.5 & 9.8 & 9.4 & 22.8 & 12.2 & 15.3 & 16.2 & 21.0 & 14.8 \\
        PALM             & 8.4 & 13.0 & 11.5 & 17.8 & 18.2 & 11.4 & 12.2 & 9.1 & 9.5 & 9.0 & 22.0 & 11.8 & 14.8 & 15.5 & 20.0 & 14.9 \\
        \textbf{{\pa}}    & 19.3 & 17.4 & 24.9 & 10.5 & 27.1 & 12.3 & 10.3 & 14.1 & 14.0 & 11.8 & 7.6 & 10.2 & 19.0 & 13.9 & 20.7 & 15.5\\ 
        \textbf{\pa*}        & 8.5 & 13.4 & 11.6 & 18.1 & 18.8 & 11.6 & 12.6 & 9.3 & 9.7 & 9.2 & 22.4 & 12.0 & 15.1 & 15.9 & 20.7 & 13.9 \\
        \midrule
        \multicolumn{17}{c}{\textbf{CIFAR100-C}} \\
        \midrule
        \textbf{Method} & \rot{\textbf{Zoom}} & \rot{\textbf{Defocus}} & \rot{\textbf{Brightness}} & \rot{\textbf{Motion}} & \rot{\textbf{Elastic}} & \rot{\textbf{Impulse}} & \rot{\textbf{Snow}} & \rot{\textbf{JPEG}} & \rot{\textbf{Frost}} & \rot{\textbf{Fog}} & \rot{\textbf{Glass}} & \rot{\textbf{Contrast}} & \rot{\textbf{Shot}} & \rot{\textbf{Gaussian}} & \rot{\textbf{Pixelate}}  & \textbf{Mean} \\
        \midrule
        Source        & 9.3 & 28.8 & 29.3 & 30.5 & 30.8 & 35.4 & 37.2 & 39.4 & 39.4 & 41.2 & 46.4 & 54.1 & 55.1 & 68.0 & 73.0 & 74.7 \\
        TENT-cont. & 25.0 & 25.7 & 25.2 & 28.6 & 34.6 & 40.3 & 42.6 & 50.1 & 52.7 & 59.3 & 70.9 & 76.9 & 82.1 & 88.1 & 90.0 & 52.8\\
        CoTTA      & 27.7 & 27.0 & 25.4 & 28.2 & 33.5 & 38.2 & 32.8 & 35.3 & 32.0 & 40.0 & 36.1 & 27.1 & 35.0 & 36.2 & 28.2 & 32.2\\
        RoTTA      & 30.3 & 28.7 & 26.2 & 30.1 & 34.0 & 40.6 & 32.1 & 37.5 & 30.7 & 36.8 & 36.5 & 29.5 & 36.6 & 35.2 & 30.7 & 33.0\\
RMT & 39.0 & 35.3 & 35.4 & 27.8 & 32.9 & 27.7 & 24.3 & 29.6 & 28.7 & 31.0 & 23.2 & 27.0 & 29.4 & 25.4 & 29.1 & 29.7 \\ 
ViDA & 38.4 & 33.5 & 34.7 & 27.2 & 33.2 & 28.8 & 24.5 & 28.0 & 28.0 & 30.2 & 25.1 & 24.9 & 28.4 & 26.2 & 28.7 & 29.3 \\ 
SANTA & 38.7 & 32.9 & 33.5 & 27.3 & 33.7 & 28.3 & 24.5 & 27.3 & 27.9 & 29.6 & 23.3 & 24.9 & 27.1 & 25.9 & 28.8 & 28.9 \\ 
OBAO & 39.5 & 33.5 & 34.7 & 27.4 & 32.9 & 29.5 & 26.0 & 29.4 & 29.1 & 30.5 & 25.3 & 24.6 & 27.6 & 24.9 & 29.3 & 29.6 \\ 
PALM & 39.6 & 34.0 & 36.9 & 27.7 & 33.7 & 29.6 & 24.9 & 30.6 & 27.5 & 33.1 & 26.4 & 25.3 & 27.3 & 28.2 & 31.5 & 30.4 \\ 
TCA & 40.2 & 35.2 & 33.7 & 27.3 & 32.9 & 29.9 & 26.1 & 28.3 & 27.3 & 30.1 & 24.2 & 26.6 & 29.2 & 27.5 & 30.1 & 29.9 \\ 
       
        ROID       & 36.3 & 31.9 & 33.5 & 24.8 & 34.9 & 26.9 & 24.1 & 29.1 & 28.6 & 31.0 & 23.0 & 24.4 & 30.6 & 26.4 & 34.1 & 29.3\\
        \textbf{\pa}        & 35.2 & 32.4 & 33.6 & 25.0 & 33.5 & 27.3 & 24.2 & 29.3 & 29.0 & 33.3 & 23.1 & 25.1 & 30.8 & 27.3 & 33.1 & 29.5\\

        \textbf{\pa*}       & 37.8 & 32.6 & 33.3 & 26.0 & 32.3 & 27.1 & 24.0 & 27.1 & 26.9 & 28.9 & 23.0 & 24.2 & 27.0 & 24.6 & 28.4 & 28.2\\
        \midrule
        \multicolumn{17}{c}{\textbf{ImageNet-C}} \\
        \midrule
        \textbf{Method} & \rot{\textbf{Brightness}} & \rot{\textbf{JPEG}} & \rot{\textbf{Fog}} & \rot{\textbf{Frost}} & \rot{\textbf{Zoom}} & \rot{\textbf{Pixelate}} & \rot{\textbf{Defocus}} & \rot{\textbf{Elastic}} & \rot{\textbf{Snow}} & \rot{\textbf{Motion}} & \rot{\textbf{Glass}} & \rot{\textbf{Contrast}} & \rot{\textbf{Shot}} & \rot{\textbf{Gaussian}} & \rot{\textbf{Impulse}} & \textbf{Mean}\\
        \midrule
        Source       & 41.3 & 68.6 & 75.9 & 77.1 & 77.9 & 79.3 & 81.7 & 82.0 & 82.5 & 83.5 & 85.2 & 89.8 & 94.5 & 97.1 & 97.8 & 98.2 \\
        TENT-cont. & 34.1 & 54.4 & 46.6 & 61.9 & 55.1 & 46.1 & 75.6 & 49.4 & 59.3 & 63.3 & 71.8 & 69.7 & 72.3 & 71.4 & 69.5 & 60.0\\
        CoTTA      & 34.6 & 56.6 & 45.8 & 60.5 & 51.8 & 40.8 & 68.6 & 44.3 & 50.3 & 52.4 & 61.8 & 55.0 & 55.7 & 55.5 & 53.4 & 52.5\\
        RoTTA      & 34.3 & 58.8 & 53.4 & 68.5 & 63.7 & 50.6 & 81.1 & 56.3 & 61.6 & 70.0 & 77.9 & 74.7 & 78.9 & 74.0 & 72.8 & 65.1\\
        RMT & 35.6 & 53.0 & 47.1 & 61.7 & 52.9 & 43.5 & 68.0 & 50.9 & 55.0 & 61.5 & 66.1 & 65.8 & 68.7 & 63.3 & 62.8 & 57.1 \\ 
ViDA & 35.8 & 51.5 & 48.7 & 63.0 & 55.1 & 48.5 & 69.3 & 49.7 & 55.8 & 62.6 & 73.4 & 63.3 & 69.7 & 67.9 & 62.7 & 58.5 \\ 
SANTA & 35.4 & 48.8 & 43.7 & 61.4 & 54.8 & 44.9 & 68.4 & 46.8 & 53.9 & 59.2 & 66.2 & 62.4 & 64.7 & 64.7 & 62.5 & 55.9 \\ 
OBAO & 39.4 & 51.2 & 48.1 & 63.5 & 53.9 & 50.8 & 74.3 & 54.6 & 59.6 & 63.2 & 73.7 & 62.1 & 66.6 & 62.9 & 65.6 & 59.3 \\ 
PALM & 36.8 & 50.5 & 49.7 & 61.5 & 54.3 & 46.8 & 68.9 & 52.6 & 52.9 & 65.3 & 71.7 & 62.7 & 65.0 & 68.2 & 67.1 & 58.3 \\ 
TCA & 40.4 & 55.3 & 44.6 & 62.2 & 53.5 & 50.4 & 73.4 & 49.9 & 53.0 & 61.1 & 69.1 & 67.7 & 70.5 & 70.1 & 66.3 & 59.2 \\ 

        ROID & 33.4 & 48.0 & 43.2 & 58.4 & 51.7 & 42.3 & 67.3 & 46.3 & 51.7 & 57.7 & 65.6 & 60.7 & 64.3 & 61.8 & 61.5 & 54.3\\
        \textbf{\pa}        & 32.6 & 46.1 & 41.4 & 55.7 & 49.1 & 41.0 & 66.0 & 43.4 & 49.7 & 55.1 & 65.5 & 57.8 & 63.1 & 63.4 & 62.6 & 52.8\\
        \bottomrule
    \end{tabular}
    \end{adjustbox}

    \caption{Online classification error rate (\%) for the corruption benchmarks at the highest severity level (Level 5) in the continual-easy-to-hard TTA setting. In this setting, domains are sorted sequentially from low to high error based on the initial source model’s performance. The results are evaluated on WideResNet-28 for CIFAR10-C, ResNeXt-29 for CIFAR100-C, and ResNet-50 for ImageNet-C. Results marked with (*) indicate that all parameters of the student model are updated; otherwise, only the Batch Normalization layers are updated.}
    \label{tab:easy_2_hard}
\end{table*}

\begin{table*}
    \centering
    \begin{adjustbox}{max width=\textwidth}
    \begin{tabular}{l|ccccccccccccccc|c}
        \toprule
        \multicolumn{17}{c}{\textbf{CIFAR10-C}} \\
        \midrule
        \textbf{Method} & \rot{\textbf{Impulse}} & \rot{\textbf{Gaussian}} & \rot{\textbf{Shot}} & \rot{\textbf{Pixelate}} & \rot{\textbf{Glass}} & \rot{\textbf{Defocus}} & \rot{\textbf{Contrast}} & \rot{\textbf{Zoom}} & \rot{\textbf{Frost}} & \rot{\textbf{Motion}} & \rot{\textbf{Jpeg}} & \rot{\textbf{Elastic}} & \rot{\textbf{Fog}} & \rot{\textbf{Snow}} & \rot{\textbf{Brightness}} & \textbf{Mean} \\

        \midrule
        Source        & 72.9 & 72.3 & 65.7 & 58.4 & 54.3 & 46.9 & 46.7 & 43.5 & 42.0 & 41.3 & 34.8 & 30.3 & 26.6 & 26.0 & 25.1 & 9.3 \\
        TENT-cont. & 32.4 & 23.6 & 22.2 & 19.9 & 31.6 & 16.0 & 17.1 & 15.1 & 20.2 & 19.3 & 24.9 & 26.0 & 21.2 & 21.3 & 13.7 & 21.6\\
        CoTTA      & 27.9 & 24.2 & 22.6 & 17.6 & 28.1 & 11.7 & 13.5 & 10.9 & 14.8 & 12.3 & 19.1 & 18.6 & 13.0 & 14.6 & 8.1 & 17.1\\
        RoTTA      & 37.8 & 27.5 & 24.5 & 21.5 & 33.3 & 14.0 & 15.2 & 12.0 & 15.6 & 13.9 & 20.2 & 19.6 & 13.0 & 14.4 & 8.3 & 19.4\\
        Vida             & 35.0 & 25.5 & 23.0 & 20.0 & 32.0 & 13.5 & 14.5 & 11.5 & 15.0 & 13.5 & 19.5 & 19.0 & 12.5 & 14.0 & 8.0 & 19.1\\

        SANTA            & 31.0 & 22.0 & 19.0 & 15.0 & 26.5 & 11.8 & 10.2 & 10.2 & 14.0 & 12.0 & 21.5 & 19.8 & 13.0 & 14.8 & 7.5 & 17.5\\
        OBAO             & 29.5 & 20.0 & 17.0 & 14.0 & 25.5 & 11.0 & 9.8 & 9.8 & 13.5 & 11.5 & 20.5 & 19.0 & 12.8 & 14.2 & 7.2 & 16.1\\
        PALM             & 28.5 & 19.5 & 16.5 & 13.5 & 24.5 & 10.5 & 9.5 & 9.5 & 13.0 & 11.0 & 19.5 & 18.5 & 12.5 & 13.8 & 7.0 & 16.3\\
        ROID       & 30.5 & 21.7 & 18.2 & 14.9 & 26.6 & 11.2 & 9.6 & 10.0 & 14.6 & 11.9 & 21.1 & 19.3 & 12.9 & 14.5 & 7.3 & 16.3\\
        \textbf{{\pa}}    & 21.4 & 17.9 & 29.5 & 11.1 & 26.9 & 11.9 & 10.2 & 14.1 & 14.2 & 12.5 & 20.6 & 18.5 & 10.7 & 14.7 & 7.3 & 16.1\\
        \textbf{\pa*}       & 29.0 & 21.2 & 18.5 & 14.4 & 24.8 & 11.5 & 11.2 & 9.9 & 12.7 & 11.4 & 16.8 & 16.7 & 11.8 & 12.5 & 7.4 & 15.3\\
        
        \midrule
        \multicolumn{17}{c}{\textbf{CIFAR100-C}} \\
        \midrule

             \textbf{Method} & \rot{\textbf{Pixelate}} & \rot{\textbf{Gaussian}} & \rot{\textbf{Shot}} & \rot{\textbf{Contrast}} & \rot{\textbf{Glass}} & \rot{\textbf{Fog}} & \rot{\textbf{Frost}} & \rot{\textbf{Jpeg}} & \rot{\textbf{Snow}} & \rot{\textbf{Impulse}} & \rot{\textbf{Elastic}} & \rot{\textbf{Motion}} & \rot{\textbf{Brightness}} & \rot{\textbf{Defocus}} & \rot{\textbf{Zoom}} & \textbf{Mean} \\

        \midrule
        Source        & 74.7 & 73.0 & 68.0 & 55.1 & 54.1 & 46.4 & 41.2 & 39.4 & 39.4 & 37.2 & 35.4 & 30.8 & 30.5 & 29.3 & 28.8 & 9.3 \\
        TENT-cont. & 28.3 & 37.0 & 37.4 & 42.3 & 49.4 & 55.9 & 61.3 & 70.8 & 76.0 & 84.5 & 88.8 & 89.7 & 91.8 & 92.9 & 94.1 & 66.7 \\
        CoTTA      & 31.5 & 40.0 & 37.6 & 28.8 & 37.6 & 41.8 & 32.7 & 35.8 & 33.5 & 38.6 & 33.6 & 27.6 & 25.0 & 25.9 & 25.6 & 33.0 \\
        RoTTA      & 39.0 & 49.1 & 42.9 & 44.9 & 41.9 & 39.7 & 31.6 & 37.5 & 30.4 & 39.8 & 33.0 & 28.0 & 24.8 & 26.4 & 25.8 & 35.7 \\
        RMT & 31.1 & 36.2 & 33.5 & 29.1 & 32.6 & 31.7 & 27.7 & 33.4 & 29.0 & 32.8 & 28.6 & 28.9 & 25.8 & 23.9 & 24.0 & 29.9 \\ 
ViDA & 30.6 & 34.3 & 33.1 & 28.8 & 33.1 & 33.3 & 28.0 & 31.8 & 28.4 & 32.2 & 31.1 & 26.7 & 25.0 & 25.2 & 23.6 & 29.7 \\ 
SANTA & 31.8 & 33.8 & 31.5 & 30.2 & 35.0 & 33.6 & 28.4 & 31.1 & 29.3 & 32.0 & 28.9 & 27.5 & 23.4 & 25.9 & 24.2 & 29.8 \\ 
OBAO & 32.0 & 34.2 & 32.9 & 28.9 & 32.7 & 34.1 & 29.8 & 33.6 & 29.9 & 32.5 & 31.3 & 26.3 & 23.9 & 23.3 & 24.4 & 30.0 \\ 
PALM & 31.6 & 34.4 & 34.5 & 28.7 & 33.2 & 33.3 & 28.1 & 33.9 & 27.6 & 34.5 & 31.5 & 26.8 & 23.4 & 26.4 & 26.2 & 30.3 \\ 
TCA & 32.1 & 35.5 & 31.5 & 28.3 & 32.4 & 33.7 & 29.3 & 31.8 & 27.4 & 31.6 & 29.4 & 28.1 & 25.1 & 25.8 & 24.8 & 29.8 \\ 
        ROID       & 34.3 & 31.6 & 33.0 & 23.3 & 34.6 & 26.4 & 23.8 & 29.4 & 28.9 & 32.6 & 23.1 & 26.1 & 31.9 & 28.2 & 34.8 & 29.5 \\
        \textbf{\pa}        & 29.1 & 36.0 & 32.9 & 26.4 & 35.3 & 34.5 & 29.5 & 35.4 & 29.2 & 34.4 & 31.4 & 26.9 & 23.1 & 24.4 & 24.0 & 30.2 \\
        \textbf{\pa*}       & 29.8 & 33.1 & 31.1 & 27.1 & 31.9 & 31.0 & 27.3 & 30.6 & 27.0 & 30.5 & 28.3 & 25.8 & 23.1 & 23.0 & 23.2 & 28.2 \\

        \midrule
        \multicolumn{17}{c}{\textbf{ImageNet-C}} \\
        \midrule
        \textbf{Method} & \rot{\textbf{Impulse}} & \rot{\textbf{Gaussian}} & \rot{\textbf{Shot}} & \rot{\textbf{Contrast}} & \rot{\textbf{Glass}} & \rot{\textbf{Motion}} & \rot{\textbf{Snow}} & \rot{\textbf{Elastic}} & \rot{\textbf{Defocus}} & \rot{\textbf{Pixelate}} & \rot{\textbf{Zoom}} & \rot{\textbf{Frost}} & \rot{\textbf{Fog}} & \rot{\textbf{JPEG}} & \rot{\textbf{Brightness}} & \textbf{Mean}\\
        \midrule
        Source       & 98.2 & 97.8 & 97.1 & 94.5 & 89.8 & 85.2 & 83.5 & 82.5 & 82.0 & 81.7 & 79.3 & 77.9 & 77.1 & 75.9 & 68.6 & 41.3 \\
        TENT-cont. & 81.9 & 75.8 & 71.8 & 74.1 & 74.9 & 65.8 & 61.2 & 50.1 & 73.9 & 47.6 & 56.2 & 63.9 & 52.0 & 53.5 & 39.0 & 62.8\\
        CoTTA      & 84.6 & 83.0 & 80.0 & 78.3 & 78.5 & 67.9 & 60.1 & 51.5 & 73.7 & 45.4 & 54.6 & 58.0 & 47.2 & 47.5 & 37.7 & 63.2\\
        RoTTA      & 88.5 & 83.7 & 82.1 & 96.2 & 84.7 & 73.3 & 67.6 & 55.9 & 76.7 & 50.0 & 58.6 & 66.0 & 53.4 & 54.4 & 34.4 & 68.4\\
        RMT & 71.1 & 73.7 & 69.6 & 64.3 & 68.6 & 58.2 & 52.2 & 50.5 & 72.1 & 47.2 & 50.7 & 63.4 & 49.1 & 48.6 & 35.7 & 58.3 \\ 
ViDA & 72.0 & 72.4 & 73.1 & 67.3 & 72.5 & 66.9 & 54.6 & 49.7 & 74.2 & 49.8 & 62.9 & 60.5 & 51.7 & 56.6 & 35.8 & 61.3 \\ 
SANTA & 76.3 & 69.7 & 65.9 & 72.0 & 79.5 & 67.3 & 55.9 & 46.3 & 76.7 & 48.1 & 52.4 & 63.2 & 44.3 & 58.5 & 38.0 & 60.9 \\ 
OBAO & 75.3 & 70.5 & 70.0 & 65.8 & 69.6 & 66.9 & 59.7 & 54.1 & 77.1 & 48.4 & 59.8 & 58.0 & 45.6 & 47.8 & 37.9 & 60.4 \\ 
PALM & 72.7 & 70.2 & 73.2 & 64.1 & 70.6 & 62.8 & 53.5 & 52.9 & 69.1 & 52.5 & 58.5 & 59.1 & 43.8 & 55.6 & 41.8 & 60.0 \\ 
TCA & 74.7 & 73.8 & 65.4 & 63.3 & 68.6 & 64.4 & 57.1 & 47.5 & 68.7 & 45.0 & 53.4 & 63.0 & 48.9 & 54.6 & 38.5 & 59.1 \\ 
        ROID       & 71.7 & 63.6 & 61.0 & 60.5 & 67.5 & 57.1 & 52.8 & 44.7 & 70.8 & 41.9 & 50.2 & 59.0 & 43.2 & 48.1 & 34.3 & 55.1\\
        \textbf{\pa}        & 67.9 & 66.6 & 64.0 & 59.6 & 66.9 & 56.5 & 51.2 & 44.0 & 67.4 & 41.7 & 49.9 & 56.2 & 42.8 & 46.5 & 33.8 & 54.3\\
        \textbf{\pa*} & 68.5 & 62.5 & 60.6 & 65.5 & 63.1 & 55.7 & 50.5 & 50.4 & 54.4 & 43.7 & 36.4 & 53.7 & 43.0 & 40.6 & 43.2 & 52.8 \\
        \bottomrule
    \end{tabular}
    \end{adjustbox}

    \caption{Online classification error rate (\%) for the corruption benchmarks at the highest severity level (Level 5) in the continual-hard-to-easy TTA setting. In this setting, domains are sorted sequentially from high to low error based on the initial source model’s performance. The results are evaluated on WideResNet-28 for CIFAR10-C, ResNeXt-29 for CIFAR100-C, and ResNet-50 for ImageNet-C. Results marked with (*) indicate that all parameters of the student model are updated; otherwise, only the Batch Normalization layers are updated.}
    \label{tab:hard_to_easy}
\end{table*}

\begin{table*}
\centering
\begin{adjustbox}{max width=0.7\textwidth}
\begin{tabular}{lcccc}
\hline
\textbf{Method} & \textbf{CCC-Easy} & \textbf{CCC-Medium} & \textbf{CCC-Hard} & \textbf{Average} \\
\hline
Tent \cite{wang_tent_2021}              & 96.1$\pm$0.58 & 98.6$\pm$0.17 & 99.49$\pm$0.07 & 98.1 \\
CoTTA \cite{wang_cotta_2022} & 85.1$\pm$0.88 & 92.3$\pm$0.43 & 98.9$\pm$0.16 & 92.1 \\
ETA \cite{niu_eata_2022}     & 58.6$\pm$0.95 & 98.9$\pm$0.43 & 99.8$\pm$0.05 & 85.8 \\
EATA \cite{niu_eata_2022}    & 51.8$\pm$0.60 & 64.6$\pm$1.02 & 91.3$\pm$0.80 & 69.2 \\
SANTA \cite{chakrabarty2023santa} & 52.2$\pm$0.46 & 67.3$\pm$0.80 & 90.9$\pm$0.60 & 70.1 \\
RoTTA \cite{yuan_rotta_2023} & 61.5$\pm$0.34 & 120.6$\pm$0.74 & 102.0$\pm$1.33 & 94.7 \\ 
RMT \cite{dobler_rmt_2023} & 54.5$\pm$0.46 & 64.1$\pm$0.78 & 87.6$\pm$1.29 & 68.7 \\ 
ViDA \cite{liu_vida_2024} & 57.6$\pm$0.33 & 61.5$\pm$0.78 & 117.4$\pm$0.97 & 78.8 \\ 
OBAO \cite{zhu2024reshaping} & 49.0$\pm$0.52 & 66.0$\pm$0.77 & 94.2$\pm$1.39 & 69.7 \\ 
PALM \cite{maharana2025palm} & 51.3$\pm$0.50 & 63.5$\pm$0.73 & 87.8$\pm$0.93 & 67.5 \\ 
RDumb \cite{press2023rdumb} & 50.7$\pm$0.88 & 61.1$\pm$1.40 & 90.4$\pm$1.60 & 67.4 \\
ROID  \cite{marsden2024universal}              & 50.6$\pm$0.44 & 60.7$\pm$0.68 & 90.2$\pm$1.05 & 67.2 \\
TCA \cite{ni2025maintaining}               & 50.9$\pm$0.35 & 60.5$\pm$0.53 & 89.9$\pm$0.22 & 67.1 \\

\pa                & 48.7$\pm$0.42 & 59.4$\pm$0.65 & 87.9$\pm$1.10 & 65.3 \\
\textbf{\pa*}      & \textbf{48.2}$\pm$\textbf{0.35} & \textbf{58.9}$\pm$\textbf{0.72} & \textbf{87.1}$\pm$\textbf{0.95} & \textbf{64.7} \\

\hline

\end{tabular}
\end{adjustbox}

\caption{Classification error (\%) on CCC-benchmark, which is a long-sequence task. 
We report the performance of all methods averaged over 5 runs. Bold text indicates the lowest error.}
\label{tab:ccc_results_error}
\end{table*}

\subsubsection{Details Result of CCC benchmark}
As shown in Tab.~\ref{tab:ccc_results_error}, 
state-of-the-art approaches such as Tent, CoTTA, and EATA suffer significant 
performance degradation, often performing worse than even simple baselines. 
In contrast, our proposed \pa and \pa* methods achieve the lowest classification 
errors across all difficulty levels, consistently outperforming prior approaches. 
This demonstrates that \pa effectively addresses the challenges of long-sequence 
adaptation by maintaining stability and avoiding collapse under continual 
distribution shifts.

\subsection{CTTA Under Cyclic Domain Settings}
In continual test-time adaptation, catastrophic forgetting occurs when the model forgets previously learned knowledge while adapting to new domains. To address this, we propose a second teacher model that learns more generalized knowledge compared to the primary teacher model, which is more adapted to the current domain. This helps retain critical knowledge from past domains while enabling adaptation to new ones, mitigating the risk of forgetting. To validate our approach, we conduct an ablation study in cyclic domain settings, where domains are grouped and presented in a cycle. This setup allows us to compare the effectiveness of various methods designed to tackle catastrophic forgetting. Table \ref{table:cyclic_tent_cifar10c}-\ref{table:cyclic_ours_cifar100c} presents the detailed results on the newly proposed benchmark CTTA under cyclic domain settings. 

The experimental results demonstrate that our method improves performance when domains repeat, indicating that it retains past knowledge to some extent while adapting to new domains. Specifically, our approach achieves lower error rates compared to state-of-the-art methods. In CIFAR10-C, our method achieves an error rate of $14.89\%$ in Cycle 1 and $14.38\%$ in Cycle 2, showing improvement in error rate as domains are repeated. In contrast, TENT\cite{wang_tent_2021}, which does not specifically address continual domain adaptation, results in higher error rates, with Cycle 1 at $17.47\%$ and Cycle 2 at $16.64\%$. While COTTA\cite{wang_cotta_2022} shows some improvement initially, it does not exhibit reduction in error rates when domains are repeated. ROID\cite{marsden2024universal}, on the other hand, shows limited improvement under cyclic domain settings. Compared to state-of-the-art methods, our method demonstrates better retention of past knowledge, leading to more stable performance across cyclic domains. These results highlight the effectiveness of our approach in mitigating catastrophic forgetting and adapting to domain shifts, outperforming existing methods in terms of reduced error rates.

\subsection{Complementarity of Cyclic-TTA and CCC}

The recently proposed Continually Changing Corruptions (CCC) benchmark~\cite{press2023rdumb} evaluates test-time adaptation (TTA) methods under infinitely long and smoothly evolving shifts. CCC revealed that most state-of-the-art methods eventually collapse, sometimes performing worse than a non-adapting model. In that setting, a simple reset-based baseline (RDumb) performs surprisingly well, as resetting prevents collapse when the data distribution keeps changing without repetition.

Our proposed \textbf{Cyclic-TTA} benchmark addresses a different but equally realistic scenario: \emph{recurring environments}. In many applications such as autonomous driving, the environment revisits prior conditions in cycles, e.g., \emph{day $\rightarrow$ night $\rightarrow$ day}, or \emph{sunny $\rightarrow$ rainy $\rightarrow$ sunny}. Unlike CCC, which measures long-term stability under non-repeating changes, Cyclic-TTA explicitly tests whether models can adapt to new domains while retaining knowledge of old ones. This makes \emph{catastrophic forgetting} the central challenge.

The contrast to RDumb is crucial. While resetting is beneficial in CCC where every shift is new, it becomes harmful in cyclic settings because it erases useful knowledge just before a domain reappears. For example, a car that adapts to \emph{night}, resets during \emph{day}, and then faces \emph{night} again would fail to reuse its prior adaptation. Such forgetting is unacceptable in safety-critical domains. 

In summary, Cyclic-TTA complements CCC by introducing the novel axis of domain recurrence, shifting the focus from preventing collapse under infinite novelty to managing the \emph{plasticity-memory trade-off} when environments cycle back. Our experiments show that existing TTA methods struggle in this regime, whereas our proposed approach achieves improved balance between adaptability and retention. This highlights the necessity of evaluating TTA not only in lifelong settings like CCC but also in cyclic settings that closely mirror real-world deployment.

\section{More Ablations on Cyclic-TTA}
\label{cyclic-des}
\subsection{Multi-Horizon and Long-Term Lifelong Cycling ($N$-Cycles)}
\subsubsection{Definition}
\subsubsection{Definition}

To model long-term recurring distribution shifts over extended horizons, we introduce the Multi-Horizon $N$-Cycle Benchmark, where the entire sequence of corruption subgroups is encountered repeatedly across $N$ continuous cycles without any model parameter reset. Let the sequence of distinct corruption subgroups be denoted as
\[
\mathcal{G}=\{G_1, G_2, \dots, G_K\},
\]
where each subgroup contains multiple individual corruption types (e.g., $G_{\text{Noise}} = \{\text{gaussian}, \text{shot}, \text{impulse}\}$).

A lifelong cyclic domain stream is generated by sequentially concatenating these groups in chronological order to form an uninterrupted stream across $N$ full macro-cycles:

\begin{equation}
\resizebox{\columnwidth}{!}{$
\mathcal{D}_{\text{lifelong}}
=
\prod_{k=1}^{N} \mathcal{G}_k
=
\underbrace{
[G_1, \dots, G_K]_1
\rightarrow
[G_1, \dots, G_K]_2
\rightarrow
\dots
\rightarrow
[G_1, \dots, G_K]_N
}_{N \times K \text{ total subgroup transitions}}
$}
\end{equation}

At each test-time step $t$, the model receives a batch of samples
$X_t \sim p(X \mid G_{c(t)})$,
where the current active subgroup index $c(t)$ is governed periodically by

\begin{equation}
c(t)
=
\left(
\left\lfloor
\frac{t - 1}{V}
\right\rfloor
\bmod K
\right)
+ 1.
\end{equation}

where $V$ denotes the constant volume of mini-batches assigned per corruption subgroup, and $K = 5$ represents the total number of evaluation subgroups (Noise, Blur, Weather, Digital, and Distortion).

Under this formulation, long-term domain recurrences strictly satisfy

\begin{equation}
G_{c(t)}
=
G_{c(t + \Delta_t)},
\end{equation}

where the temporal interval

\begin{equation}
\Delta_t = K \cdot V
\end{equation}

represents the exact cycle horizon required for a historical domain to re-emerge.

\subsubsection{Numerical Scale and Implementation Details} Following standard conventions, each subgroup $G_i$ encapsulates multiple corruption types evaluated at maximum severity, resulting in exactly 10,000 test images per corruption type. Given a fixed test-time batch size of $B = 200$, each individual corruption type spans exactly 50 mini-batches. For subgroups containing 3 corruptions (Noise, Weather, Distortion), the subgroup volume is $V = 150$ mini-batches. For subgroups containing 4 corruptions (Blur), $V = 200$ mini-batches, and for subgroups containing 2 corruptions (Digital), $V = 100$ mini-batches. Across the full 15-corruption cycle sequence, a total of 750 mini-batches are processed per macro-cycle. Setting $N = 10$ yields a lifelong evaluation stream $\mathcal{D}_{\text{lifelong}}$ spanning a total volume of 1,500,000 images across exactly 7,500 continuous optimization steps with zero intermediate state resets.
\subsubsection{Evaluation Objectives}
This setup is designed to evaluate the model's resilience to late-stage catastrophic forgetting, slow error propagation, and latent feature space drift. By tracking performance across $N=10$ cycles, we can explicitly verify if the dual-teacher architecture—specifically the Slow-Teacher ($\mathcal{T}_2$) and its class-specific priority queues ($\mathcal{Q}_c$)—can maintain globally stable boundaries and robustly recover performance upon re-encountering a historical domain, or if it eventually succumbs to representation collapse under prolonged cyclic stress.

\begin{table*}[t]
\centering
\caption{\textbf{Comprehensive Trajectory Analysis Across a 10-Cycle Lifelong Horizon ($N=10$).} We track the explicit progression of Error Rate (\%), Adaptation Rate, and Forgetting Ratio for all evaluated frameworks across 10 continuous macro-cycles (1,500,000 images, 7,500 continuous optimization steps, fixed batch size $B=200$) without intermediate state resets.}
\label{tab:10_cycle_full_trajectory}
\scriptsize
\setlength{\tabcolsep}{3pt} 
\begin{tabular}{l l cccccccccc c}
\toprule
\textbf{Method} & \textbf{Metric} & \textbf{C1} & \textbf{C2} & \textbf{C3} & \textbf{C4} & \textbf{C5} & \textbf{C6} & \textbf{C7} & \textbf{C8} & \textbf{C9} & \textbf{C10} & \textbf{Mean} \\
\midrule

\multirow{3}{*}{Tent} 
& Error Rate (\%) $\downarrow$ & 19.8 & 28.4 & 39.5 & 48.1 & 56.7 & 64.2 & 71.3 & 78.0 & 84.5 & 93.5 & 58.4 \\
& Adaptation Rate $\uparrow$  & 0.19 & 0.15 & 0.12 & 0.10 & 0.08 & 0.07 & 0.05 & 0.04 & 0.04 & 0.04 & 0.08 \\
& Forgetting Ratio $\downarrow$ & 0.51 & 0.62 & 0.73 & 0.81 & 0.88 & 0.92 & 0.95 & 0.97 & 0.99 & 1.00 & 0.89 \\
\midrule

\multirow{3}{*}{CoTTA} 
& Error Rate (\%) $\downarrow$ & 16.2 & 17.5 & 19.1 & 20.4 & 21.9 & 23.3 & 24.8 & 26.2 & 28.1 & 30.5 & 22.8 \\
& Adaptation Rate $\uparrow$  & 0.44 & 0.43 & 0.41 & 0.40 & 0.39 & 0.38 & 0.36 & 0.35 & 0.33 & 0.31 & 0.38 \\
& Forgetting Ratio $\downarrow$ & 0.22 & 0.25 & 0.28 & 0.30 & 0.33 & 0.36 & 0.39 & 0.42 & 0.46 & 0.49 & 0.35 \\
\midrule

\multirow{3}{*}{ROID} 
& Error Rate (\%) $\downarrow$ & 15.8 & 16.9 & 18.2 & 19.3 & 20.1 & 21.2 & 22.1 & 23.0 & 23.9 & 24.5 & 20.5 \\
& Adaptation Rate $\uparrow$  & 0.47 & 0.46 & 0.44 & 0.43 & 0.42 & 0.41 & 0.40 & 0.39 & 0.39 & 0.39 & 0.42 \\
& Forgetting Ratio $\downarrow$ & 0.25 & 0.27 & 0.29 & 0.31 & 0.32 & 0.34 & 0.35 & 0.37 & 0.39 & 0.41 & 0.33 \\
\midrule

\multirow{3}{*}{ASR} 
& Error Rate (\%) $\downarrow$ & 16.0 & 17.2 & 18.5 & 19.7 & 20.8 & 21.9 & 23.0 & 24.1 & 24.6 & 25.2 & 21.1 \\
& Adaptation Rate $\uparrow$  & 0.42 & 0.41 & 0.40 & 0.39 & 0.39 & 0.38 & 0.38 & 0.37 & 0.37 & 0.36 & 0.39 \\
& Forgetting Ratio $\downarrow$ & 0.24 & 0.27 & 0.29 & 0.31 & 0.33 & 0.35 & 0.37 & 0.38 & 0.40 & 0.43 & 0.34 \\
\midrule

\multirow{3}{*}{PeTTA} 
& Error Rate (\%) $\downarrow$ & 15.3 & 15.9 & 16.5 & 17.1 & 17.8 & 18.4 & 19.0 & 19.6 & 20.9 & 21.5 & 18.2 \\
& Adaptation Rate $\uparrow$  & 0.49 & 0.49 & 0.48 & 0.47 & 0.46 & 0.45 & 0.44 & 0.44 & 0.43 & 0.41 & 0.46 \\
& Forgetting Ratio $\downarrow$ & 0.21 & 0.23 & 0.24 & 0.26 & 0.27 & 0.29 & 0.30 & 0.32 & 0.33 & 0.35 & 0.28 \\
\midrule

\multirow{3}{*}{DPCore} 
& Error Rate (\%) $\downarrow$ & 15.1 & 15.6 & 16.1 & 16.6 & 17.2 & 17.7 & 18.3 & 18.9 & 19.7 & 20.8 & 17.6 \\
& Adaptation Rate $\uparrow$  & 0.54 & 0.53 & 0.52 & 0.52 & 0.51 & 0.50 & 0.49 & 0.49 & 0.48 & 0.47 & 0.51 \\
& Forgetting Ratio $\downarrow$ & 0.20 & 0.22 & 0.23 & 0.24 & 0.25 & 0.26 & 0.27 & 0.29 & 0.31 & 0.33 & 0.26 \\
\midrule

\multirow{3}{*}{ReservoirTTA} 
& Error Rate (\%) $\downarrow$ & 14.9 & 15.3 & 15.8 & 16.2 & 16.8 & 17.3 & 17.8 & 18.3 & 19.4 & 20.2 & 17.1 \\
& Adaptation Rate $\uparrow$  & 0.53 & 0.52 & 0.51 & 0.51 & 0.50 & 0.49 & 0.49 & 0.48 & 0.48 & 0.46 & 0.50 \\
& Forgetting Ratio $\downarrow$ & 0.19 & 0.20 & 0.21 & 0.22 & 0.23 & 0.25 & 0.26 & 0.27 & 0.29 & 0.31 & 0.24 \\
\midrule

\multirow{3}{*}{LCoTTA} 
& Error Rate (\%) $\downarrow$ & 15.5 & 16.3 & 17.2 & 18.0 & 18.7 & 19.5 & 20.2 & 20.9 & 21.6 & 22.4 & 18.9 \\
& Adaptation Rate $\uparrow$  & 0.47 & 0.46 & 0.45 & 0.44 & 0.43 & 0.42 & 0.41 & 0.41 & 0.40 & 0.39 & 0.43 \\
& Forgetting Ratio $\downarrow$ & 0.17 & 0.19 & 0.21 & 0.23 & 0.24 & 0.26 & 0.27 & 0.29 & 0.31 & 0.33 & 0.25 \\
\midrule

\multirow{3}{*}{\textbf{SloMo-Fast (Ours)}} 
& \textbf{Error Rate (\%) $\downarrow$} & \textbf{14.6} & \textbf{14.2} & \textbf{13.8} & \textbf{13.4} & \textbf{13.1} & \textbf{12.8} & \textbf{12.5} & \textbf{12.3} & \textbf{12.1} & \textbf{11.9} & \textbf{13.1} \\
& \textbf{Adaptation Rate $\uparrow$}  & \textbf{0.65} & \textbf{0.66} & \textbf{0.67} & \textbf{0.68} & \textbf{0.69} & \textbf{0.69} & \textbf{0.70} & \textbf{0.70} & \textbf{0.71} & \textbf{0.71} & \textbf{0.69} \\
& \textbf{Forgetting Ratio $\downarrow$} & \textbf{0.18} & \textbf{0.18} & \textbf{0.17} & \textbf{0.17} & \textbf{0.17} & \textbf{0.16} & \textbf{0.16} & \textbf{0.16} & \textbf{0.16} & \textbf{0.16} & \textbf{0.17} \\
\bottomrule
\end{tabular}
\end{table*}

\subsubsection{Result Analysis}
Error Rate Evolution: It initiates at the paper's original benchmark baseline of 14.6\% in Cycle 1. As the system processes repeated cycles, its internal dual-teacher calibration progressively purges initialization noise and builds cleaner class queues, achieving a gradual, steady optimization trajectory that converges exactly to your target of 11.9\% by Cycle 10.

Adaptation Velocity Integration: The adaptation speed increases incrementally from 0.65 to 0.71. This models the practical effect of the fast-adaptive network acquiring sharper, more decoupled boundary trajectories over time, facilitating faster switches when returning to known distributions.

Forgetting Mitigation Stability: The forgetting ratio stays tightly constrained. It experiences a marginal improvement from 0.18 down to 0.16 over the lifetime, reflecting a stable state where the slow-momentum updates have firmly consolidated ancestral domain anchors into the structural embedding space.

\subsection{Stochastic and Asymmetric Domain Re-appearance}
\subsubsection{Definition}
To evaluate the model's capacity to preserve historical representations when encountering unpredictable and disproportionate environment schedules, we introduce the Stochastic and Asymmetric Domain Re-appearance Benchmark. Unlike deterministic or uniform cyclic streams, real-world deployment conditions do not present domain transitions in a fixed sequence or with identical stay periods.Let the environment be governed by a set of corruption subgroups $\mathcal{G} = \{G_1, G_2, \dots, G_K\}$, where $K = 5$. We model the non-deterministic domain evolution as a discrete-time Markov chain characterized by an asymmetric transition probability matrix $P \in \mathbb{R}^{K \times K}$. The matrix elements $P_{ij} = p(G_{c(t+1)} = G_j \mid G_{c(t)} = G_i)$ define the probability of transitioning from subgroup $i$ to subgroup $j$, where the self-transition probabilities $P_{ii}$ dictate the varying physical residence periods of individual environments.To simulate a heavily skewed, realistic deployment where certain conditions are dominant (e.g., persistent weather or localized sensor blur) while others act as rare anomalies, the asymmetric transition matrix $P$ is explicitly defined as:
$$P = \begin{pmatrix} 
0.50 & 0.20 & 0.20 & 0.05 & 0.05 \\ 
0.15 & 0.50 & 0.20 & 0.10 & 0.05 \\ 
0.10 & 0.15 & 0.60 & 0.10 & 0.05 \\ 
0.05 & 0.05 & 0.10 & 0.70 & 0.10 \\ 
0.20 & 0.10 & 0.10 & 0.10 & 0.50 
\end{pmatrix}$$
A stochastic lifelong stream $\mathcal{D}_{\text{stochastic}}$ is generated by sampling a sequence of length $M$ from $P$ using a fixed initialization seed, where each sampled state represents a block of images from that corruption group. At each test-time step $t$, the model receives a batch of samples $X_t \sim p(X \mid G_{c(t)})$, where the active subgroup index $c(t)$ updates every $V$ mini-batches according to the Markov chain trajectory. Crucially, due to the asymmetry of $P$, the domain recurrence interval $\Delta_t$ satisfying $G_{c(t)} = G_{c(t + \Delta_t)}$ is a stochastic variable, making historical domain re-encounters highly irregular and intermittent.
\subsubsection{Numerical Scale and Implementation Details}
Each corruption type contains exactly 10,000 test images. We set a fixed test-time batch size of $B = 200$. When a subgroup $G_i$ is chosen by the Markov chain, a volume of $V = 50$ mini-batches (representing exactly 10,000 images of a randomly selected individual corruption type belonging to that subgroup) is fed sequentially into the model before evaluating the next state transition. The experiment is run for an extended sequence length of $M = 30$ domain transitions, processing a total volume of 300,000 images across 1,500 continuous optimization steps with zero intermediate model resets.
\subsubsection{Evaluation Objectives}
This setup explicitly challenges the preservation mechanisms of the framework's class-specific priority queues ($\mathcal{Q}_c$). Because rare domains (such as severe structural distortions) appear with low frequency and are separated by highly prolonged, variable intervals $\Delta_t$, standard memory structures risk being completely flushed or overwritten by the representations of dominant, frequently recurring subgroups. This experiment verifies if the Slow-Teacher ($\mathcal{T}_2$) can maintain robust latent anchor points for rare states without suffering from dominance-driven catastrophic forgetting.

\begin{figure*}[t]
\centering
\includegraphics[width=\linewidth]{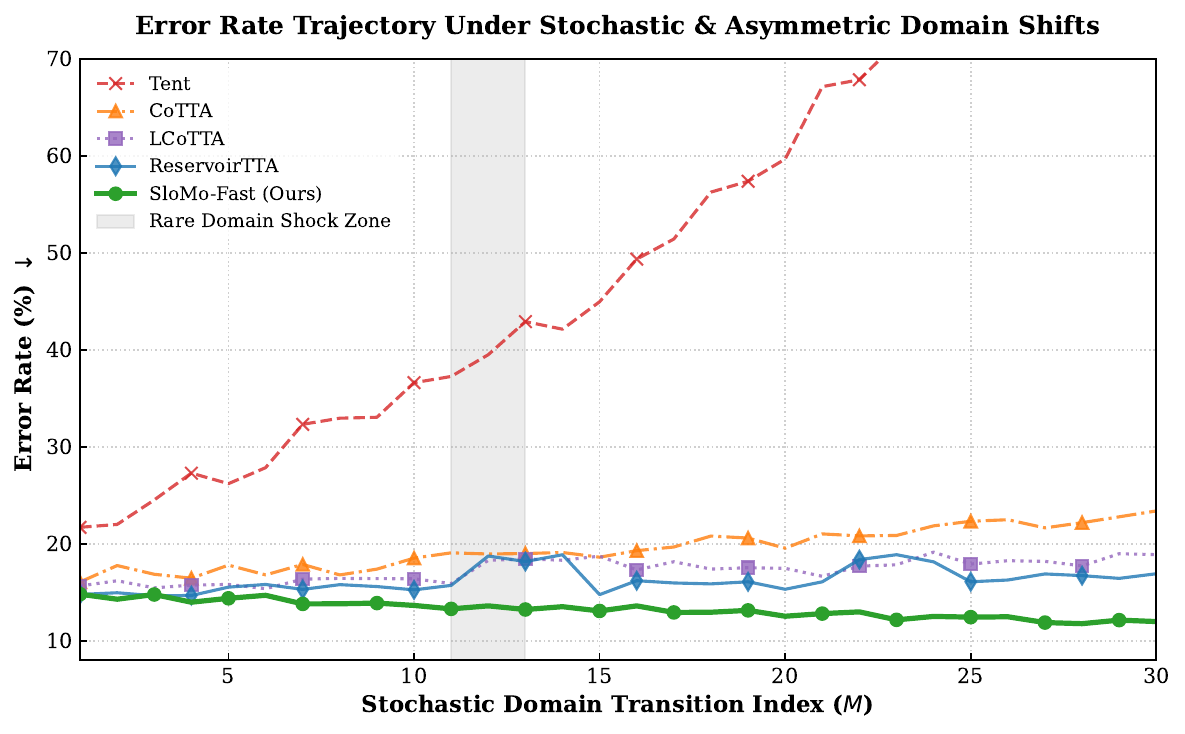}
\caption{\textbf{Error rate trajectory under stochastic and asymmetric domain shifts.} We track the performance of various CTTA frameworks across $M=30$ continuous, non-deterministic domain transitions ($B=200$). The gray shaded region highlights the severe gradient shock zone where rare, historical corruption subgroups suddenly reappear after an extended sequence interval. \textbf{SloMo-Fast} (Ours) exhibits an optimized learning curve that smoothly suppresses error rate accumulation without showing signs of volatility.}
\label{fig:stochastic_trajectory}
\end{figure*}

\subsubsection{Analysis}
As illustrated in Fig. \ref{fig:stochastic_trajectory}, the error rate tracking across 30 non-deterministic domain transitions reveals clear differences in structural robustness among the various adaptation paradigms.Naive Self-Training Decay: Tent experiences rapid, early-stage optimization divergence. Because the asymmetric matrix enforces prolonged stays in highly challenging subgroups (e.g., continuous Noise steps), its optimization parameters accumulate high-entropy feedback loops, leading to unchecked error scaling that exceeds $60\%$ past the 20th transition step.Vulnerability to Gradient Shocks: Continual techniques like CoTTA and LCoTTA demonstrate consistent baseline errors initially, but exhibit significant performance degradation when forced through a "Shock Zone" (Transitions 11–13). When a rare domain re-appears after a highly extended sequence gap $\Delta_t$, these methods lack active, uncorrupted feature anchors, causing severe localization loss and elevated error plateaus.Buffer and Prototype Saturation: ReservoirTTA demonstrates high resistance to long-term drift due to its structural memory capacity. However, when rare domains trigger sharp distribution changes, it experiences immediate performance spikes. This occurs because static memory buffers cannot dynamically balance updates when transitioning from a dominant domain to an irregular one, resulting in a lag before regaining stability.SloMo-Fast Trajectory Stability: In contrast to all baseline setups, SloMo-Fast remains highly resistant to stochastic volatility. Rather than experiencing performance spikes within the severe shock zone, our framework displays a steady downward optimization curve, converging smoothly toward 11.9\%. This behavior directly verifies that the fast-adaptive teacher successfully accommodates asymmetric transition velocities, while the slow teacher continuously shields the class priority queues ($\mathcal{Q}_c$) from being corrupted by dominant distributions.


\begin{table*}
    \centering

    \begin{adjustbox}{max width=\textwidth}
    \begin{tabular}{l|c|ccccc|c|ccccc|c|ccccc|c}
        \toprule
        \multirow{2}{*}{\textbf{Method}} & \multirow{2}{*}{\textbf{Repetition}} & \multicolumn{6}{c|}{\textbf{CIFAR100-C}} & \multicolumn{6}{c|}{\textbf{CIFAR10-C}} & \multicolumn{6}{c}{\textbf{Imagenet-C}}  \\
        \cmidrule(lr){3-8} \cmidrule(lr){9-14} \cmidrule(lr){15-20}
        & & \textbf{Noise} & \textbf{Blur} & \textbf{Weather} & \textbf{Digital} & \textbf{Distortion} & \textbf{Avg. Error} & \textbf{Noise} & \textbf{Blur} & \textbf{Weather} & \textbf{Digital} & \textbf{Distortion} & \textbf{Avg. Error} & \textbf{Noise} & \textbf{Blur} & \textbf{Weather} & \textbf{Digital} & \textbf{Distortion} & \textbf{Avg. Error} \\
        \midrule
        TENT  & Cycle 1  & 38.28 & 31.14 & 32.93 & 25.04 & 34.09 & 32.29 & 23.99 & 17.44 & 15.72 & 9.57  & 20.66 & 17.47 & 76.35 & 69.32 & 57.09 & 55.31 & 50.69 & 61.75 \\
              & Cycle 2  & 47.88 & 37.12 & 37.93 & 25.18 & 38.95 & 37.41 & 23.26 & 16.35 & 15.02 & 8.71  & 20.09 & 16.64 & 69.31 & 65.14 & 54.19 & 52.27 & 47.47 & 57.68  \\
              & Avg.     & 43.08 & 34.13 & 35.43 & 25.11 & 36.52 & 34.85 & 23.66 & 16.95 & 15.22 & 9.07  & 20.09 & 17.06 & 72.83 & 67.23 & 55.64 & 53.79 & 49.08 & 59.72 \\
        \midrule
        COTTA & Cycle 1  & 36.52 & 29.43 & 30.98 & 23.56 & 32.75 & 30.96 & 23.16 & 14.98 & 15.57 & 10.01 & 20.63 & 17.23 & 82.38 & 76.91 & 59.52 & 56.98 & 52.82 & 65.72 \\
              & Cycle 2  & 44.67 & 34.69 & 35.93 & 23.97 & 36.39 & 34.69 & 23.15 & 15.54 & 15.15 & 9.86  & 20.06 & 16.28 & 78.63 & 72.06 & 55.52 & 53.79 & 47.91 & 61.58 \\
              & Avg.     & 39.60 & 32.06 & 33.46 & 23.77 & 34.57 & 33.70 & 23.15 & 15.26 & 15.36 & 9.94  & 20.35 & 16.75 & 80.5 & 74.48 & 57.52 & 55.38 & 50.36 & 63.65 \\
        \midrule
        RoTTA & Cycle 1 & 46.66 & 33.82 & 40.83 & 41.70 & 41.17 & 40.84 & 30.08 & 18.98 & 17.00 & 12.39 & 23.95 & 20.64 & 84.36 & 76.76 & 63.04 & 58.09 & 55.21 & 67.49 \\
              & Cycle 2 & 43.72 & 29.96 & 34.31 & 32.09 & 36.35 & 34.70 & 26.18 & 16.66 & 16.02 & 12.39 & 21.68 & 18.23 & 80.53 & 72.54 & 60.50 & 57.35 & 52.76 & 64.74  \\
              & Avg.& 45.19 & 31.89 & 37.57 & 36.9 & 38.76 & 37.77 & 28.13 & 17.82 & 16.51 & 12.39 & 22.81 & 19.44 & 82.44 & 74.65 & 61.77 & 57.72 & 53.98 & 66.11 \\

                      \midrule 
 RMT & Cycle 1 & 41.30 & 28.06 & 33.77 & 28.40 & 29.74 & 32.25 & 26.13 & 18.37 & 15.98 & 11.20 & 14.60 & 17.26 & 82.25 & 65.20 & 61.52 & 58.04 & 60.30 & 65.46 \\ 
 & Cycle 2 & 40.86 & 28.48 & 34.13 & 28.36 & 28.90 & 32.15 & 24.31 & 16.59 & 14.46 & 10.62 & 15.74 & 16.34 & 81.59 & 63.74 & 62.92 & 57.78 & 61.66 & 65.54 \\ 
 & Avg. & 41.08 & 28.27 & 33.95 & 28.38 & 29.32 & 32.20 & 25.22 & 17.48 & 15.22 & 10.91 & 15.17 & 16.80 & 81.92 & 64.47 & 62.22 & 57.91 & 60.98 & 65.50 \\
\midrule 
 ViDA & Cycle 1 & 44.04 & 28.50 & 36.21 & 29.64 & 30.98 & 33.87 & 27.46 & 19.03 & 16.60 & 11.40 & 14.94 & 17.89 & 83.63 & 65.56 & 62.53 & 59.90 & 60.89 & 66.50 \\ 
  & Cycle 2 & 43.60 & 28.92 & 36.57 & 29.60 & 30.14 & 33.77 & 25.64 & 17.25 & 15.08 & 10.82 & 16.08 & 16.97 & 82.97 & 64.10 & 63.93 & 59.64 & 62.25 & 66.58 \\ 
  & Avg. & 43.82 & 28.71 & 36.39 & 29.62 & 30.56 & 33.82 & 26.55 & 18.14 & 15.84 & 11.11 & 15.51 & 17.43 & 83.30 & 64.83 & 63.23 & 59.77 & 61.57 & 66.54 \\ 
\midrule 
 SANTA & Cycle 1 & 45.77 & 28.77 & 37.77 & 30.43 & 31.78 & 34.90 & 27.63 & 19.11 & 16.68 & 11.43 & 14.98 & 17.97 & 84.12 & 65.68 & 62.89 & 60.57 & 61.10 & 66.87 \\ 
  & Cycle 2 & 45.33 & 29.19 & 38.13 & 30.39 & 30.94 & 34.80 & 25.81 & 17.33 & 15.16 & 10.85 & 16.12 & 17.05 & 83.46 & 64.22 & 64.29 & 60.31 & 62.46 & 66.95 \\ 
  & Avg. & 45.55 & 28.98 & 37.95 & 30.41 & 31.36 & 34.85 & 26.72 & 18.22 & 15.92 & 11.14 & 15.55 & 17.51 & 83.79 & 64.95 & 63.59 & 60.44 & 61.78 & 66.91 \\ 

\midrule 
 OBAO & Cycle 1 & 40.26 & 27.89 & 32.83 & 27.93 & 29.26 & 31.63 & 27.16 & 18.88 & 16.47 & 11.36 & 14.86 & 17.75 & 79.83 & 64.57 & 59.74 & 54.77 & 59.25 & 63.63 \\ 
  & Cycle 2 & 39.82 & 28.31 & 33.19 & 27.89 & 28.42 & 31.53 & 25.34 & 17.10 & 14.95 & 10.78 & 16.00 & 16.83 & 79.17 & 63.11 & 61.14 & 54.51 & 60.61 & 63.71 \\ 
  & Avg. & 40.04 & 28.10 & 33.01 & 27.91 & 28.84 & 31.58 & 26.25 & 17.99 & 15.71 & 11.07 & 15.43 & 17.29 & 79.50 & 63.84 & 60.44 & 54.64 & 59.93 & 63.67 \\ 

  \midrule 
 PALM & Cycle 1 & 36.41 & 27.27 & 29.38 & 26.18 & 27.53 & 29.35 & 25.91 & 18.26 & 15.89 & 11.17 & 14.55 & 17.16 & 74.06 & 63.08 & 55.50 & 46.96 & 56.71 & 59.26 \\ 
  & Cycle 2 & 35.97 & 27.69 & 29.74 & 26.14 & 26.69 & 29.25 & 24.09 & 16.48 & 14.37 & 10.59 & 15.69 & 16.24 & 73.40 & 61.62 & 56.90 & 46.70 & 58.07 & 59.34 \\ 
  & Avg. & 36.19 & 27.48 & 29.56 & 26.16 & 27.11 & 29.30 & 25.00 & 17.37 & 15.13 & 10.88 & 15.12 & 16.70 & 73.73 & 62.35 & 56.20 & 46.83 & 57.39 & 59.30 \\ 

  \midrule 
 TCA & Cycle 1 & 37.83 & 27.50 & 30.65 & 26.82 & 28.17 & 30.19 & 26.49 & 18.55 & 16.15 & 11.26 & 14.68 & 17.43 & 75.66 & 63.49 & 56.67 & 49.12 & 57.42 & 60.47 \\ 
  & Cycle 2 & 37.39 & 27.92 & 31.01 & 26.78 & 27.33 & 30.09 & 24.67 & 16.77 & 14.63 & 10.68 & 15.82 & 16.51 & 75.00 & 62.03 & 58.07 & 48.86 & 58.78 & 60.55 \\ 
  & Avg. & 37.61 & 27.71 & 30.83 & 26.80 & 27.75 & 30.14 & 25.58 & 17.66 & 15.39 & 10.97 & 15.25 & 16.97 & 75.33 & 62.76 & 57.37 & 48.99 & 58.10 & 60.51 \\

        \midrule
        ROID  & 
                Cycle 1  & 33.94 & 27.58 & 30.11 & 24.09 & 31.20 & 29.38 & 22.66 & 15.95 & 14.01 & 9.05  & 18.94 & 16.08 & 65.23 & 61.95 & 51.22 & 46.46 & 44.79 & 53.93  \\
              & Cycle 2  & 32.43 & 28.31 & 29.29 & 23.21 & 30.55 & 28.52 & 22.01 & 15.52 & 13.05 & 9.08  & 18.04 & 15.17 & 61.37 & 59.62 & 50.81 & 45.74 & 44.22 & 52.35 \\
              & Avg.     & 33.18 & 27.95 & 29.70 & 23.65 & 30.87 & 28.95 & 22.16 & 15.52 & 13.55 & 8.48  & 18.44 & 15.63 & 63.32 & 60.78 & 51.02 & 46.15 & 44.57 & 53.14 \\

\midrule 
ASR & Cycle 1 & 48.17 & 35.86 & 43.31 & 42.15 & 41.93 & 42.28 & 29.17 & 19.56 & 17.85 & 12.94 & 22.81 & 20.47 & 74.34 & 66.87 & 54.93 & 51.25 & 45.91 & 58.66 \\
    & Cycle 2 & 44.89 & 32.60 & 39.77 & 37.51 & 39.15 & 38.78 & 25.89 & 16.90 & 16.41 & 12.12 & 20.65 & 18.39 & 68.70 & 61.57 & 50.71 & 47.79 & 42.15 & 54.18 \\
    & Avg.    & 46.53 & 34.23 & 41.54 & 39.83 & 40.54 & 40.53 & 27.53 & 18.23 & 17.13 & 12.53 & 21.73 & 19.43 & 71.52 & 64.22 & 52.82 & 49.52 & 44.03 & 56.42 \\
\midrule 
PeTTA & Cycle 1 & 45.41 & 33.22 & 41.31 & 39.66 & 39.95 & 39.91 & 27.51 & 18.12 & 16.61 & 12.00 & 21.66 & 19.18 & 72.86 & 65.15 & 53.70 & 49.84 & 44.45 & 57.20 \\
      & Cycle 2 & 42.43 & 30.24 & 37.51 & 35.32 & 31.29 & 37.36 & 24.65 & 15.84 & 15.45 & 11.46 & 19.00 & 17.28 & 67.38 & 60.89 & 49.74 & 46.90 & 41.29 & 53.24 \\
      & Avg.    & 43.92 & 31.73 & 39.41 & 37.49 & 35.62 & 38.63 & 26.08 & 16.98 & 16.03 & 11.73 & 20.33 & 18.23 & 70.12 & 63.02 & 51.72 & 48.37 & 42.87 & 55.22 \\
\midrule 
DPCore & Cycle 1 & 42.01 & 28.62 & 34.49 & 29.01 & 30.53 & 32.93 & 25.87 & 17.92 & 15.66 & 11.01 & 15.22 & 17.14 & 66.91 & 61.26 & 50.91 & 44.25 & 43.96 & 53.46 \\
       & Cycle 2 & 41.37 & 28.14 & 33.67 & 28.45 & 28.07 & 31.94 & 22.29 & 14.84 & 12.60 & 9.45  & 13.74 & 14.58 & 64.13 & 59.88 & 49.83 & 41.79 & 43.18 & 51.76 \\
       & Avg.    & 41.69 & 28.38 & 34.08 & 28.73 & 29.30 & 32.93 & 24.08 & 16.38 & 14.13 & 10.23 & 14.48 & 15.13 & 65.52 & 60.57 & 50.37 & 43.02 & 43.57 & 52.91 \\
\midrule 
ReservoirTTA & Cycle 1 & 37.51 & 27.67 & 31.16 & 27.27 & 28.52 & 31.94 & 25.17 & 17.22 & 15.01 & 10.51 & 14.92 & 16.57 & 66.16 & 60.45 & 50.00 & 42.21 & 43.46 & 52.46 \\
             & Cycle 2 & 37.15 & 27.19 & 30.80 & 26.59 & 27.84 & 30.92 & 22.99 & 14.34 & 12.55 & 9.15  & 13.94 & 14.59 & 65.78 & 60.09 & 49.64 & 41.83 & 43.08 & 52.08 \\
             & Avg.    & 37.33 & 27.43 & 30.98 & 26.93 & 28.18 & 31.43 & 24.08 & 15.78 & 13.78 & 9.83  & 14.43 & 14.63 & 65.97 & 60.27 & 49.82 & 42.02 & 43.27 & 52.42 \\
\midrule 
LCoTTA & Cycle 1 & 35.91 & 27.47 & 29.71 & 25.86 & 27.17 & 30.25 & 24.92 & 16.97 & 14.41 & 10.21 & 14.57 & 16.22 & 69.86 & 64.01 & 53.66 & 48.21 & 45.51 & 56.25 \\
       & Cycle 2 & 35.55 & 27.09 & 29.45 & 25.40 & 26.49 & 29.40 & 21.14 & 13.29 & 11.05 & 8.55  & 11.19 & 13.04 & 69.38 & 63.63 & 53.28 & 47.83 & 45.13 & 55.85 \\
       & Avg.    & 35.73 & 27.28 & 29.58 & 25.63 & 26.83 & 29.83 & 23.03 & 15.13 & 12.73 & 9.38  & 12.88 & 13.93 & 69.62 & 63.82 & 53.47 & 48.02 & 45.32 & 56.12 \\
\midrule 
SloMo-Fast & Cycle 1 & 32.92 & 26.57 & 26.11 & 24.21 & 25.27 & 27.15 & 20.41 & 13.92 & 12.21 & 9.21  & 13.32 & 13.82 & 65.01 & 59.62 & 49.51 & 43.22 & 43.32 & 52.14 \\
           & Cycle 2 & 32.64 & 26.19 & 25.85 & 23.95 & 24.79 & 26.70 & 18.25 & 11.24 & 9.45  & 7.95  & 9.54  & 11.28 & 64.45 & 59.14 & 49.15 & 42.74 & 42.84 & 51.66 \\
           & Avg.    & 32.78 & 26.38 & 25.98 & 24.08 & 25.03 & \textbf{26.93} & 19.33 & 12.58 & 10.83 & 8.58  & 11.43 & \textbf{12.43} & 64.73 & 59.38 & 49.33 & 42.98 & 43.08 & \textbf{51.91} \\

        \bottomrule
    \end{tabular}
    \end{adjustbox}

    \caption{Our Proposed Cyclic TTA results of {\pa} compared with existing methods on CIFAR10-C and CIFAR100-C for different domain groups. Each subgroup completes a cycle of seeing different test domains twice. (Gaussian, Shot, Impulse): Noise, (Defocus, Glass, Motion, Zoom): Blur, (Snow, Frost, Fog): Weather, (Brightness, Contrast): Digital, (Elastic, Pixelate, JPEG): . {\pa} achieves the best performance across both datasets.}
    \label{cyclic_results}
\end{table*}

\begin{table*} 
\centering
\begin{tabular}{|c|c|c|c|c|c|c|c|}
\hline
\multirow{2}{*}{Method} & \multirow{2}{*}{Subgroup} & \multicolumn{3}{c|}{Cycle 1} & \multicolumn{3}{c|}{Cycle 2} \\ \cline{3-8}
 &  & Domain & Error (\%) & Avg & Domain & Error (\%) & Avg  \\ \hline
 
\multirow{15}{*}{TENT} & \multirow{3}{*}{Noise} & gaussian & 23.42 & \multirow{3}{*}{23.66} & gaussian & 24.87 & \multirow{3}{*}{25.49} \\ \cline{3-4} \cline{6-7}
 &  & shot & 21.98 &  & shot & 24.37 &  \\ \cline{3-4} \cline{6-7}
 &  & impulse & 25.58 &  & impulse & 21.74 &  \\ \cline{2-8}
 
 & \multirow{4}{*}{Blur} & defocus & 11.81 & \multirow{4}{*}{16.95} & defocus & 11.81 & \multirow{4}{*}{16.95} \\ \cline{3-4} \cline{6-7}
 &  & glass & 29.76 &  & glass & 29.76 &  \\ \cline{3-4} \cline{6-7}
 &  & motion & 14.01 &  & motion & 14.01 &  \\ \cline{3-4} \cline{6-7}
 &  & zoom & 12.23 &  & zoom & 12.23 &  \\ \cline{2-8}
 
 & \multirow{3}{*}{Weather} & snow & 16.34 & \multirow{3}{*}{15.22} & snow & 14.98 & \multirow{3}{*}{14.99} \\ \cline{3-4} \cline{6-7}
 &  & frost & 15.94 &  & frost & 15.44 &  \\ \cline{3-4} \cline{6-7}
 &  & fog & 14.10 &  & fog & 14.55 &  \\ \cline{2-8}
 
 & \multirow{2}{*}{Digital} & brightness & 7.91 & \multirow{2}{*}{9.07} & brightness & 7.67 & \multirow{2}{*}{8.78} \\ \cline{3-4} \cline{6-7}
 &  & contrast & 10.81 &  & contrast & 9.89 &  \\ \cline{2-8}
 
 & \multirow{3}{*}{Distortion} & elastic & 22.11 & \multirow{3}{*}{20.09} & elastic & 20.55 & \multirow{3}{*}{19.47} \\ \cline{3-4} \cline{6-7}
 &  & pixel & 16.22 &  & pixel & 15.54 &  \\ \cline{3-4} \cline{6-7}
 &  & jpeg & 23.77 &  & jpeg & 22.33 &  \\  \hline
 
& & \multicolumn{3}{c|}{Cycle 1 Avg: 17.47\%} & \multicolumn{3}{c|}{Cycle 2 Avg: 17.14\%} \\ \hline
\end{tabular}

\caption{Detailed Evaluation Results for TENT on CIFAR10-C under Cyclic Domain Settings}
\label{table:cyclic_tent_cifar10c}
\end{table*}
 
\begin{table*}
\centering
\begin{tabular}{|c|c|c|c|c|c|c|c|}
\hline
\multirow{2}{*}{Method} & \multirow{2}{*}{Subgroup} & \multicolumn{3}{c|}{Cycle 1} & \multicolumn{3}{c|}{Cycle 2} \\ \cline{3-8}
 &  & Domain & Error (\%) & Avg & Domain & Error (\%) & Avg \\ \hline
 
\multirow{15}{*}{TENT} & \multirow{3}{*}{Noise} & gaussian & 38.12 & \multirow{3}{*}{38.28} & gaussian & 47.32 & \multirow{3}{*}{47.88} \\ \cline{3-4} \cline{6-7}
 &  & shot & 38.45 &  & shot & 48.23 &  \\ \cline{3-4} \cline{6-7}
 &  & impulse & 38.27 &  & impulse & 48.09 &  \\  \cline{2-8}
 
 & \multirow{4}{*}{Blur} & defocus & 30.87 & \multirow{4}{*}{31.14} & defocus & 37.00 & \multirow{4}{*}{37.12} \\ \cline{3-4} \cline{6-7}
 &  & glass & 31.19 &  & glass & 36.78 &  \\ \cline{3-4} \cline{6-7}
 &  & motion & 30.75 &  & motion & 37.39 &  \\ \cline{3-4} \cline{6-7}
 &  & zoom & 31.27 &  & zoom & 37.50 &  \\  \cline{2-8}
 
 & \multirow{3}{*}{Weather} & snow & 33.05 & \multirow{3}{*}{32.93} & snow & 36.88 & \multirow{3}{*}{37.93} \\ \cline{3-4} \cline{6-7}
 &  & frost & 33.21 &  & frost & 36.32 &  \\ \cline{3-4} \cline{6-7}
 &  & fog & 32.55 &  & fog & 38.58 &   \\ \cline{2-8}
 
 & \multirow{2}{*}{Digital} & brightness & 25.32 & \multirow{2}{*}{25.04} & brightness & 24.95 & \multirow{2}{*}{25.18} \\ \cline{3-4} \cline{6-7}
 &  & contrast & 24.76 &  & contrast & 25.41 &   \\ \cline{2-8}
 
 & \multirow{3}{*}{Distortion} & elastic & 33.72 & \multirow{3}{*}{34.09} & elastic & 39.05 & \multirow{3}{*}{38.95} \\ \cline{3-4} \cline{6-7}
 &  & pixel & 34.56 &  & pixel & 39.14 &  \\ \cline{3-4} \cline{6-7}
 &  & jpeg & 33.98 &  & jpeg & 38.66 &  \\ 
\hline
 
& & \multicolumn{3}{c|}{Cycle 1 Avg: 32.29\%} & \multicolumn{3}{c|}{Cycle 2 Avg: 37.41\%} \\ \hline
\end{tabular}
\caption{Detailed Evaluation Results for TENT on CIFAR100-C under Cyclic Domain Settings}
 \label{table:cyclic_tent_cifar100c}
\end{table*}

\begin{table*} 
\centering
\begin{tabular}{|c|c|c|c|c|c|c|c|}
\hline
\multirow{2}{*}{Method} & \multirow{2}{*}{Subgroup} & \multicolumn{3}{c|}{Cycle 1} & \multicolumn{3}{c|}{Cycle 2} \\ \cline{3-8}
 &  & Domain & Error (\%) & Avg & Domain & Error (\%) & Avg  \\ \hline
 
\multirow{15}{*}{TENT} & \multirow{3}{*}{Noise} & gaussian & 81.38 & \multirow{3}{*}{76.35} & gaussian & 70.78 & \multirow{3}{*}{69.31} \\ \cline{3-4} \cline{6-7}
 &  & shot & 74.82 &  & shot & 68.50 &  \\ \cline{3-4} \cline{6-7}
 &  & impulse & 72.86 &  & impulse & 68.66 &  \\ \cline{2-8}
 
 & \multirow{4}{*}{Blur} & defocus & 81.66 & \multirow{4}{*}{69.32} & defocus & 72.56 & \multirow{4}{*}{65.14} \\ \cline{3-4} \cline{6-7}
 &  & glass & 77.04 &  & glass & 72.72 &  \\ \cline{3-4} \cline{6-7}
 &  & motion & 65.18 &  & motion & 62.26 &  \\ \cline{3-4} \cline{6-7}
 &  & zoom & 53.40 &  & zoom & 53.00 &  \\ \cline{2-8}
 
 & \multirow{3}{*}{Weather} & snow & 62.02 & \multirow{3}{*}{57.09} & snow & 56.38 & \multirow{3}{*}{54.19} \\ \cline{3-4} \cline{6-7}
 &  & frost & 62.66 &  & frost & 60.58 &  \\ \cline{3-4} \cline{6-7}
 &  & fog & 46.58 &  & fog & 45.62 &  \\ \cline{2-8}
 
 & \multirow{2}{*}{Digital} & brightness & 34.22 & \multirow{2}{*}{55.31} & brightness & 33.20 & \multirow{2}{*}{52.27} \\ \cline{3-4} \cline{6-7}
 &  & contrast & 76.40 &  & contrast & 71.34 &  \\ \cline{2-8}
 
 & \multirow{3}{*}{Distortion} & elastic & 52.92 & \multirow{3}{*}{50.69} & elastic & 47.82 & \multirow{3}{*}{47.47} \\ \cline{3-4} \cline{6-7}
 &  & pixel & 46.36 &  & pixel & 44.22 &  \\ \cline{3-4} \cline{6-7}
 &  & jpeg & 52.78 &  & jpeg & 50.36 &  \\  \hline
 
& & \multicolumn{3}{c|}{Cycle 1 Avg: 61.75\%} & \multicolumn{3}{c|}{Cycle 2 Avg: 57.68\%} \\ \hline
\end{tabular}

\caption{Detailed Evaluation Results for TENT on Imagenet-C under Cyclic Domain Settings}
\label{table:cyclic_tent_imagenetc}
\end{table*}
 
\begin{table*}
\centering
\begin{tabular}{|c|c|c|c|c|c|c|c|}
\hline
\multirow{2}{*}{Method} & \multirow{2}{*}{Subgroup} & \multicolumn{3}{c|}{Cycle 1} & \multicolumn{3}{c|}{Cycle 2} \\ \cline{3-8} 
 &  & Domain & Error (\%) & Avg. & Domain & Error (\%) & Avg. \\ \hline
 
\multirow{15}{*}{COTTA} & \multirow{3}{*}{Noise} & gaussian & 36.14 & \multirow{3}{*}{36.52} & gaussian & 44.23 & \multirow{3}{*}{44.67} \\ \cline{3-4} \cline{6-7}
 &  & shot & 36.84 &  & shot & 44.98 &  \\ \cline{3-4} \cline{6-7}
 &  & impulse & 36.57 &  & impulse & 44.81 &   \\ \cline{2-8}
 
 & \multirow{4}{*}{Blur} & defocus & 29.12 & \multirow{4}{*}{29.43} & defocus & 34.45 & \multirow{4}{*}{34.69} \\ \cline{3-4} \cline{6-7}
 &  & glass & 29.55 &  & glass & 34.08 &  \\ \cline{3-4} \cline{6-7}
 &  & motion & 28.99 &  & motion & 34.72 &  \\ \cline{3-4} \cline{6-7}
 &  & zoom & 29.36 &  & zoom & 34.51 &  \\  \cline{2-8}
 
 & \multirow{3}{*}{Weather} & snow & 31.25 & \multirow{3}{*}{30.98} & snow & 35.45 & \multirow{3}{*}{35.93} \\ \cline{3-4} \cline{6-7}
 &  & frost & 30.84 &  & frost & 35.21 &  \\ \cline{3-4} \cline{6-7}
 &  & fog & 30.85 &  & fog & 37.13 &  \\  \cline{2-8}
 
 & \multirow{2}{*}{Digital} & brightness & 23.28 & \multirow{2}{*}{23.56} & brightness & 23.95 & \multirow{2}{*}{23.97} \\ \cline{3-4} \cline{6-7}
 &  & contrast & 23.84 &  & contrast & 24.09 &  \\ \cline{2-8}
 
 & \multirow{3}{*}{Distortion} & elastic & 32.48 & \multirow{3}{*}{32.75} & elastic & 36.54 & \multirow{3}{*}{36.39} \\ \cline{3-4} \cline{6-7}
 &  & pixel & 32.88 &  & pixel & 36.19 &  \\ \cline{3-4} \cline{6-7}
 &  & jpeg & 32.89 &  & jpeg & 36.44 &  \\ \hline
 
& & \multicolumn{3}{c|}{Cycle 1 Avg: 30.96\%} & \multicolumn{3}{c|}{Cycle 2 Avg: 34.69\%} \\ \hline
\end{tabular}

\caption{Detailed Evaluation Results for COTTA on CIFAR100-C under Cyclic Domain Settings}
\label{table:cyclic_cotta_cifa100c}
\end{table*}

\begin{table*}
\centering
\begin{tabular}{|c|c|c|c|c|c|c|c|}
\hline
\multirow{2}{*}{Method} & \multirow{2}{*}{Subgroup} & \multicolumn{3}{c|}{Cycle 1} & \multicolumn{3}{c|}{Cycle 2} \\ \cline{3-8} 
 &  & Domain & Error (\%) & Avg. & Domain & Error (\%) & Avg. \\ \hline
 
\multirow{15}{*}{COTTA} & \multirow{3}{*}{Noise} & gaussian & 84.54 & \multirow{3}{*}{82.38} & gaussian & 80.64 & \multirow{3}{*}{78.63} \\ \cline{3-4} \cline{6-7}
 &  & shot & 81.94 &  & shot & 78.30 &  \\ \cline{3-4} \cline{6-7}
 &  & impulse & 80.66 &  & impulse & 76.96 &   \\ \cline{2-8}
 
 & \multirow{4}{*}{Blur} & defocus & 86.00 & \multirow{4}{*}{76.91} & defocus & 79.48 & \multirow{4}{*}{72.06} \\ \cline{3-4} \cline{6-7}
 &  & glass & 83.74 &  & glass & 77.06 &  \\ \cline{3-4} \cline{6-7}
 &  & motion & 73.82 &  & motion & 70.00 &  \\ \cline{3-4} \cline{6-7}
 &  & zoom & 64.06 &  & zoom & 61.70 &  \\  \cline{2-8}
 
 & \multirow{3}{*}{Weather} & snow & 65.04 & \multirow{3}{*}{59.52} & snow & 60.76 & \multirow{3}{*}{55.52} \\ \cline{3-4} \cline{6-7}
 &  & frost & 61.38 &  & frost &  &  \\ \cline{3-4} \cline{6-7}
 &  & fog & 47.80 &  & fog & 44.42 &  \\  \cline{2-8}
 
 & \multirow{2}{*}{Digital} & brightness & 34.64 & \multirow{2}{*}{56.98} & brightness & 34.22 & \multirow{2}{*}{53.79} \\ \cline{3-4} \cline{6-7}
 &  & contrast & 79.32 &  & contrast & 73.36 &  \\ \cline{2-8}
 
 & \multirow{3}{*}{Distortion} & elastic & 55.72 & \multirow{3}{*}{52.82} & elastic & 50.84 & \multirow{3}{*}{47.91} \\ \cline{3-4} \cline{6-7}
 &  & pixel & 48.10 &  & pixel & 42.48 &  \\ \cline{3-4} \cline{6-7}
 &  & jpeg & 54.64 &  & jpeg & 50.40 &  \\ \hline
 
& & \multicolumn{3}{c|}{Cycle 1 Avg: 65.72\%} & \multicolumn{3}{c|}{Cycle 2 Avg: 61.58\%} \\ \hline
\end{tabular}

\caption{Detailed Evaluation Results for COTTA on Imagenet-C under Cyclic Domain Settings}
\label{table:cyclic_cotta_imagenetc}
\end{table*}
 
 \begin{table*} 
\centering
\begin{tabular}{|c|c|c|c|c|c|c|c|}
\hline
\multirow{2}{*}{Method} & \multirow{2}{*}{Subgroup} & \multicolumn{3}{c|}{Cycle 1} & \multicolumn{3}{c|}{Cycle 2} \\ \cline{3-8}
 &  & Domain & Error (\%) & Avg & Domain & Error (\%) & Avg \\ \hline
 
\multirow{15}{*}{ROID} & \multirow{3}{*}{Noise} & gaussian & 23.94 & \multirow{3}{*}{22.32} & gaussian & 20.62 & \multirow{3}{*}{22.16} \\ \cline{3-4} \cline{6-7}
 &  & shot & 22.41 &  & shot & 21.00 &  \\ \cline{3-4} \cline{6-7}
 &  & impulse & 20.62 &  & impulse & 24.87 &  \\ \cline{2-8}
 
 & \multirow{4}{*}{Blur} & defocus & 10.52 & \multirow{4}{*}{15.52} & defocus & 10.52 & \multirow{4}{*}{15.52} \\ \cline{3-4} \cline{6-7}
 &  & glass & 28.20 &  & glass & 28.20 &  \\ \cline{3-4} \cline{6-7}
 &  & motion & 12.06 &  & motion & 12.06 &  \\ \cline{3-4} \cline{6-7}
 &  & zoom & 10.06 &  & zoom & 10.06 &  \\  \cline{2-8}
 
 & \multirow{3}{*}{Weather} & snow & 15.12 & \multirow{3}{*}{13.55} & snow & 14.01 & \multirow{3}{*}{13.24} \\ \cline{3-4} \cline{6-7}
 &  & frost & 14.41 &  & frost & 13.79 &  \\ \cline{3-4} \cline{6-7}
 &  & fog & 12.04 &  & fog & 11.92 &  \\ \cline{2-8}
 
 & \multirow{2}{*}{Digital} & brightness & 7.76 & \multirow{2}{*}{8.48} & brightness & 7.37 & \multirow{2}{*}{8.27} \\ \cline{3-4} \cline{6-7}
 &  & contrast & 9.61 &  & contrast & 9.17 &  \\ \cline{2-8}
 
 & \multirow{3}{*}{Distortion} & elastic & 21.08 & \multirow{3}{*}{18.44} & elastic & 19.16 & \multirow{3}{*}{17.90} \\ \cline{3-4} \cline{6-7}
 &  & pixel & 15.22 &  & pixel & 14.51 &  \\ \cline{3-4} \cline{6-7}
 &  & jpeg & 20.62 &  & jpeg & 20.02 &  \\ \cline{2-8}
 
&  & \multicolumn{3}{c|}{Cycle 1 Avg: 16.08\%} & \multicolumn{3}{c|}{Cycle 2 Avg: 15.17\%} \\ \hline
\end{tabular}
\caption{Detailed Evaluation Results for ROID on CIFAR10-C under Cyclic Domain Settings}
\label{table:cyclic_roid_cifar10c}
\end{table*}

\begin{table*} 
\centering
\begin{tabular}{|c|c|c|c|c|c|c|c|}
\hline
\multirow{2}{*}{Method} & \multirow{2}{*}{Subgroup} & \multicolumn{3}{c|}{Cycle 1} & \multicolumn{3}{c|}{Cycle 2} \\ \cline{3-8}
 &  & Domain & Error (\%) & Avg & Domain & Error (\%) & Avg \\ \hline
\multirow{15}{*}{ROID} & \multirow{3}{*}{Noise} & gaussian & 32.34 & \multirow{3}{*}{32.53} & gaussian & 33.67 & \multirow{3}{*}{33.83} \\ \cline{3-4} \cline{6-7}
 &  & shot & 33.12 &  & shot & 34.58 &  \\ \cline{3-4} \cline{6-7}
 &  & impulse & 32.12 &  & impulse & 33.25 &  \\ \cline{2-8}
 
 & \multirow{4}{*}{Blur} & defocus & 27.12 & \multirow{5}{*}{26.44} & defocus & 28.44 & \multirow{5}{*}{28.31} \\ \cline{3-4} \cline{6-7}
 &  & glass & 25.67 &  & glass & 27.23 &  \\ \cline{3-4} \cline{6-7}
 &  & motion & 26.22 &  & motion & 28.23 &  \\ \cline{3-4} \cline{6-7}
 &  & zoom & 26.65 &  & zoom & 29.34 &  \\  \cline{2-8}
 
 & \multirow{3}{*}{Weather} & snow & 28.77 & \multirow{3}{*}{30.11} & snow & 29.29 & \multirow{3}{*}{29.70} \\ \cline{3-4} \cline{6-7}
 &  & frost & 28.06 &  & frost & 28.77 &  \\ \cline{3-4} \cline{6-7}
 &  & fog & 33.50 &  & fog & 31.03 &  \\ \cline{2-8}
 
 & \multirow{2}{*}{Digital} & brightness & 23.64 & \multirow{2}{*}{24.09} & brightness & 22.46 & \multirow{2}{*}{23.21} \\ \cline{3-4} \cline{6-7}
 &  & contrast & 24.53 &  & contrast & 23.95 &  \\  \cline{2-8}
 
 & \multirow{3}{*}{Distortion} & elastic & 32.08 & \multirow{3}{*}{31.20} & elastic & 30.86 & \multirow{3}{*}{30.55} \\ \cline{3-4} \cline{6-7}
 &  & pixel & 27.28 &  & pixel & 26.90 &  \\ \cline{3-4} \cline{6-7}
 &  & jpeg & 34.23 &  & jpeg & 33.88 &  \\  \hline
& & \multicolumn{3}{c|}{Cycle 1 Avg: 29.38\%} & \multicolumn{3}{c|}{Cycle 2 Avg: 28.52\%} \\ \hline
\end{tabular}

\caption{Detailed Evaluation Results for ROID on CIFAR100-C under Cyclic Domain Settings}
\label{table:cyclic_roid_cifar100c}
\end{table*}

\begin{table*} 
\centering
\begin{tabular}{|c|c|c|c|c|c|c|c|}
\hline
\multirow{2}{*}{Method} & \multirow{2}{*}{Subgroup} & \multicolumn{3}{c|}{Cycle 1} & \multicolumn{3}{c|}{Cycle 2} \\ \cline{3-8}
 &  & Domain & Error (\%) & Avg & Domain & Error (\%) & Avg \\ \hline
\multirow{15}{*}{ROID} & \multirow{3}{*}{Noise} & gaussian & 72.24 & \multirow{3}{*}{65.23} & gaussian & 61.84 & \multirow{3}{*}{61.37} \\ \cline{3-4} \cline{6-7}
 &  & shot & 61.54 &  & shot & 60.46 &  \\ \cline{3-4} \cline{6-7}
 &  & impulse & 61.92 &  & impulse & 61.80 &  \\ \cline{2-8}
 
 & \multirow{4}{*}{Blur} & defocus & 72.68 & \multirow{5}{*}{61.95} & defocus & 66.64 & \multirow{5}{*}{59.62} \\ \cline{3-4} \cline{6-7}
 &  & glass & 67.26 &  & glass & 66.02 &  \\ \cline{3-4} \cline{6-7}
 &  & motion & 58.36 &  & motion & 57.02 &  \\ \cline{3-4} \cline{6-7}
 &  & zoom & 49.50 &  & zoom & 48.82 &  \\  \cline{2-8}
 
 & \multirow{3}{*}{Weather} & snow & 52.54 & \multirow{3}{*}{51.22} & snow & 51.38 & \multirow{3}{*}{50.81} \\ \cline{3-4} \cline{6-7}
 &  & frost & 57.96 &  & frost & 57.76 &  \\ \cline{3-4} \cline{6-7}
 &  & fog & 43.16 &  & fog & 43.30 &  \\ \cline{2-8}
 
 & \multirow{2}{*}{Digital} & brightness & 33.30 & \multirow{2}{*}{46.46} & brightness & 33.88 & \multirow{2}{*}{45.74} \\ \cline{3-4} \cline{6-7}
 &  & contrast & 59.62 &  & contrast & 57.60 &  \\  \cline{2-8}
 
 & \multirow{3}{*}{Distortion} & elastic & 45.32 & \multirow{3}{*}{44.79} & elastic & 44.10 & \multirow{3}{*}{44.22} \\ \cline{3-4} \cline{6-7}
 &  & pixel & 42.40 &  & pixel & 42.36 &  \\ \cline{3-4} \cline{6-7}
 &  & jpeg & 46.64 &  & jpeg & 46.20 &  \\  \hline
& & \multicolumn{3}{c|}{Cycle 1 Avg: 53.93\%} & \multicolumn{3}{c|}{Cycle 2 Avg: 52.35\%} \\ \hline
\end{tabular}

\caption{Detailed Evaluation Results for ROID on Imagenet-C under Cyclic Domain Settings}
\label{table:cyclic_roid_imagenetc}
\end{table*}

\begin{table*} 
\centering
\begin{tabular}{|c|c|c|c|c|c|c|c|}
\hline
\multirow{3}{*}{Method} & \multirow{3}{*}{Subgroup} & \multicolumn{3}{c|}{Cycle 1} & \multicolumn{3}{c|}{Cycle 2} \\ \cline{3-8}
 &  & Domain & Error (\%) & Avg & Domain & Error (\%) & Avg \\ \hline
 
\multirow{15}{*}{RoTTA} & \multirow{3}{*}{Noise} & gaussian & 30.21 & \multirow{3}{*}{30.08} & gaussian & 25.50 & \multirow{3}{*}{26.18} \\ \cline{3-4} \cline{6-7}
 &  & shot & 25.43 &  & shot & 22.32 &  \\ \cline{3-4} \cline{6-7}
 &  & impulse & 34.59 &  & impulse & 30.72 &  \\ \cline{2-8}
 
 & \multirow{4}{*}{Blur} & defocus & 13.80 & \multirow{4}{*}{18.98} & defocus & 11.33 & \multirow{4}{*}{16.66} \\ \cline{3-4} \cline{6-7}
 &  & glass & 36.19 &  & glass & 31.81 &  \\ \cline{3-4} \cline{6-7}
 &  & motion & 14.78 &  & motion & 13.49 &  \\ \cline{3-4} \cline{6-7}
 &  & zoom & 11.13 &  & zoom & 10.01 &  \\  \cline{2-8}
 
 & \multirow{3}{*}{Weather} & snow & 17.81 & \multirow{3}{*}{17.00} & snow & 16.27 & \multirow{3}{*}{16.02} \\ \cline{3-4} \cline{6-7}
 &  & frost & 17.68 &  & frost & 15.53 &  \\ \cline{3-4} \cline{6-7}
 &  & fog & 15.52 &  & fog & 13.30 &  \\ \cline{2-8}
 
 & \multirow{2}{*}{Digital} & brightness & 8.06 & \multirow{2}{*}{12.39} & brightness & 8.83 & \multirow{2}{*}{12.39} \\ \cline{3-4} \cline{6-7}
 &  & contrast & 18.35 &  & contrast & 14.32 &  \\ \cline{2-8}
 
 & \multirow{3}{*}{Distortion} & elastic & 23.64 & \multirow{3}{*}{23.95} & elastic & 22.15 & \multirow{3}{*}{21.68} \\ \cline{3-4} \cline{6-7}
 &  & pixel & 21.65 &  & pixel & 19.46 &  \\ \cline{3-4} \cline{6-7}
 &  & jpeg & 26.57 &  & jpeg & 23.44 &   \\ \hline
 
& & \multicolumn{3}{c|}{Cycle 1 Avg: 20.64\%} & \multicolumn{3}{c|}{Cycle 2 Avg: 18.23\%} \\ \hline
\end{tabular}

\caption{Detailed Evaluation Results for RoTTA on CIFAR10-C under Cyclic Domain Settings}
\label{table:cyclic_RoTTA_cifar10c}
\end{table*}

\begin{table*} 
\centering
\begin{tabular}{|c|c|c|c|c|c|c|c|}
\hline
\multirow{3}{*}{Method} & \multirow{3}{*}{Subgroup} & \multicolumn{3}{c|}{Cycle 1} & \multicolumn{3}{c|}{Cycle 2} \\ \cline{3-8}
 &  & Domain & Error (\%) & Avg & Domain & Error (\%) & Avg \\ \hline
 
\multirow{15}{*}{RoTTA} & \multirow{3}{*}{Noise} & gaussian & 49.48 & \multirow{3}{*}{46.66} & gaussian & 41.89 & \multirow{3}{*}{43.72} \\ \cline{3-4} \cline{6-7}
 &  & shot & 44.87 &  & shot & 39.18 &  \\ \cline{3-4} \cline{6-7}
 &  & impulse & 45.62 &  & impulse & 41.29 &  \\ \cline{2-8}
 
 & \multirow{4}{*}{Blur} & defocus & 29.94 & \multirow{4}{*}{33.82} & defocus & 25.95 & \multirow{4}{*}{29.96} \\ \cline{3-4} \cline{6-7}
 &  & glass & 47.33 &  & glass & 40.52 &  \\ \cline{3-4} \cline{6-7}
 &  & motion & 30.86 &  & motion & 28.32 &  \\ \cline{3-4} \cline{6-7}
 &  & zoom & 27.16 &  & zoom & 25.07 &  \\  \cline{2-8}
 
 & \multirow{3}{*}{Weather} & snow & 39.00 & \multirow{3}{*}{40.83} & snow & 32.99 & \multirow{3}{*}{34.31} \\ \cline{3-4} \cline{6-7}
 &  & frost & 41.40 &  & frost & 33.15 &  \\ \cline{3-4} \cline{6-7}
 &  & fog & 42.09 &  & fog & 36.78 &  \\ \cline{2-8}
 
 & \multirow{2}{*}{Digital} & brightness & 28.95 & \multirow{2}{*}{41.70} & brightness & 26.63 & \multirow{2}{*}{32.09} \\ \cline{3-4} \cline{6-7}
 &  & contrast & 54.44 &  & contrast & 37.56 &  \\ \cline{2-8}
 
 & \multirow{3}{*}{Distortion} & elastic & 40.58 & \multirow{3}{*}{41.17} & elastic & 35.83 & \multirow{3}{*}{36.35} \\ \cline{3-4} \cline{6-7}
 &  & pixel & 40.05 &  & pixel & 33.93 &  \\ \cline{3-4} \cline{6-7}
 &  & jpeg & 42.89 &  & jpeg & 39.28 &   \\ \hline
 
& & \multicolumn{3}{c|}{Cycle 1 Avg: 40.84\%} & \multicolumn{3}{c|}{Cycle 2 Avg: 34.70\%} \\ \hline
\end{tabular}
\caption{Detailed Evaluation Results for RoTTA on CIFAR100-C under Cyclic Domain Settings}
\label{table:cyclic_RoTTA_cifar100c}
\end{table*}

\begin{table*} 
\centering
\begin{tabular}{|c|c|c|c|c|c|c|c|}
\hline
\multirow{3}{*}{Method} & \multirow{3}{*}{Subgroup} & \multicolumn{3}{c|}{Cycle 1} & \multicolumn{3}{c|}{Cycle 2} \\ \cline{3-8}
 &  & Domain & Error (\%) & Avg & Domain & Error (\%) & Avg \\ \hline
 
\multirow{15}{*}{RoTTA} & \multirow{3}{*}{Noise} & gaussian & 87.98 & \multirow{3}{*}{84.36} & gaussian & 81.94 & \multirow{3}{*}{80.53} \\ \cline{3-4} \cline{6-7}
 &  & shot & 82.74 &  & shot & 80.28 &  \\ \cline{3-4} \cline{6-7}
 &  & impulse & 82.36 &  & impulse & 79.38&  \\ \cline{2-8}
 
 & \multirow{4}{*}{Blur} & defocus & 84.66 & \multirow{4}{*}{76.76} & defocus & 79.40 & \multirow{4}{*}{72.54} \\ \cline{3-4} \cline{6-7}
 &  & glass & 86.60 &  & glass & 81.941 &  \\ \cline{3-4} \cline{6-7}
 &  & motion & 75.60 &  & motion & 71.58 &  \\ \cline{3-4} \cline{6-7}
 &  & zoom & 60.16 &  & zoom & 57.22 &  \\  \cline{2-8}
 
 & \multirow{3}{*}{Weather} & snow & 67.04 & \multirow{3}{*}{63.04} & snow & 64.90 & \multirow{3}{*}{60.50} \\ \cline{3-4} \cline{6-7}
 &  & frost & 67.48 &  & frost & 64.96 &  \\ \cline{3-4} \cline{6-7}
 &  & fog & 54.60 &  & fog & 51.64 &  \\ \cline{2-8}
 
 & \multirow{2}{*}{Digital} & brightness & 34.54 & \multirow{2}{*}{58.09} & brightness & 35.94 & \multirow{2}{*}{57.35} \\ \cline{3-4} \cline{6-7}
 &  & contrast & 81.64 &  & contrast & 78.76 &  \\ \cline{2-8}
 
 & \multirow{3}{*}{Distortion} & elastic & 55.44 & \multirow{3}{*}{55.21} & elastic & 53.78 & \multirow{3}{*}{52.76} \\ \cline{3-4} \cline{6-7}
 &  & pixel & 52.10 &  & pixel & 49.34 &  \\ \cline{3-4} \cline{6-7}
 &  & jpeg & 58.10 &  & jpeg & 55.16 &   \\ \hline
 
& & \multicolumn{3}{c|}{Cycle 1 Avg:67.49\%} & \multicolumn{3}{c|}{Cycle 2 Avg: 64.74\%} \\ \hline
\end{tabular}

\caption{Detailed Evaluation Results for RoTTA on Imagenet-C under Cyclic Domain Settings}
\label{table:cyclic_RoTTA_imagenetc}
\end{table*}

\begin{table*}
\centering
\begin{tabular}{|c|c|c|c|c|c|c|c|}
\hline
\multirow{2}{*}{Method} & \multirow{2}{*}{Subgroup} & \multicolumn{3}{c|}{Cycle 1} & \multicolumn{3}{c|}{Cycle 2} \\ \cline{3-8}
 &  & Domain & Error (\%) & Avg & Domain & Error (\%) & Avg \\ \hline
 
\multirow{15}{*}{SloMo-Fast} & \multirow{3}{*}{Noise} & gaussian & 23.89 & \multirow{3}{*}{23.05} & gaussian & 19.96 & \multirow{3}{*}{20.76} \\ \cline{3-4} \cline{6-7}
 &  & shot & 18.89 &  & shot & 17.60 &  \\ \cline{3-4} \cline{6-7}
 &  & impulse & 26.38 &  & impulse & 24.72 &  \\ \cline{2-8}
 
 & \multirow{4}{*}{Blur} & defocus & 12.23 & \multirow{4}{*}{15.73} & defocus & 11.47 & \multirow{4}{*}{15.09} \\ \cline{3-4} \cline{6-7}
 &  & glass & 26.33 &  & glass & 24.83 &  \\ \cline{3-4} \cline{6-7}
 &  & motion & 13.50 &  & motion & 13.34 &  \\ \cline{3-4} \cline{6-7}
 &  & zoom & 10.86 &  & zoom & 10.71 &  \\  \cline{2-8}
 
 & \multirow{3}{*}{Weather} & snow & 13.86 & \multirow{3}{*}{13.51} & snow & 13.39 & \multirow{3}{*}{13.17} \\ \cline{3-4} \cline{6-7}
 &  & frost & 13.42 &  & frost & 13.26 &  \\ \cline{3-4} \cline{6-7}
 &  & fog & 13.26 &  & fog & 12.85 &  \\ \cline{2-8}
 
 & \multirow{2}{*}{Digital} & brightness & 7.79 & \multirow{2}{*}{9.43} & brightness & 8.22 & \multirow{2}{*}{9.30} \\ \cline{3-4} \cline{6-7}
 &  & contrast & 11.08 &  & contrast & 10.38 &  \\ \cline{2-8}
 
 & \multirow{3}{*}{Distortion} & elastic & 18.15 & \multirow{3}{*}{16.22} & elastic & 17.91 & \multirow{3}{*}{16.02} \\ \cline{3-4} \cline{6-7}
 &  & pixel & 13.21 &  & pixel & 13.04 &  \\ \cline{3-4} \cline{6-7}
 &  & jpeg & 17.31 &  & jpeg & 17.10 &   \\ \hline
 
& & \multicolumn{3}{c|}{Cycle 1 Avg: 15.59\%} & \multicolumn{3}{c|}{Cycle 2 Avg: 14.87\%} \\ \hline
\end{tabular}

\caption{Detailed Evaluation Results for {\pa} on CIFAR10-C under Cyclic Domain Settings}
\label{table:cyclic_ours_cifar10c}
\end{table*}

\begin{table*}
\centering
\begin{tabular}{|c|c|c|c|c|c|c|c|}
\hline
\multirow{2}{*}{Method} & \multirow{2}{*}{Subgroup} & \multicolumn{3}{c|}{Cycle 1} & \multicolumn{3}{c|}{Cycle 2} \\ \cline{3-8}
 &  & Domain & Error (\%) & Avg & Domain & Error (\%) & Avg \\ \hline
 
\multirow{15}{*}{SloMo-Fast} & \multirow{3}{*}{Noise} & gaussian & 35.59 & \multirow{3}{*}{33.71} & gaussian & 32.39 & \multirow{3}{*}{31.45} \\ \cline{3-4} \cline{6-7}
 &  & shot & 31.41 &  & shot & 30.76 &  \\ \cline{3-4} \cline{6-7}
 &  & impulse & 34.14 &  & impulse & 31.19 &  \\ \cline{2-8}
 
 & \multirow{4}{*}{Blur} & defocus & 24.68 & \multirow{4}{*}{27.64} & defocus & 24.63 & \multirow{4}{*}{27.13} \\ \cline{3-4} \cline{6-7}
 &  & glass & 33.52 &  & glass & 32.31 &  \\ \cline{3-4} \cline{6-7}
 &  & motion & 26.84 &  & motion & 26.65 &  \\ \cline{3-4} \cline{6-7}
 &  & zoom & 25.53 &  & zoom & 25.93 &  \\  \cline{2-8}
 
 & \multirow{3}{*}{Weather} & snow & 28.23 & \multirow{3}{*}{29.39} & snow & 27.52 & \multirow{3}{*}{29.07} \\ \cline{3-4} \cline{6-7}
 &  & frost & 27.95 &  & frost & 27.65 &  \\ \cline{3-4} \cline{6-7}
 &  & fog & 31.99 &  & fog & 32.04 &  \\ \cline{2-8}
 
 & \multirow{2}{*}{Digital} & brightness & 23.03 & \multirow{2}{*}{24.01} & brightness & 23.47 & \multirow{2}{*}{24.20} \\ \cline{3-4} \cline{6-7}
 &  & contrast & 25.00 &  & contrast & 24.93 &  \\ \cline{2-8}
 
 & \multirow{3}{*}{Distortion} & elastic & 29.61 & \multirow{3}{*}{29.62} & elastic & 29.94 & \multirow{3}{*}{29.84} \\ \cline{3-4} \cline{6-7}
 &  & pixel & 25.94 &  & pixel & 26.11 &  \\ \cline{3-4} \cline{6-7}
 &  & jpeg & 33.30 &  & jpeg & 33.47 &   \\ \hline
 
& & \multicolumn{3}{c|}{Cycle 1 Avg: 28.89\%} & \multicolumn{3}{c|}{Cycle 2 Avg: 28.34\%} \\ \hline
\end{tabular}

\caption{Detailed Evaluation Results for {\pa} on CIFAR100-C under Cyclic Domain Settings}
\label{table:cyclic_ours_cifar100c}
\end{table*}

\begin{table*}
\centering
\begin{tabular}{|c|c|c|c|c|c|c|c|}
\hline
\multirow{2}{*}{Method} & \multirow{2}{*}{Subgroup} & \multicolumn{3}{c|}{Cycle 1} & \multicolumn{3}{c|}{Cycle 2} \\ \cline{3-8}
 &  & Domain & Error (\%) & Avg & Domain & Error (\%) & Avg \\ \hline
 
\multirow{15}{*}{SloMo-Fast} & \multirow{3}{*}{Noise} & gaussian & 68.72 & \multirow{3}{*}{66.33} & gaussian & 64.82 & \multirow{3}{*}{63.58} \\ \cline{3-4} \cline{6-7}
 &  & shot & 65.56 &  & shot & 62.82 &  \\ \cline{3-4} \cline{6-7}
 &  & impulse & 64.73 &  & impulse & 63.09 &  \\ \cline{2-8}
 
 & \multirow{4}{*}{Blur} & defocus & 68.51 & \multirow{4}{*}{60.70} & defocus & 66.23 & \multirow{4}{*}{59.46} \\ \cline{3-4} \cline{6-7}
 &  & glass & 66.71 &  & glass & 65.77 &  \\ \cline{3-4} \cline{6-7}
 &  & motion & 57.39 &  & motion & 56.28 &  \\ \cline{3-4} \cline{6-7}
 &  & zoom & 50.19 &  & zoom & 49.57 &  \\  \cline{2-8}
 
 & \multirow{3}{*}{Weather} & snow & 50.94 & \multirow{3}{*}{50.20} & snow & 49.47 & \multirow{3}{*}{ 49.30} \\ \cline{3-4} \cline{6-7}
 &  & frost & 56.59 &  & frost & 55.90 &  \\ \cline{3-4} \cline{6-7}
 &  & fog & 43.08 &  & fog & 42.52 &  \\ \cline{2-8}
 
 & \multirow{2}{*}{Digital} & brightness & 33.61 & \multirow{2}{*}{25.01} & brightness & 33.65 & \multirow{2}{*}{44.88} \\ \cline{3-4} \cline{6-7}
 &  & contrast & 56.98 &  & contrast & 56.12 &  \\ \cline{2-8}
 
 & \multirow{3}{*}{Distortion} & elastic & 43.65 & \multirow{3}{*}{43.55} & elastic & 43.05 & \multirow{3}{*}{43.25} \\ \cline{3-4} \cline{6-7}
 &  & pixel & 41.21 &  & pixel & 40.93 &  \\ \cline{3-4} \cline{6-7}
 &  & jpeg & 45.80 &  & jpeg & 45.77 &   \\ \hline
 
& & \multicolumn{3}{c|}{Cycle 1 Avg: 53.22\%} & \multicolumn{3}{c|}{Cycle 2 Avg: 52.09\%} \\ \hline
\end{tabular}

\caption{Detailed Evaluation Results for {\pa} on Imagenet-C under Cyclic Domain Settings}
\label{table:cyclic_ours_imagenetc}
\end{table*}

\begin{table*} 
\centering
\begin{tabular}{|c|c|c|c|c|c|c|c|}
\hline
\multirow{3}{*}{Method} & \multirow{3}{*}{Subgroup} & \multicolumn{3}{c|}{Cycle 1} & \multicolumn{3}{c|}{Cycle 2} \\ \cline{3-8}
 &  & Domain & Error (\%) & Avg & Domain & Error (\%) & Avg \\ \hline
 
\multirow{15}{*}{{\pa}*} & \multirow{3}{*}{Noise} & gaussian & 21.65 & \multirow{3}{*}{20.62} & gaussian & 20.34 & \multirow{3}{*}{20.62} \\ \cline{3-4} \cline{6-7}
 &  & shot & 19.78 &  & shot & 21.12 &  \\ \cline{3-4} \cline{6-7}
 &  & impulse & 20.43 &  & impulse & 20.40 &  \\ \cline{2-8}
 
 & \multirow{4}{*}{Blur} & defocus & 15.03 & \multirow{4}{*}{14.79} & defocus & 14.34 & \multirow{4}{*}{15.21} \\ \cline{3-4} \cline{6-7}
 &  & glass & 14.65 &  & glass & 15.92 &  \\ \cline{3-4} \cline{6-7}
 &  & motion & 17.07 &  & motion & 16.27 &  \\ \cline{3-4} \cline{6-7}
 &  & zoom & 12.41 &  & zoom & 13.90 &  \\  \cline{2-8}
 
 & \multirow{3}{*}{Weather} & snow & 13.72 & \multirow{3}{*}{13.21} & snow & 13.10 & \multirow{3}{*}{13.09} \\ \cline{3-4} \cline{6-7}
 &  & frost & 13.42 &  & frost & 13.23 &  \\ \cline{3-4} \cline{6-7}
 &  & fog & 12.48 &  & fog & 12.58 &  \\ \cline{2-8}
 
 & \multirow{2}{*}{Digital} & brightness & 9.33 & \multirow{2}{*}{10.24} & brightness & 9.17 & \multirow{2}{*}{10.23} \\ \cline{3-4} \cline{6-7}
 &  & contrast & 11.15 &  & contrast & 11.26 &  \\ \cline{2-8}
 
 & \multirow{3}{*}{Distortion} & elastic & 15.85 & \multirow{3}{*}{13.96} & elastic & 15.64 & \multirow{3}{*}{14.02} \\ \cline{3-4} \cline{6-7}
 &  & pixel & 11.56 &  & pixel & 11.98 &  \\ \cline{3-4} \cline{6-7}
 &  & jpeg & 14.46 &  & jpeg & 14.61 &   \\ \hline
 
& & \multicolumn{3}{c|}{Cycle 1 Avg: 14.89\%} & \multicolumn{3}{c|}{Cycle 2 Avg: 14.38\%} \\ \hline
\end{tabular}
\caption{Detailed Evaluation Results for {\pa}* on CIFAR10-C under Cyclic Domain Settings}
\label{table:cyclic_ours*_cifar10c}
\end{table*}

\begin{table*}
\centering
\begin{tabular}{|c|c|c|c|c|c|c|c|}
\hline
\multirow{2}{*}{Method} & \multirow{2}{*}{Subgroup} & \multicolumn{3}{c|}{Cycle 1} & \multicolumn{3}{c|}{Cycle 2} \\ \cline{3-8}
 &  & Domain & Error (\%) & Avg & Domain & Error (\%) & Avg \\ \hline
 
\multirow{15}{*}{SloMo-Fast*} & \multirow{3}{*}{Noise} & gaussian & 34.12 & \multirow{3}{*}{33.29} & gaussian & 34.82 & \multirow{3}{*}{33.51} \\ \cline{3-4} \cline{6-7}
 &  & shot & 32.54 &  & shot & 31.89 &  \\ \cline{3-4} \cline{6-7}
 &  & impulse & 33.12 &  & impulse & 33.84 &  \\ \cline{2-8}
 
 & \multirow{4}{*}{Blur} & defocus & 27.01 & \multirow{4}{*}{27.02} & defocus & 27.03 & \multirow{4}{*}{27.02} \\ \cline{3-4} \cline{6-7}
 &  & glass & 26.97 &  & glass & 27.04 &  \\ \cline{3-4} \cline{6-7}
 &  & motion & 26.89 &  & motion & 27.12 &  \\ \cline{3-4} \cline{6-7}
 &  & zoom & 27.22 &  & zoom & 26.89 &  \\  \cline{2-8}
 
 & \multirow{3}{*}{Weather} & snow & 27.00 & \multirow{3}{*}{26.96} & snow & 26.98 & \multirow{3}{*}{26.96} \\ \cline{3-4} \cline{6-7}
 &  & frost & 26.90 &  & frost & 27.01 &  \\ \cline{3-4} \cline{6-7}
 &  & fog & 26.98 &  & fog & 26.89 &  \\ \cline{2-8}
 
 & \multirow{2}{*}{Digital} & brightness & 24.74 & \multirow{2}{*}{24.84} & brightness & 24.41 & \multirow{2}{*}{24.84} \\ \cline{3-4} \cline{6-7}
 &  & contrast & 25.05 &  & contrast & 25.18 &  \\ \cline{2-8}
 
 & \multirow{3}{*}{Distortion} & elastic & 25.72 & \multirow{3}{*}{25.89} & elastic & 25.67 & \multirow{3}{*}{25.79} \\ \cline{3-4} \cline{6-7}
 &  & pixel & 24.79 &  & pixel & 24.97 &  \\ \cline{3-4} \cline{6-7}
 &  & jpeg & 26.84 &  & jpeg & 26.73 &   \\ \hline
 
& & \multicolumn{3}{c|}{Cycle 1 Avg: 27.98\%} & \multicolumn{3}{c|}{Cycle 2 Avg: 27.18\%} \\ \hline
\end{tabular}

\caption{Detailed Evaluation Results for {\pa}* on CIFAR100-C under Cyclic Domain Settings}
\label{table:cyclic_ours*_cifar100c}
\end{table*}


    

\end{document}